%% file: main.tex
\documentclass[10pt,twocolumn,letterpaper]{article}

\usepackage[pagenumbers]{iccv} 


\input{math_commands}

\input{my_cls}

\definecolor{iccvblue}{rgb}{0.21,0.49,0.74}
\definecolor{cvprblue}{rgb}{0.21,0.49,0.74}
\usepackage[pagebackref,breaklinks,colorlinks,allcolors=iccvblue]{hyperref}

\def\paperID{13305} 
\def\confName{ICCV}
\def\confYear{2025}

\title{Transformed Low-rank Adaptation via~Tensor~Decomposition and \\ Its Applications to Text-to-image Models}

\author{
Zerui Tao$^{1, \dagger}$ \quad Yuhta Takida$^2$ \quad Naoki Murata$^2$ \quad Qibin Zhao$^1$ \quad Yuki Mitsufuji$^{2, 3}$ \\
$^1$RIKEN AIP \quad $^2$Sony AI \quad $^3$Sony Group Corporation \\
{\tt\small \{zerui.tao,qibin.zhao\}@riken.jp, \quad \{yuta.takida,naoki.murata,yuhki.mitsufuji\}@sony.com}
}

\begin{document}

\twocolumn[{
\renewcommand\twocolumn[1][]{#1}%
\vspace{-1.0cm} 
\maketitle
\begin{center}
    \centering
    \includegraphics[width=\textwidth]{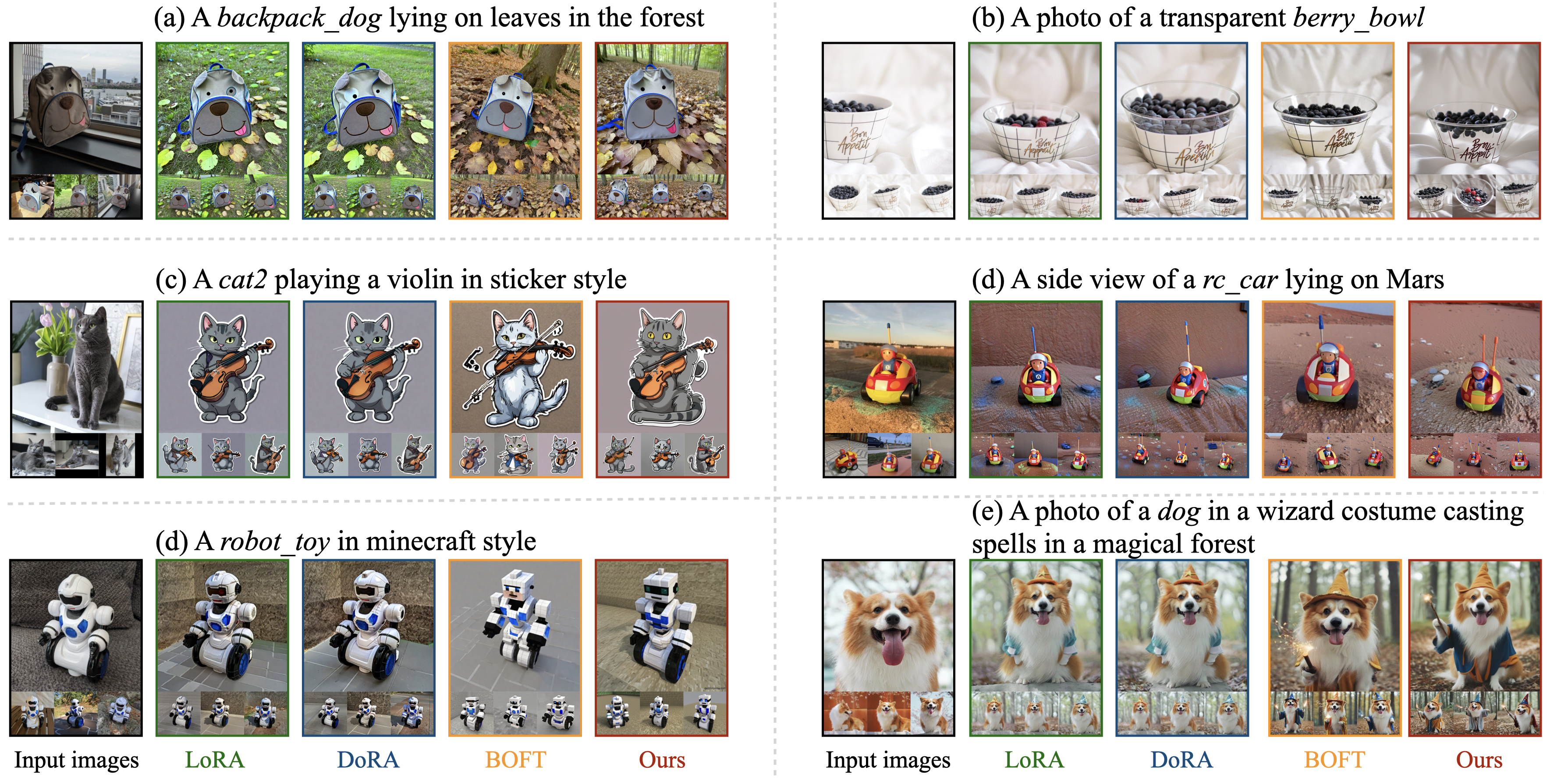}
    \captionof{figure}{Qualitative comparison of the subject-driven generation results. Results are generated using fine-tuned checkpoint by each method. For each model, we randomly generate four images, all of which are shown here. The subcaptions are the given prompts, and the subjects in italics represent training subjects.}\label{fig:db-images}
\end{center}
}]

\def\thefootnote{$\dagger$}\footnotetext{Work done during an internship at Sony AI}\def\thefootnote{\arabic{footnote}}

\input{contents/0_abs}    
\addtocontents{toc}{\protect\setcounter{tocdepth}{0}}
\input{contents/1_intro}
\input{contents/2_related_work}
\input{contents/3_model}

\input{contents/4_exp}

\input{contents/5_conclusion}
{
    \small
    \bibliographystyle{ieeenat_fullname}
    \bibliography{main}
}

\input{contents/appendix}

\end{document}

%% file: math_commands.tex
\usepackage{amsmath,amsfonts,bm}

\def\eqref#1{equation~\ref{#1}}

\def\1{\bm{1}}

\def\vzero{{\bm{0}}}
\def\vone{{\bm{1}}}

\def\vm{{\bm{m}}}

\def\vx{{\bm{x}}}
\def\vy{{\bm{y}}}

\def\mA{{\bm{A}}}
\def\mB{{\bm{B}}}

\def\mF{{\bm{F}}}

\def\mI{{\bm{I}}}

\def\mT{{\bm{T}}}

\def\mW{{\bm{W}}}
\def\mX{{\bm{X}}}
\def\mY{{\bm{Y}}}
\def\mZ{{\bm{Z}}}

\DeclareMathAlphabet{\mathsfit}{\encodingdefault}{\sfdefault}{m}{sl}
\SetMathAlphabet{\mathsfit}{bold}{\encodingdefault}{\sfdefault}{bx}{n}
\newcommand{\tens}[1]{\bm{\mathsfit{#1}}}
\def\tA{{\tens{A}}}
\def\tB{{\tens{B}}}
\def\tC{{\tens{C}}}

\def\tX{{\tens{X}}}

\def\gF{{\mathcal{F}}}

\def\gN{{\mathcal{N}}}

\def\gR{{\mathcal{R}}}



%% file: my_cls.tex
\usepackage{amssymb,amsthm}
\usepackage{mathtools}
\usepackage{siunitx}
\usepackage{multirow}
\usepackage{sidecap}

\usepackage{listings}

\newcommand{\bftab}{\fontseries{b}\selectfont}

\newcommand{\tr}{\mathrm{tr}}

\newtheorem{prop}{Proposition}

\theoremstyle{definition}

\AtEndPreamble{%
    \crefname{prop}{Prop.}{Props.}
    \Crefname{prop}{Proposition}{Propositions}
}

%% file: contents/0_abs.tex
\begin{abstract}
Parameter-Efficient Fine-Tuning (PEFT) of text-to-image models has become an increasingly popular technique with many applications.
Among the various PEFT methods,
Low-Rank Adaptation (LoRA) and its variants have gained significant attention due to their effectiveness, enabling users to fine-tune models with limited computational resources.
However, the approximation gap between the low-rank assumption and desired fine-tuning weights prevents the simultaneous acquisition of ultra-parameter-efficiency and better performance.
To reduce this gap and further improve the power of LoRA, we propose a new PEFT method that combines two classes of adaptations, namely, transform and residual adaptations.
In specific, we first apply a \textbf{full-rank} and \textbf{dense} transform to the pre-trained weight.
This learnable transform is expected to align the pre-trained weight as closely as possible to the desired weight, thereby reducing the rank of the residual weight.
Then, the residual part can be effectively approximated by more compact and parameter-efficient structures, with a smaller approximation error.
To achieve ultra-parameter-efficiency in practice, we design highly flexible and effective tensor decompositions for both the transform and residual adaptations.
Additionally, popular PEFT methods such as DoRA can be summarized under this transform plus residual adaptation scheme.
Experiments are conducted on fine-tuning Stable Diffusion models in subject-driven and controllable generation.
The results manifest that our method can achieve better performances and parameter efficiency compared to LoRA and several baselines.
\end{abstract}

%% file: contents/1_intro.tex
\section{Introduction}

In recent years, text-to-image (T2I) generative models \cite{ramesh2022hierarchical,rombach2022high,saharia2022photorealistic,podell2024sdxl,esser2024scaling} have achieved remarkable results in image synthesis.
In order to obtain the desired generative power,
these models typically consist of hundreds of millions or even billions of parameters, which are trained on huge image and text datasets.
Despite their incredible abilities, the high computational and memory costs largely prevent users from taking advantage of these models on their personalized or private datasets and tasks, such as subject-driven generation \cite{gal2023an,ruiz2023dreambooth}, controllable generation \cite{mou2024t2i,zhang2023adding}, and others \cite{xue2024raphael,gu2024mix}.
For subject-driven generation, users aim to adapt the pre-trained model to several images of the target subject, so that the model can generate new images of this subject given text prompts.
Similarly, controllable generation involves fine-tuning the model on image, text, and control signal pairs, such as landmarks, segmentations and canny edges, to facilitate the ability of generation conditioned on these signals.

To circumvent the huge resource requirements for fine-tuning the T2I models, Parameter-Efficient Fine-Tuning (PEFT) has become an important research topic.
In particular, Low-Rank Adaptation \cite[LoRA,][]{hu2022lora} and its variants have become the most widely adopted due to their simplicity, stability, and efficiency \cite{fomenko2024note}.
By assuming the fine-tuning adaptations have a low intrinsic rank, LoRA methods can greatly reduce the computational and memory costs by only updating the low-rank adaptation.
While this simple approximation works surprisingly well, it still faces several challenges and issues.
First, the low-rank approximation may yield a high recovery gap compared to the full fine-tuning adaptations, especially for difficult tasks \cite{nikdan2024rosa}.
Second, the simple matrix decomposition structure lacks flexibility in terms of adjusting the fine-tuning budget and further compression of large matrices.
Therefore, both the fine-tuning performance and the parameter efficiency of LoRA tend to be sub-optimal and can be further improved.

To address these issues, we investigate a novel PEFT method called Transformed Low-Rank Adaptation (TLoRA) via tensor decomposition.
The proposed method consists of two adaptation parts, namely, \textbf{Transform} and \textbf{Residual adaptations}.
The \textbf{Transform adaptation} applies a linear transform to the pre-trained weight so that it can be projected into a space with a lower-rank fine-tuning process. This is necessary because the pre-trained weight may not have a low-rank fine-tuning weight. Subsequently, the \textbf{Residual adaptation} approximates the residual part using more compact and efficient structures.
To effectively parameterize these adaptations, tensor decomposition techniques are adopted (as described below).
Moreover, we empirically demonstrate the effectiveness of combining these two adaptations by investigating some pre-trained and fine-tuned models in \cref{fig:motivatoin_param_sdxl_inpaint} and \cref{sec:motivation}.

\vspace{-1em}
\paragraph{Transform adaptation.}
The purpose of this learnable transform is to align the pre-trained weight as closely as possible to some target weight in order to reduce the rank of the residual adaptation.
Since the target weight space (e.g., optimal weights for the fine-tuning task) is not likely to be low-rank, we assume the transform also has a full-rank structure.
Furthermore, the transform is parameterized to be a dense matrix, allowing the information transmission among all neurons \cite{liu2024parameterefficient}.
However, since this transform matrix is large, it should be represented by parameter-efficient structures.
To meet all these requirements, we adopt the tensor-ring matrix \cite[TRM,][]{oseledets2011tensor,cichocki2016tensor,zhao2016tensor} format for the transform, which is a highly compact form for \textbf{full-rank and dense matrices}, as described in \cref{sec:ttm}.

\vspace{-1em}
\paragraph{Residual adaptation.}

Assuming the effectiveness of the transform, the residual adaptation can be effectively approximated by more compact and parameter-efficient structures.
While many efficient parameterizations are available, we focus on the tensor-ring  decomposition \cite[TR,][]{zhao2016tensor} in this work.
In our study, we implement TR using a different initialization strategy with previous PEFT literature using similar structures \cite{yang2024loretta,anjum2024tensor,chen2024quanta}, showing that TR with the transform adaptation achieves promising performances with ultra-parameter-efficiency. Details are presented in \cref{sec:residual}

\vspace{-1em}
\paragraph{}
Additionally, in \cref{sec:connection}, we show that while some popular PEFT methods (e.g., DoRA \cite{liu2024dora} and Fourier-inspired LoRA \cite{borse2024foura,gao2024parameterefficient,sehanobish2024structured}) adopt the idea of transform implicitly or explicitly, they are parameterized by either extremely sparse or fixed transforms, which may be insufficient.

To demonstrate the advantages of our model,
we conduct experiments on two tasks, subject-driven generation and controllable generation, using Stable Diffusion XL \cite[SDXL,][]{podell2024sdxl} and Stable Diffusion \cite{rombach2022high} v1.5 models respectively.
Compared to LoRA and several baselines, the results show that the transform part can effectively boost performances in most cases.
Additionally, by using the compact tensor decompositions, our model is able to achieve desirable performances and ultra-parameter-efficiency simultaneously, e.g., fine-tuning SDXL with only 0.4M parameters in \cref{fig:db-images}.

%% file: contents/2_related_work.tex
\section{Related work}

\paragraph{Text-to-image model personalization.}

T2I models have shown exceptional results in image synthesis \cite{ramesh2022hierarchical,rombach2022high,saharia2022photorealistic,podell2024sdxl,esser2024scaling}.
To personalize pre-trained models,
\citet{gal2023an} propose learning given subjects via textual inversion, while \citet{ruiz2023dreambooth} fine-tune the whole model.
ControlNet \cite{zhang2023adding} incorporates an additional network branch that can learn datasets of paired control signals and images.
While these methods may have large numbers of trainable parameters,
\citet{kumari2023multi} show that fine-tuning the cross-attention layers alone is effective enough for these tasks.
More recently, many works have focused on developing PEFT methods for these tasks \cite{han2023svdiff,qiu2023controlling,gu2024mix,yeh2024navigating,zhang2024spectrum,borse2024foura,xie2023difffit,NEURIPS2024_18c0102c,hu2025sara,chen2025para,zhuang2024time}.
There are also training-free approaches \cite{rout2024rb}, which could be slow at inference.

\vspace{-1em}
\paragraph{Parameter-efficient fine-tuning.}

Popular PEFT methods include Adapter \cite{houlsby2019parameter}, Prefix-tuning \cite{li2021prefix}, Prompt-tuning \cite{lester2021power}, LoRA \cite{hu2022lora}, and many of their variants.
LoRA has become the most popular PEFT method due to its simplicity and impressive performances \cite{fomenko2024note}.
Many variants of LoRA have been proposed \cite{zhang2023adaptive,liu2024dora,qiu2023controlling,liu2024parameterefficient,nikdan2024rosa,kopiczkovera,hyeon-woo2022fedpara,NEURIPS2024_2c570b0f,NEURIPS2024_68716328,huang2025hira,chen2024quanta}.
In DoRA \cite{liu2024dora}, the pre-trained weight is decomposed into magnitude and direction, whereas vanilla LoRA is applied to the direction. In the current work, we show that DoRA can also be connected to our method as utilizing a diagonal transform. Orthogonal Fine-Tuning \cite[OFT,][]{qiu2023controlling} applies a learnable orthogonal transform for adaptation. However, for parameter efficiency, OFT adopts block diagonal matrices, which are highly sparse. Subsequently, many methods aim to improve OFT by applying particular dense transform structures \cite{liu2024parameterefficient,maparameter,biniether,yuan2024bridging,zhang2024spectrum}. Our method adopts a similar idea of using a transform, but we design a different dense matrix parameterization using tensor decomposition.
Additionally, several works \cite{gao2024parameterefficient,borse2024foura,sehanobish2024structured} also share similarities with ours by using \emph{pre-defined and fixed} transforms.

\vspace{-1em}
\paragraph{Tensor decomposition.}

TD is a classical tool in signal processing and machine learning \cite{cichocki2016tensor}.
In particular, tensor-train \cite[TT,][]{oseledets2011tensor} and its extension, tensor-ring \cite[TR,][]{zhao2016tensor}, have shown exceptional results in model compression,
including MLP \cite{novikov2015tensorizing}, CNN \cite{wang2018wide,garipov2016ultimate}, RNN/LSTM \cite{pan2019compressing,su2020convolutional,yang2017tensor,mehta2019scaling} and Transformer \cite{ma2019tensorized,pham2022tt}.
Recently, TDs have also been applied to PEFT \citep{jie2023fact,yang2024loretta,anjum2024tensor,chen2024quanta}.
While these works utilize similar TT/TR structures to our model, they do not apply the transform adaptation.
Moreover, we study a different initialization strategy for the TR factors to make the training more stable, as we will show in our experiments.

%% file: contents/3_model.tex
\section{Proposed model}

\subsection{Preliminaries}

\paragraph{Notations.}

Notations for tensors, matrices, vectors and scalars basically follow the conventions in the \emph{Deep Learning} textbook \cite{Goodfellow-et-al-2016,dlbook_notation}.
Furthermore, we use square brackets to denote entries or slices of an array. For example, given a tensor $\tX \in \mathbb{R}^{I \times J \times K}$, we have matrix slices $\mX_{i::} = \tX[i, :, :] \in \mathbb{R}^{J \times K}$, vector slices $\vx_{ij:} = \tX[i, j, :] \in \mathbb{R}^{K}$ and scalar entries $x_{ijk} = \tX[i, j, k] \in \mathbb{R}$.
We use $\mI_I$ to denote identity matrix of shape $I  \times I$.
The Kronecker product is denoted by $\otimes$. The matrix trace is denoted by $\tr(\cdot)$. $\text{diag}(\vm)$ denotes diagonal matrix with diagonal elements $\vm$.
For indices of tensorization, we adopt the little-endian convention \cite{cichocki2016tensor}. For a vector $\vx \in \mathbb{R}^{I}$ with $I = \prod_{d=1}^D I_d$, if it is tensorized into $\tX \in \mathbb{R}^{I_1 \times \cdots I_D}$, we have $\vx[\overline{i_1\dots i_D}] = \tX[i_1, \dots, i_D]$, where $\overline{i_1\dots i_D} = i_1 + (i_2 - 1) I_1 + (i_3 - 1) I_1 I_2 + \cdots (i_D - 1) I_1 \cdots I_{D-1}$.

\vspace{-1em}
\paragraph{Low-rank adaptation \cite{hu2022lora}.}

Consider a linear layer $\vy = \mW_0 \vx$, where the weight has shape $I \times I$. Given a pre-trained weight $\mW_0$, LoRA aims to learn an additive adaptation $\bm{\Delta}$ of the same shape, i.e.,
\begin{equation}\label{eq:lora-original}
\vy' = (\mW_0 + \bm{\Delta}) \vx, \quad s.t. \,\bm{\Delta} = \mB \mA,
\end{equation}
where the adaptation is parameterized by a low-rank matrix decomposition to reduce trainable parameters.
The low-rank matrices $\mB \in \mathbb{R}^{I \times R}$ and $\mA \in \mathbb{R}^{R \times I}$ are optimized through the given fine-tuning tasks, and have a total of $2 I R$ trainable parameters, which is much smaller than the original size $I^2$ if $R \ll I$. However, as some works \cite{nikdan2024rosa,borse2024foura} indicate, the desired fine-tuning weight may not be low-rank.

\vspace{-1em}
\paragraph{Orthogonal fine-tuning \cite{qiu2023controlling}.}

Instead of the additive adaptation in \cref{eq:lora-original}, OFT introduces an orthogonal transformation of the pre-trained weight for adaptation,
\begin{equation}\label{eq:oft-original}
\vy' = (\mW_0 \mT) \vx,
\end{equation}
where $\mT$ is a trainable orthogonal matrix of shape $I \times I$. However, directly optimizing $\mT$ is computationally infeasible due to its large size. In \cite{qiu2023controlling}, $\mT$ is parameterized as a block diagonal matrix, which is extremely sparse for small parameter budgets.
This sparsity is considered to reduce the information transfer and connectivity among neurons, which is deleterious to the fine-tuning.
Therefore, parameter-efficient matrices with dense entries are more desirable for OFT \cite{liu2024parameterefficient,maparameter,biniether,yuan2024bridging,zhang2024spectrum}.

\input{contents/fig_and_tab/sdxl_inpaint_param}

\subsection{Our motivation}\label{sec:motivation}

Supposing the target of fine-tuning is to approximate some desired weight $\mW_*$, the assumption behind LoRA is that the difference $\bm{\Delta}_* = \mW_* - \mW_0$ is low-rank. However, this is unlikely to be true in real large foundation models \cite{nikdan2024rosa}.

Considering this approximation problem, if the target $\bm{\Delta}_*$ has a high rank, LoRA inevitably causes high approximation error, as illustrated in \cref{fig:motivation-rotation,fig:motivation-true}.
To address this issue, we first apply a learnable linear transform $\mT$ on $\mW_0$ to align $\mW_0$ on $\mW_*$ and then approximate the residual part by another compact structure. Here, the difference becomes $\bm{\Delta}'_* = \mW_* - \mW_0 \mT$.
After the transform, $\bm{\Delta}'_*$ should have a smaller rank than the original $\bm{\Delta}_*$ so that
we can use much more compact structures for approximation of this residual part.
The overall fine-tuning structure becomes,
\begin{equation}\label{eq:proposed-model}
    \vy' = (\mW_0 {\color{cvprblue} \mT} + {\color{cvprblue} \bm{\Delta}}) \vx,
\end{equation}
where $\mT$ and $\bm{\Delta}$ learnable compact parameterizations.
To meet different requirements and desired properties,
we design TRM form \cite{oseledets2011tensor,zhao2016tensor} for the transform $\mT$ and TR form \cite{oseledets2011tensor,zhao2016tensor} for the residual $\bm{\Delta}$ in \cref{sec:ttm,sec:residual} respectively.

To empirically illustrate the effectiveness of the transform, we conduct a simulation study on the pretrained SDXL \cite{podell2024sdxl} model and the fine-tuned SDXL-Inpaint \cite{SDXL-inpaint} model.
We consider three settings: (a) Additive low-rank difference, where the desired weight $\mW_*$ is generated by some low-rank update $\mW_* = \mW_0 + \mB_* \mA_*$. (b) Orthogonal rotation, where the desired weight $\mW_*$ is generated by an orthogonal transform $\mW_* = \mW_0 \mT_*$. (c) True fully fine-tuned weight, where $\mW_*$ is obtained by full-parameter fine-tuning, which is identical to SDXL-Inpaint \cite{SDXL-inpaint}.
We approximate $\mW_*$ using:
(\textsc{I}) OFT in \cref{eq:oft-original}.
(\textsc{II}) LoRA in \cref{eq:lora-original}.
(\textsc{III}) TR, which replaces $\bm{\Delta}$ in \cref{eq:lora-original} with TR.
(\textsc{IV}) OFT + LoRA, which uses OFT for $\mT$ and LoRA for $\bm{\Delta}$ in \cref{eq:proposed-model}.
(\textsc{V}) TRM + LoRA, which uses TRM for $\mT$ and LoRA for $\bm{\Delta}$ in \cref{eq:proposed-model}.
(\textsc{VI}) TRM + TR, which uses TRM for $\mT$ and TR for $\bm{\Delta}$ in \cref{eq:proposed-model}.
To showcase the expressiveness, we illustrate the approximation error with different adaptation budgets in \cref{fig:motivatoin_param_sdxl_inpaint}.
More details and results on different layers/models can be found in \cref{app-sec:motivation}.
Based on the results, we have several observations.
\begin{itemize}
    \item LoRA and OFT work well under their own assumptions, i.e. \cref{fig:motivation-lora,fig:motivation-rotation} respectively, which may not hold in the real case (\cref{fig:motivation-true}). Adding the TRM transform can take the best of both worlds and performs well across these settings.
    More importantly, the TRM transform generally improves LoRA and TR in \cref{fig:motivation-rotation,fig:motivation-true}.
    \item Compared to LoRA, the TR can be advantageous when the parameter budget is extremely small. In \cref{fig:motivation-true}, TRM + TR achieves comparable error with LoRA  $R=1$ (the leftmost point) using less than $10\%$ sizes. TR can also adapt the budget size more smoothly.
    However, when the rank increases, the improvement may not be significant.
    \item The simple combination OFT+LoRA does not reduce the approximation error. In \cref{fig:motivation-true}, it introduces additional parameters compared to LoRA, while the error does not change. We hypothesis this is because that the orthogonal transform does not change the column space spanned by the pre-trained weight, therefore limits the ability to align with the target weight $\mW_*$.
\end{itemize}
Although this is a simple simulation study, these observations align well with our experiments in \cref{sec:exp}.

\subsection{Tensor-ring matrix transform adaptation}\label{sec:ttm}

In this subsection, we introduce the structure of the transform $\mT$. Recall that the purpose of $\mT$ is to align the pre-trained weight $\mW_0$ to the target weight $\mW_*$ as closely as possible, in order to reduce the rank of the residual adaptation $\bm{\Delta}$.
Since the target weight space (e.g., optimal weights for the fine-tuning task) is not likely to be low-rank, we assume $\mT$ also has a full-rank structure.
Furthermore, to address the sparsity problem in OFT, $\mT$ is expected to have dense entries. As this transform matrix is large, it should be represented by parameter-efficient structures.

To meet the above requirements, we adopt the tensor-ring matrix (TRM) form \cite{oseledets2011tensor,zhao2016tensor,cichocki2016tensor}.
Given a matrix $\mT \in \mathbb{R}^{I \times J}$, suppose $I = \prod_{d=1}^D I_d$, $J = \prod_{d=1}^D J_d$ and that it can be re-arranged (tensorized) into many sub-arrays.
The TRM factorizes the matrix into contractions of $D$ 4th-order factors $\tA^d \in \mathbb{R}^{I_d \times J_d \times R_{d} \times R_{d+1}}, \forall d = 1, \dots, D$, which are called core tensors.
The sequence $[R_1, \dots, R_{D+1}]$ with $R_{D+1} = R_1$ is the TR rank. When one of the $R_d$ is equal to 1, TRM reduces to the tensor-train matrix \cite[TTM,][]{oseledets2011tensor} form. For simplicity, we assume $R = R_1 = \cdots = R_{D+1}$ and $I = J$ throughout this work.
The TRM assumes each entry of the matrix is computed by,
\begin{equation}\label{eq:ttm-def}
\begin{multlined}[0.8\linewidth]
\mT[\overline{i_1 \cdots i_D}, \overline{j_1 \cdots j_D}] = \tr(\tA^1[i_1, j_1, :, :] \cdots \tA^D[i_D, j_D, :, :]).
\end{multlined}
\end{equation}
For simplicity, we denote the TRM format as $\mT = \text{TRM}(\tA^{1:D})$, where $\tA^{1:D}$ denotes $\{\tA^1, \dots, \tA^D \}$.

\begin{SCfigure}[1][t]
    \centering
    \vspace{-0.5\baselineskip}
    \includegraphics[width=0.5\linewidth]{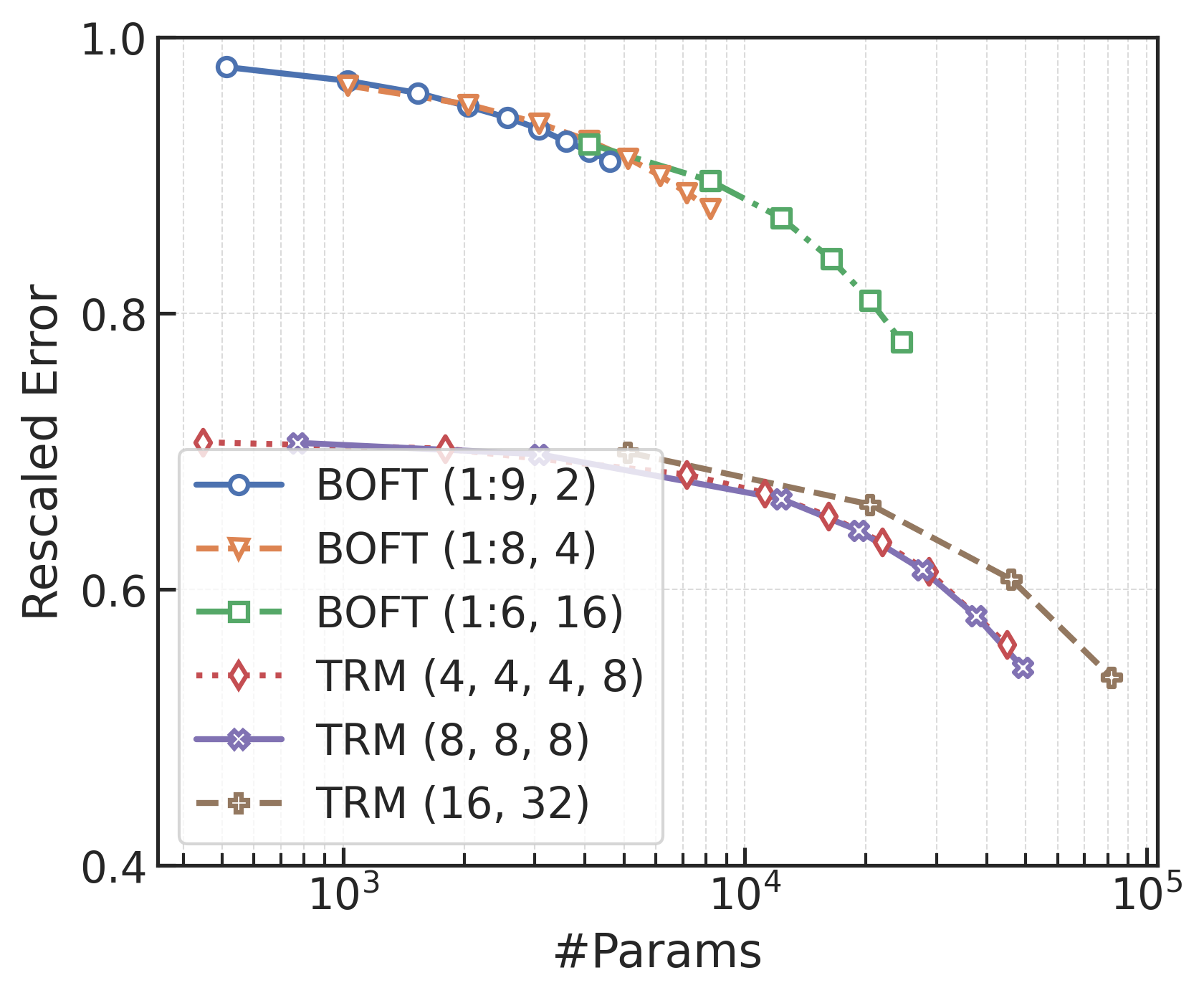}
    \caption{Expressiveness of BOFT and TRM. For BOFT, the first number in the bracket indicates the number of blocks, and the second number indicates the number of Butterfly factors. For TRM, the numbers denote the sizes of sub-indices $I_d$.}\label{fig:boft-vs-ttm}
\end{SCfigure}

TRM is a highly compact form for representation of dense and full-rank matrices. If the factors $\tA^{1:D}$ are dense and full-rank, then $\text{TRM}(\tA^{1:D})$ becomes a dense and full-rank matrix.
Moreover, the memory cost of the TRM is $\mathcal{O}(D I^{2/D} R^2)$ if $I = J$, which is much lower than the original $\mathcal{O}(I^2)$ if we adjust proper hyper-parameters $D$ and $R$.
To show the expressive power of TRM, we conduct a simulation study to compare it with the Butterfly OFT \cite[BOFT,][]{liu2024parameterefficient}, which is proposed to represent dense matrices in OFT.
Specifically, we randomly generate an orthogonal matrix with shape $512 \times 512$, and approximate it using BOFT and TRM.
The approximation error for different parameter sizes is shown in \cref{fig:boft-vs-ttm}.
TRM achieves a smaller approximation error when the number of parameters is small. Moreover, as we increase the number of parameters, the approximation error of TRM consistently decreases and is better than that of BOFT.
More details are provided in \cref{app-sec:model}.

\vspace{-1em}
\paragraph{Initialization.}

To ensure that the fine-tuned model is the same as the original model at initialization \cite{hu2022lora}, 
the transform $\mT$ should be an identity matrix. Fortunately, for TRM, we can easily construct an identity matrix as follows.

\begin{prop}\label{prop:init}
    For the TRM in \cref{eq:ttm-def},
    if we initialize every factor as
    \[
    \tA^d[:, :, r, r'] = \mI_{I_d} / R, \, \forall d = 1, \dots, D, \text{and} \, r, r' = 1, \dots, R,
    \]
    the resulting TRM $\mT$ is an identity matrix.
\end{prop}

\vspace{-1em}
\paragraph{Constrained transform.}

While the transform can be powerful, sometimes it is beneficial to have constraints on it, limiting the fine-tuned weight to be within a small region of the pre-trained one.
In particular, previous methods adopt identity regularization \cite{qiu2023controlling} or orthogonal regularization \cite{biniether,maparameter,yuan2024bridging} during fine-tuning.
However, directly computing these regularizations on the transform $\mT$ could be computationally expensive. By using the TRM, we show that they can be computed efficiently on the core tensors $\tA^{1:D}$ with much smaller sizes.

For identity regularization $\lVert \mT - \mI_I \rVert_F$, we have
\begin{equation*}
    \mathcal{R}_I(\tA^{1:D}) = \sum_{d=1}^D \sum^{R}_{r, r'=1} \left\lVert \tA^d[:, :, r, r'] - \mI_{I_d} / R \right\rVert_F.
\end{equation*}
According to \cref{prop:init}, this would encourage the transform to be identity.
For orthogonal regularization, directly computing $\lVert \mT \mT^\intercal - \mI_I \rVert_F$ is computationally expensive due to the additional matrix multiplication. Instead, we have
\begin{equation*}
\begin{multlined}[0.8\linewidth]
    \mathcal{R}_O(\tA^{1:D}) = \\
    \sum_{d=1}^D \sum^{I_d, I_d}_{i_d, j_d=1} \left\lVert \sum^{I_d}_{l=1} (\tA^d[i_d, l, :, :] \otimes \tA^d[j_d, l, :, :]) - \mI_{R^2} / R \right\rVert_F.
\end{multlined}
\end{equation*}

\begin{prop}\label{prop:orthogonal}
The matrix product of two TRM $\mX = \text{TRM}(\tA^{1:D})$ and $\mY = \text{TRM}(\tB^{1:D})$ is still a TRM $\mX \mY^\intercal = \text{TRM}(\tC^{1:D})$, where each core tensor satisfies $\tC^d[i_d, j_d, :, :] = \sum_{l_d} \tA^d[i_d, l_d, :, :] \otimes \tB^d[j_d, l_d, :, :]$ for each $d = 1, \dots, D$ and $i_d, j_d = 1, \dots, I_d$.
\end{prop}

Therefore, combining \cref{prop:init,prop:orthogonal}, we can compute the orthogonal regularization as shown above. In practice, we observe that the identity regularization performs better than the orthogonal regularization. This observation is consistent with the findings in \cite{qiu2023controlling,biniether,maparameter}.

\subsection{Tensor-ring residual adaptation}\label{sec:residual}

Given the powerful transform, we expect the residual part $\bm{\Delta}$ in \cref{eq:proposed-model} to be approximated by very compact structures.
Specifically,
we adopt the tensor-ring (TR) decomposition \cite{zhao2016tensor}.
Suppose the residual part $\bm{\Delta} \in \mathbb{R}^{I \times J}$, with $I = \prod_{d=1}^D I_d$, $J = \prod_{d=1}^D J_d$. 
The TR format is similar to TRM, but parameterizes the matrix into a more compact and low-rank structure. Specifically, it factorizes the matrix into $2D$ 3rd-order core tensors, denoted as $\tB^d \in \mathbb{R}^{I_d \times R \times R}$ and $\tC^d \in \mathbb{R}^{J_d \times R \times R}, \forall d = 1, \dots, D$. Each element is computed using the following contraction,
\begin{equation}\label{eq:ttv-def}
\begin{multlined}[0.8\linewidth]
\bm{\Delta}[\overline{i_1 \cdots i_D}, \overline{j_1 \cdots j_D}] = \\ \tr(\tB^1[i_1, :, :] \cdots \tB^D[i_D, :, :]\tC^1[j_1, :, :] \cdots \tC^D[j_D, :, :]).
\end{multlined}
\end{equation}
The space complexity is $\mathcal{O}(D I^{1/D} R^2)$, which is even smaller than that of TRM.

\vspace{-1em}
\paragraph{Initialization.}

Due to the high-order structure, the TR format may be more sensitive to initialization. Previous works adopting TR in PEFT \cite{yang2024loretta,abronin2024tqcompressor} utilize random Gaussian initialization for all factors $\tB^{1:D}$ and $\tC^{1:C}$, hence losing the zero-initialization of the overall adaptation as in LoRA.
This may cause optimization instability and result in the loss the information from pre-trained models.

In particular, we can write the TR layer as a sequence of linear layers.
Given two tensors $\tA$ of shape $I_d \times R \times R$ and $\tX$ of shape $I_1 \cdots \times I_d \times R$,
we define the $\times_2$ contraction as,
\[
\tA \times_2 \tX = \sum_{l=1}^{I_d} \sum_{r=1}^R \tA[l, :, r] \tX[I_1, \dots, l, r].
\]
The result is of shape $I_1 \times \cdots \times I_{d-1} \times R$.
Therefore, the TR residual layer can be defined as,
\[
\text{TR}(\tB, \tC) \vx = \tr( \tB^1 \cdots \times_2 \tB^D \times_2 \tC^1 \cdots \times_2 \tC^D \times_2 \tX ),
\]
which is reformulated as multiple linear layers.
To ensure zero initialization, we initialize $\tB^1$ as a zero tensor. Each element of other core tensors $\tB^{2:D}$ and $\tC^{1:D}$ is initialized independently from Gaussian distribution $\gN(0, \sigma^2)$.
The initialization strategy now basically follows the $\mu$P framework \cite{yang2023spectral}, i.e., the standard deviation should be $\sigma = \Theta(\sqrt{n\_out} / n\_in)$, where $n\_out = R$ and $n\_in = I_d R$.

\subsection{Connections with previous methods}\label{sec:connection}

In this subsection, we show that some previous methods implicitly or explicitly adopt the idea of a transform to seek a low-rank space. However, instead of using a learnable dense transform such as TRM, they rely on extremely sparse or fixed transforms, which may be insufficient.

\vspace{-1em}
\paragraph{DoRA.}

\citet{liu2024dora} propose to decompose the pre-trained weight $\mW_0$ into magnitude and direction, and fine-tune them separately,
\begin{equation*}
    \mW' = \frac{\mW_0 + {\color{cvprblue} \mB \mA}}{\lVert \mW_0 + \mB \mA \rVert_c} \cdot \text{diag}({\color{cvprblue} \vm}),
\end{equation*}
where the blue parts $\vm \in \mathbb{R}^{1 \times J}$, $\mB \in \mathbb{R}^{I \times R}$ and $\mA \in \mathbb{R}^{R \times J}$ are trainable parameters. $\lVert \cdot \rVert_c$ denotes the column-wise norm. For practical implementation, \citet{liu2024dora} suggests not updating the norm part and treating it as a constant. Therefore, we omit it and re-write the DoRA as,
\begin{equation}\label{eq:dora}
    \mW' \propto \mW_0 \text{diag}(\vm) + \mB \mA \text{diag}(\vm) \approxeq \mW_0 \text{diag}(\vm) + \mB \mA.
\end{equation}
Here we merge the last $\vm$ into $\mA$ since they are all trainable parameters.
\cref{eq:dora} resembles our model \cref{eq:proposed-model} by parameterizing the transform by a diagonal matrix $\text{diag}(\vm)$.
Compared to DoRA, TRM is a more flexible dense transform.

\vspace{-1em}
\paragraph{Methods with fixed transform.}

Some works assume that the fine-tuning process has better low-rank structures under particular \emph{fixed} transforms. Therefore, they propose to project the adaptation $\bm{\Delta}$ using some fixed transforms explicitly or implicitly, such as Fourier low-rank adaptation \cite{gao2024parameterefficient,borse2024foura} and low-displacement-rank adaptation \cite{sehanobish2024structured,chen2024quanta}.
To illustrate this, we take FouRA \cite{borse2024foura} as an example, which adopts the Discrete Fourier Transform (DFT) on $\bm{\Delta}$, i.e.,
$\bm{\Delta}_{\text{foura}} = \gF^{-1}( {\color{cvprblue} \mB}  \cdot \gF ({\color{cvprblue} \mA})),$
where $\gF$ is the DFT.
We omit the gating function in FouRA, which could be an orthogonal contribution to the DFT.
Empirically, \citet[][Fig. 4]{borse2024foura} shows $\bm{\Delta}_{\text{foura}}$ has smaller residual singular values than the original LoRA $\bm{\Delta}$, which may explain its better performances.
We can consider this as fine-tuning the pre-trained weight under the DFT,
\begin{equation*}
\vy =  \gF^{-1} (\gF(\mW_0) + {\color{cvprblue} \mB} \cdot \gF({\color{cvprblue} \mA})) \vx,
\end{equation*}
which is essentially learning the low-rank $\bm{\Delta}$ in the space of the transformed weight $\gF(\mW_0)$.
While this class of methods apply fixed transforms, our method adopts a learnable transform, which would adaptively learn this projection across different models and tasks.

%% file: contents/fig_and_tab/sdxl_inpaint_param.tex
\begin{figure*}[t]
    \centering
    \begin{subfigure}[b]{0.25\linewidth}
    \includegraphics[width=\linewidth]{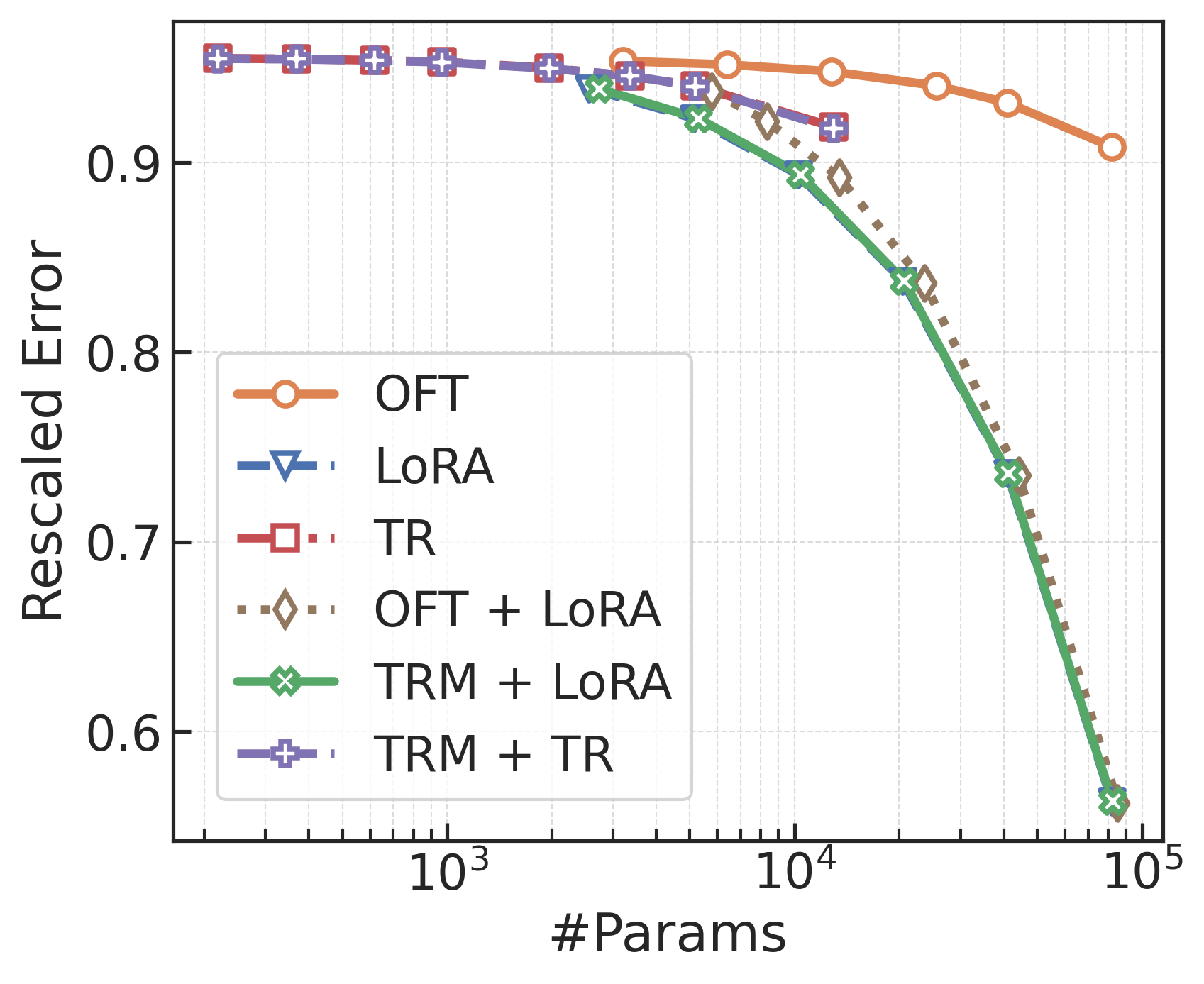}
    \caption{Additive low-rank difference}\label{fig:motivation-lora}
    \end{subfigure}
    \hspace{3em}
    \begin{subfigure}[b]{0.25\linewidth}
    \includegraphics[width=\linewidth]{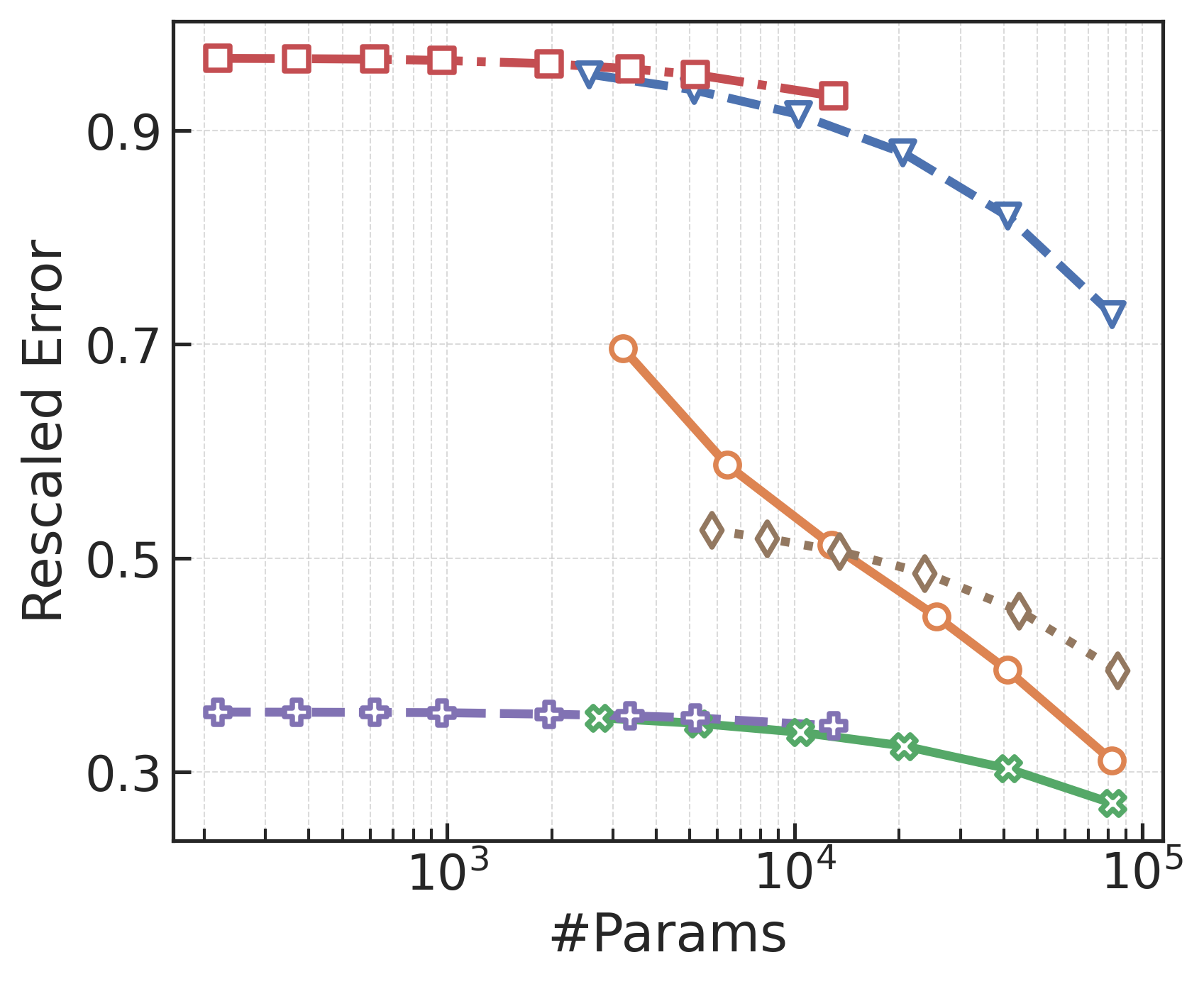}
    \caption{Orthogonal rotation}\label{fig:motivation-rotation}
    \end{subfigure}
    \hspace{3em}
    \begin{subfigure}[b]{0.25\linewidth}
    \includegraphics[width=\linewidth]{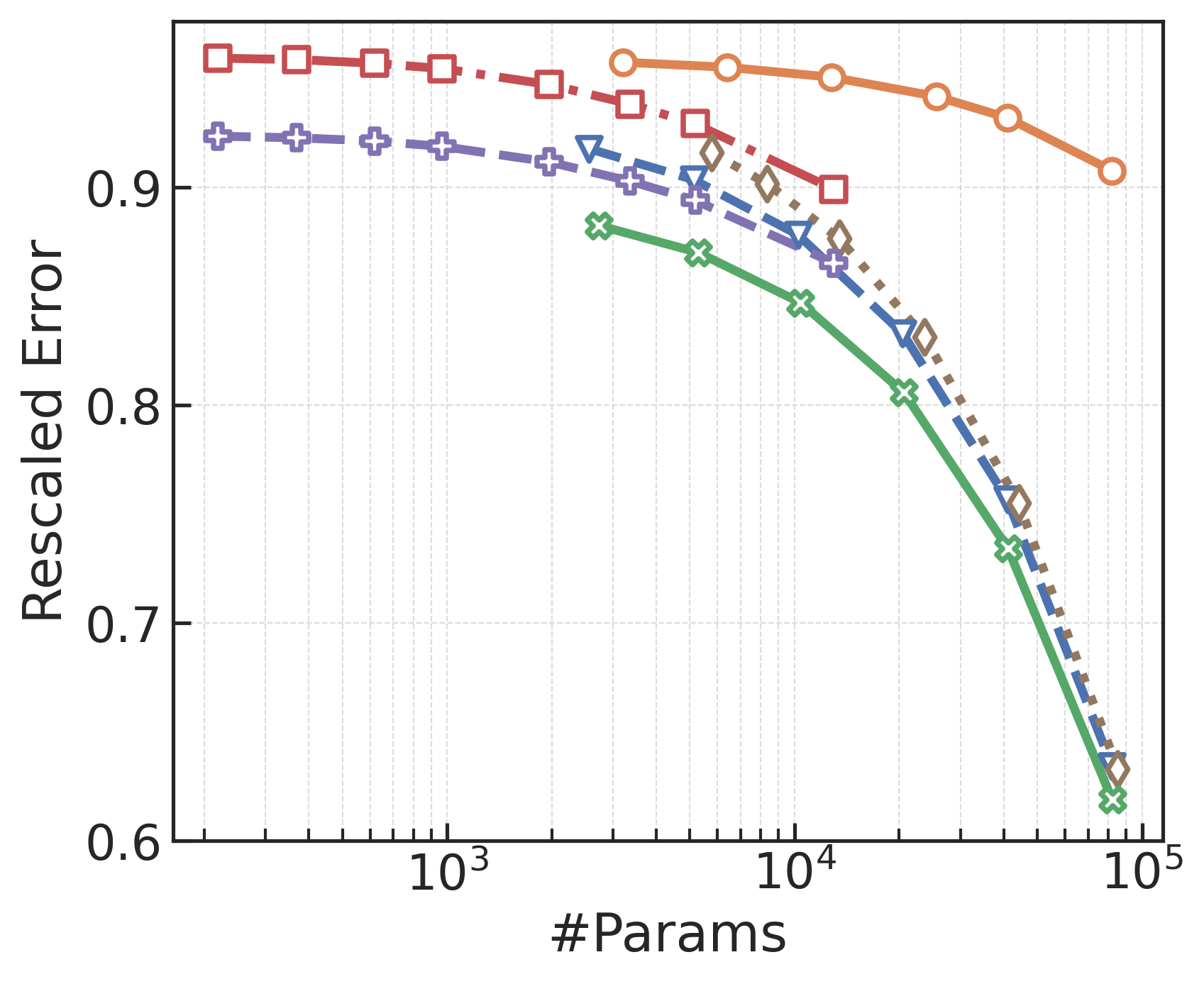}
    \caption{True fully fine-tuned weight}\label{fig:motivation-true}
    \end{subfigure}
    \caption{Simulation study on pre-trained SDXL \cite{podell2024sdxl} and fine-tuned SDXL-Inpaint \cite{SDXL-inpaint} weights. SDXL-Inpaint is fine-tuned on image-mask pairs to facilitate imputation ability of the SDXL base model. We investigate the approximation of UNet attention layers, which are counterpart of our fine-tuning targets in experiments.  We test the approximation error on three cases: (a) Additive low-rank difference, which LoRA can effectively compensate for the difference. (b) Orthogonal rotation, which follows the assumption of OFT and is full-rank. (c) True fully fine-tuned weights, which could be mixtures of additive and rotative effects.}\label{fig:motivatoin_param_sdxl_inpaint}
    \vspace{-1em}
\end{figure*}
\captionsetup[subfigure]{justification=centering,singlelinecheck=true}

%% file: contents/4_exp.tex
\input{contents/fig_and_tab/db_results}
\input{contents/fig_and_tab/db_param}

\input{contents/fig_and_tab/controlnet}

\section{Experiments}\label{sec:exp}

In this section, we present our experimental analysis. More details about settings and results are shown in \cref{app-sec:exp}. The code is available at \url{https://github.com/taozerui/tlora_diffusion}.

\subsection{Subject-driven generation}\label{sec:exp-db}

\paragraph{Experimental setting.}

We conduct subject-driven generation on the DreamBooth dataset \cite{ruiz2023dreambooth}.
We fine-tune the SDXL \cite{podell2024sdxl} model using the Direct Consistency Optimization \cite[DCO,][]{lee2024direct} algorithm. Following previous works \cite{ruiz2023dreambooth,lee2024direct}, we set the batch size to 1 and use the AdamW optimizer with constant learning rates.
For all methods, learning rates are tuned from \{\num{5e-4}, \num{1e-4}, \num{5e-5}\}.
We train the model for 20 epochs for each individual subject.

\vspace{-1em}
\paragraph{Baselines.}

We compare our method with LoRA \cite{hu2022lora}, DoRA \cite{liu2024dora}, OFT \cite{qiu2023controlling}, BOFT \cite{liu2024parameterefficient}, ETHER \cite{biniether} and LoRETTA \cite{yang2024loretta}.
For LoRA and DoRA, the rank is set to 1, to compare performances under extremely limited parameter conditions.
The block size is set to be 2 for both OFT and BOFT, while the number of Butterfly factors for BOFT is 2, denoted as $\text{OFT}_{2}, \text{BOFT}_{(2, 2)}$ respectively.
ETHER is a modification of OFT/BOFT by using Householder reflection for orthogonal matrices.
For ETHER, we choose the parameterization without orthogonal constraints, denoted as ETHER+, which is reported to have better performance \cite{biniether}. And the number of blocks is set to 1.
For LoRETTA, we choose a LoRA rank of 8 and a TT rank of 6.
For our method, we choose TRM transform adaptation with rank 1 and TR residual adaptation with rank 2.
The settings and numbers of trainable parameters are shown in \cref{tab:db_param}.
By choosing flexible TRM/TR forms, our method can achieve a much smaller size even compared to LoRA with rank 1.

\vspace{-1em}
\paragraph{Evaluation.}
The performance is evaluated in three aspects: subject alignment, prompt alignment and sample diversity, as in \cite{qiu2023controlling}.
Specifically, the subject alignment is measured by the image similarity between generated samples and training samples using CLIP-I \cite{radford2021learning} and DINOv2 \cite{oquab2024dinov2}.
The prompt alignment is measured by the image-text similarity between generated samples and given prompts using CLIP-T \cite{radford2021learning}.
Finally, the sampling diversity is measured by the distances between generated samples and training samples using LPIPS \cite{zhang2018unreasonable}.
Note that there are complex trade-offs among these metrics.
Typically, during the fine-tuning process, the subject alignment metrics would increase at the cost of losing prompt alignment and sample diversity, due to language shift \cite{ruiz2023dreambooth} and over-fitting.

To give a comprehensive visualization of the fine-tuning process, we evaluate the performances every 2 epochs and plot the Pareto curves of these metrics.
The numerical evaluation is shown in \cref{fig:db-exp}.
Our method achieves the best overall results with the smallest number of parameters.
In particular, comparing DoRA with LoRA, we find that adding the magnitude can improve the performance, which is similar to adding the transform part.
Moreover, our method further outperforms DoRA, which may be due to the usage of a more effective dense TRM transform. Also, compared to the tensor decomposition baseline LoRETTA, we can see the advantage of using the TRM transform. Interestingly, we find that tensor decomposition methods, including LoRETTA and our method, have better sampling diversity in \cref{fig:db-lpips-clipi,fig:db-lpips-dino}.

Qualitative results are illustrated in \cref{fig:db-images}.
We use the same random seeds to generate these images. We find that in many cases, LoRA and DoRA tend to produce similar images.
Images from our model sometimes look similar to those from BOFT. However, our model generally has better subject- and text-alignments.

\begin{figure*}[t]
    \centering
    \includegraphics[width=.95\linewidth]{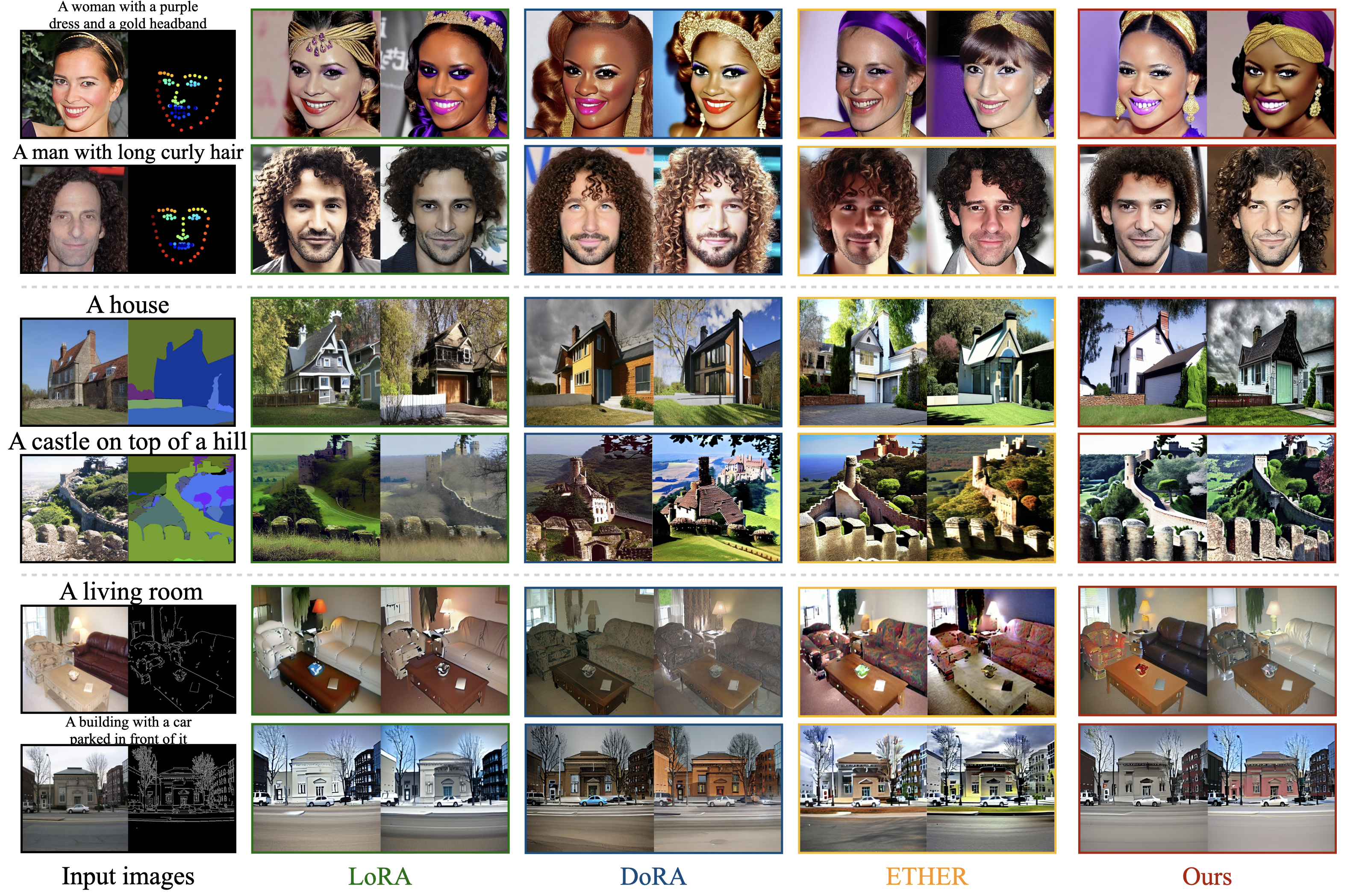}
    \caption{Qualitative results of controllable generation.}\label{fig:control-images}
    \vspace{-1em}
\end{figure*}

\subsection{Controllable generation}

\paragraph{Experimental setting.}

For controllable generation, we follow the setting in \cite{zhang2023adding,qiu2023controlling}.
Specifically, we fine-tune the Stable Diffusion \cite[SD,][]{rombach2022high} v1.5 model with datasets of images and conditional signals.
Three tasks are evaluated: (1) Landmark to Image (L2I) on the CelebA-HQ \cite{xia2021tedigan} dataset, (2) Segmentation to Image (S2I) on the ADE20K \cite{zhou2017scene} dataset, and (3) Canny to Image (C2I) on the ADE20K dataset. The baselines are the same as in \cref{sec:exp-db}, with different hyper-parameter settings to adjust the trainable parameters.
For our model, we adopt four settings. First, to show the effect of adding the TRM transform adaptation, we apply TRM with rank 2 on top of LoRA with rank 2 and 4, denoted as TLoRA\textsuperscript{*}\textsubscript{(2, 2)} and TLoRA\textsuperscript{*}\textsubscript{(2, 4)}. Then, we also test TRM with rank 2 and TR residual adaptation with ranks 6 and 8 to achieve a smaller fine-tuning size, denoted as TLoRA\textsubscript{(2, 6)} and TLoRA\textsubscript{(2, 8)} respectively.

\vspace{-1em}
\paragraph{Results.}

For the L2I task, we measure the MSE between the detected face landmark and the ground truth.
For the S2I task, we adopt the pre-trained SegFormer B4 model \cite{xie2021segformer} provided in HuggingFace for segmentation of generated images.
Then, the mean Intersection over Union (mIoU), all ACC (aACC) and mean ACC (mACC) metrics are reported.
For C2I, we evaluate the IoU and F1 score of the Canny edges of generated images.
For each test sample, we generate six images to report the mean and standard deviation.

The numerical results are shown in \cref{tab:control}.
Unlike previous observations in \cite{qiu2023controlling,liu2024parameterefficient}, we find LoRA is actually a strong baseline for these tasks.
Nonetheless, our model achieves the best or comparable results.
For the L2I task, we find LoRA achieves the best performance. DoRA is inferior to LoRA for this task, which may indicate that the transform part is not important here. However, our model can still achieve a comparable performance.
For S2I and C2I tasks, our method by adding the transform significantly improves performances.
Unlike in \cref{sec:exp-db}, since here we use larger ranks, the TR residual does not have advantages in L2I and S2I tasks, which is consistent with our observations in \cref{sec:motivation}.
Moreover, we find LoRETTA is very unstable for these datasets.
It either diverges or generates unrealistic images.
This may be because of its non-zero initialization, which destroys the information from the pre-trained model.
As a comparison, our method, which also adopts TR residual adaptation, works well for these datasets.

\cref{fig:control-images} showcases test examples and generated images.
Our model also achieves good signal alignment, prompt alignment and image quality.
For instance, in the L2I task, our model not only generates images aligned with the face landmark, but also preserves good prompt alignment.
In the S2I task, LoRA generates images with lower signal alignment, which is consistent with the numerical results.
In the C2I task, the image quality of LoRA is not good, although it achieves high accuracy on control signals.

%% file: contents/fig_and_tab/db_results.tex
\begin{figure}[t]
    \centering
    \begin{subfigure}[b]{0.48\linewidth}
    \includegraphics[width=\linewidth]{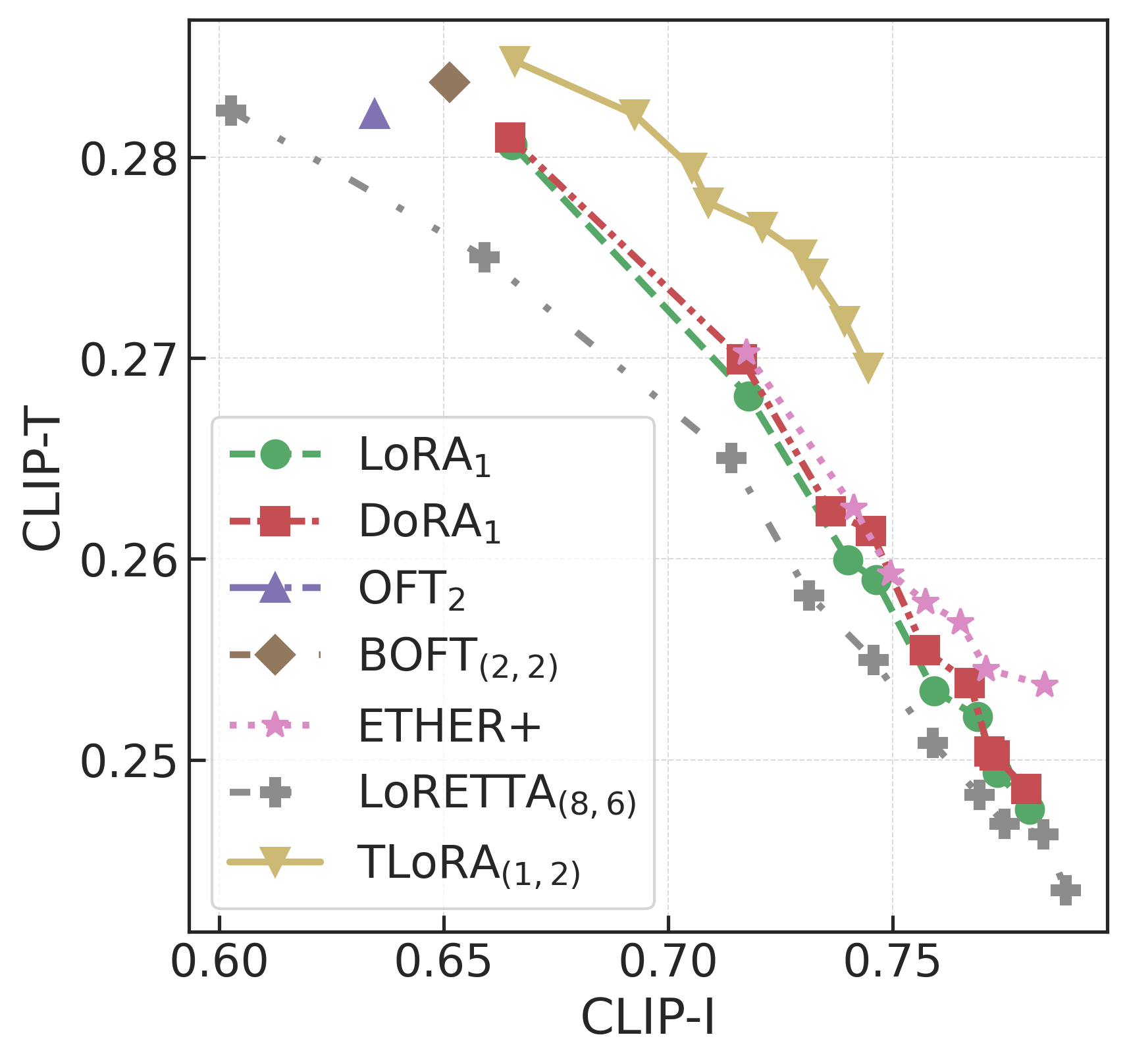}
    \caption{CLIP-T \& CLIP-I}
    \end{subfigure}
    \hfill
    \begin{subfigure}[b]{0.48\linewidth}
    \includegraphics[width=\linewidth]{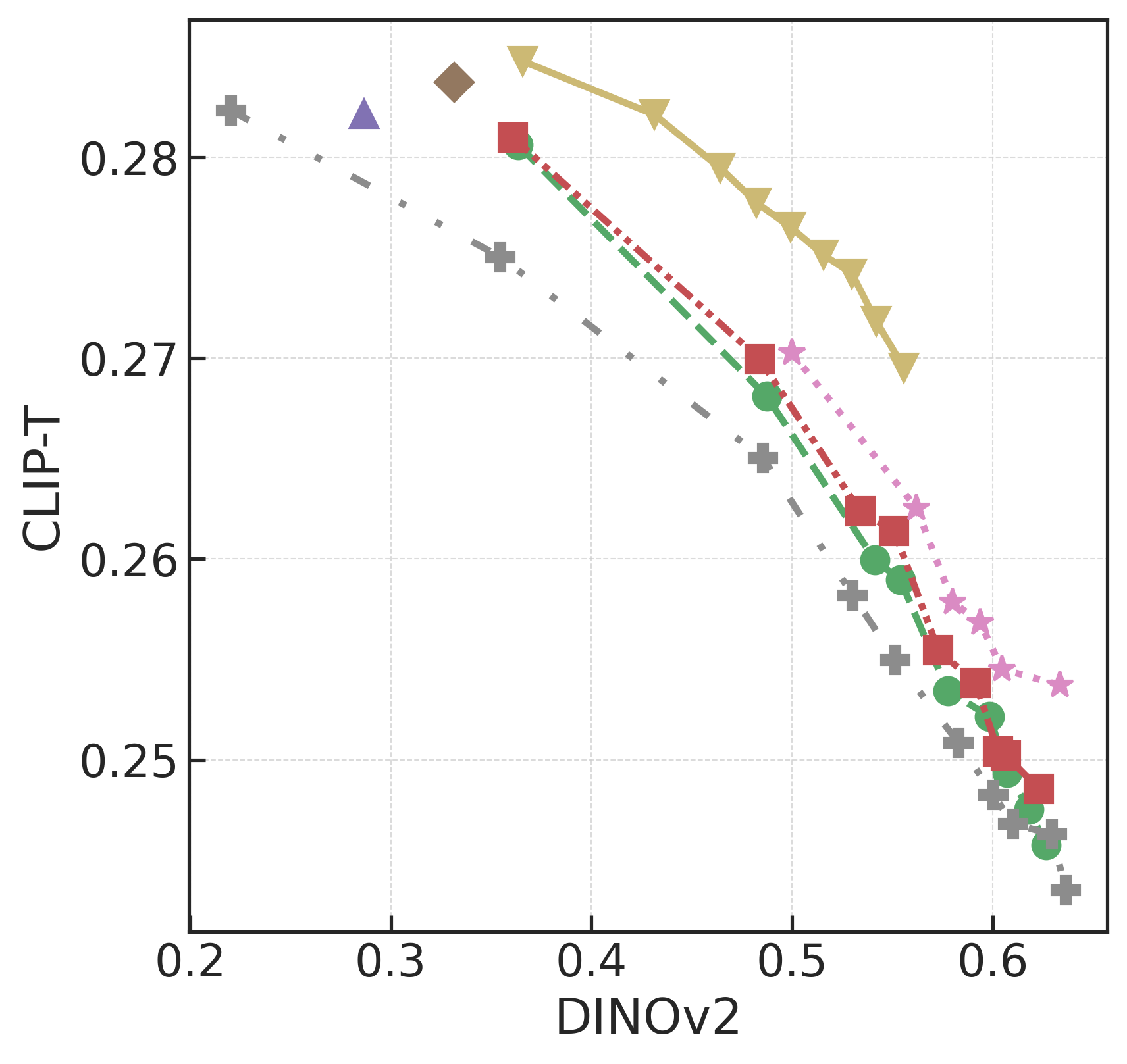}
    \caption{CLIP-T \& DINOv2}
    \end{subfigure}
    
    \begin{subfigure}[b]{0.48\linewidth}
    \includegraphics[width=\linewidth]{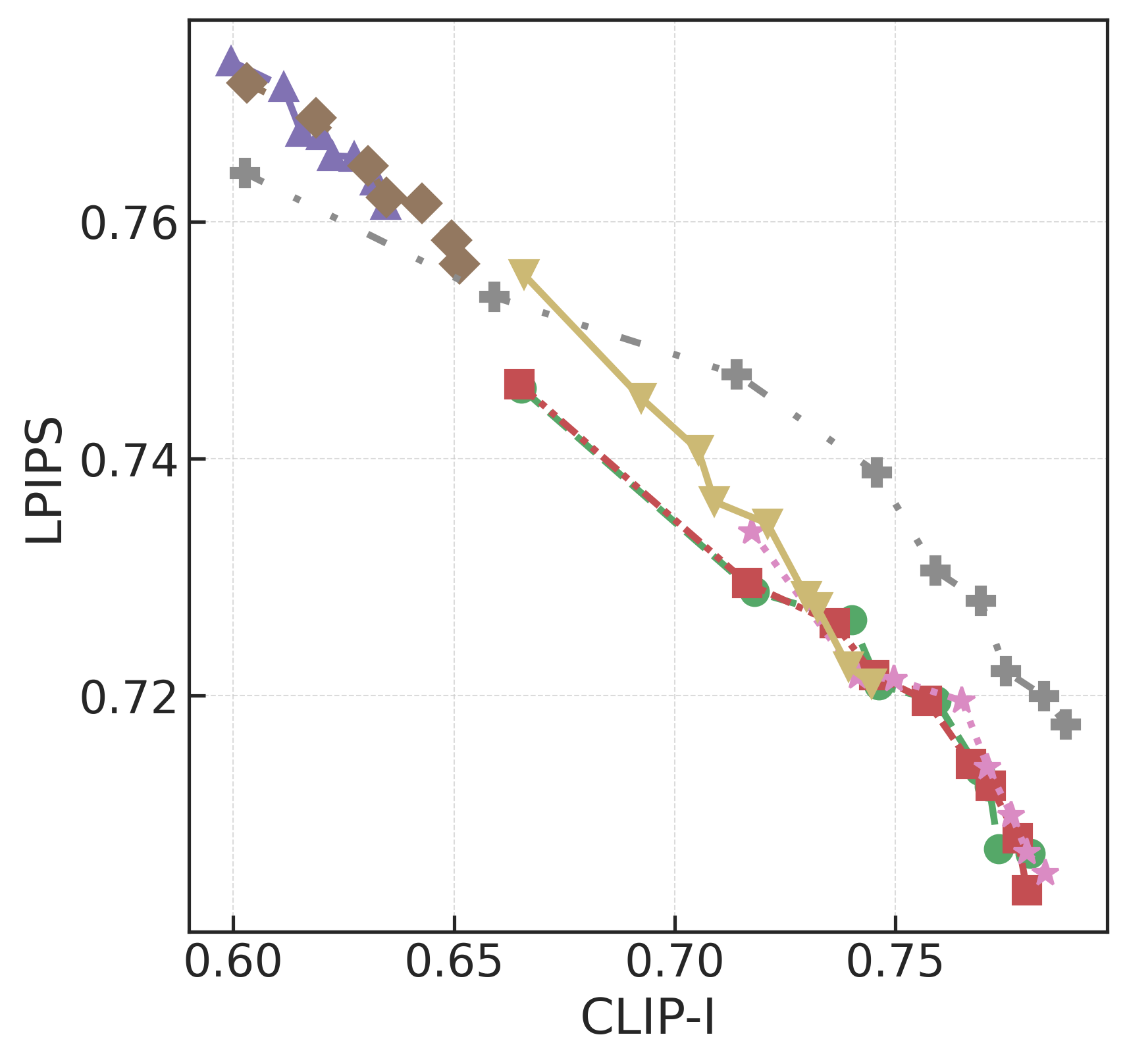}
    \caption{LPIPS \& CLIP-I}\label{fig:db-lpips-clipi}
    \end{subfigure}
    \hfill
    \begin{subfigure}[b]{0.48\linewidth}
    \includegraphics[width=\linewidth]{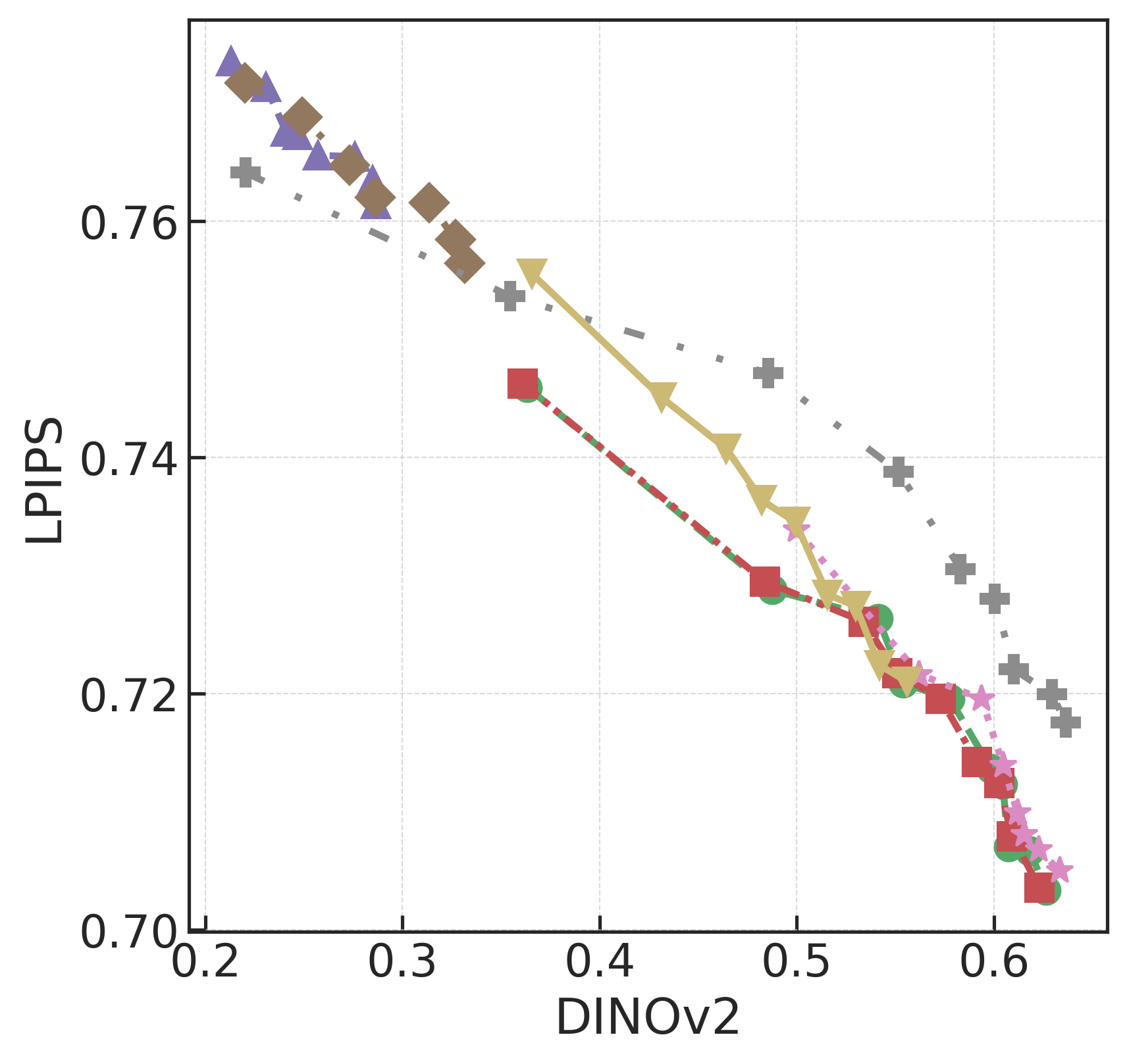}
    \caption{LPIPS \& DINOv2}\label{fig:db-lpips-dino}
    \end{subfigure}
    \caption{Subject-driven generation results. The larger DINOv2 and CLIP-I values indicate better subject alignment, larger CLIP-T values indicate better prompt alignment and larger LPIPS values indicate larger sample diversity.}\label{fig:db-exp}
    
    \vspace{-1em}
\end{figure}

%% file: contents/fig_and_tab/db_param.tex
\begin{table}[b]
\scriptsize
\setlength{\tabcolsep}{3pt}
\centering
  \caption{Number of trainable parameters.}\label{tab:db_param}
    \begin{tabular}{lccccccc}
    \toprule
    Method & LoRA & DoRA & OFT & BOFT & ETHER+ & LoRETTA & TLoRA \\
    \midrule
    Setting & $r$=1 & $r$=1 & $b$=2 & ($m$=2,$b$=2) & $n$=1 & (8, 6) & (1, 2) \\
    \#Param (M) & 1.45 & 2.12 & 2.24 & 3.81 & 1.57 & 0.99 & \bftab{0.40} \\
    \bottomrule
    \bottomrule
    \end{tabular}
\vspace{-2em}
\end{table}

%% file: contents/fig_and_tab/controlnet.tex
\begin{table*}[t]
\small
\centering
  \caption{Controllable generation results. We report the mean value of six samples, with the standard deviation in subscripts. Best results are highlighted in bold font. \textsuperscript{$\S$}Results are taken from the OFT \cite[][Table 2]{qiu2023controlling} paper. \textsuperscript{$\dag$}Results are taken from the BOFT \cite[][Table 6]{liu2024parameterefficient} paper.}\label{tab:control}
    \begin{tabular}{lccccccc}
    \toprule
    \multirow{2}*{Method} & \multirow{2}*{\#Param (M)} & L2I & \multicolumn{3}{c}{S2I} & \multicolumn{2}{c}{C2I} \\
     \cmidrule(lr){3-3} \cmidrule(lr){4-6} \cmidrule(lr){7-8}
     &  & Error$\downarrow$ & mIoU$\uparrow$ & aACC$\uparrow$ & mACC$\uparrow$ & IoU$\uparrow$ & F1$\uparrow$ \\
    \midrule
    SD\textsuperscript{$\S$} & 0 & 146.19\textsubscript{NA} & 7.72\textsubscript{NA} & 33.61\textsubscript{NA} & 14.40\textsubscript{NA} & NA & NA\\
    ControlNet\textsuperscript{$\S$} & 361.30 & 7.61\textsubscript{NA} & 20.88\textsubscript{NA} & 61.42\textsubscript{NA} & 35.52\textsubscript{NA} & NA & NA \\
    LoRA\textsubscript{$r$=4} & 0.80 & {\bftab{5.32}}\textsubscript{0.03} & 27.72\textsubscript{0.17} & 64.99\textsubscript{0.26} & 40.22\textsubscript{0.30} & 0.180\textsubscript{\num{3e-4}} & 0.305\textsubscript{\num{5e-4}} \\
    LoRETTA\textsubscript{(32, 8)} & 0.41 & 118.14\textsubscript{0.55} & 11.16\textsubscript{0.36} & 40.46\textsubscript{0.11} & 20.27\textsubscript{0.65} & 0.105\textsubscript{\num{2e-4}} & 0.189\textsubscript{\num{3e-4}} \\
    DoRA\textsubscript{$r$=4} & 0.90 & 6.43\textsubscript{0.11} & 27.11\textsubscript{0.27} & 65.70\textsubscript{0.37} & 39.16\textsubscript{0.48} & 0.143\textsubscript{\num{4e-4}} & 0.248\textsubscript{\num{7e-4}} \\
    ETHER+ & 0.40 & 6.40\textsubscript{0.08} & 27.66\textsubscript{0.29} & 65.76\textsubscript{0.13} & 40.37\textsubscript{0.38} & 0.189\textsubscript{\num{3e-4}} & 0.317\textsubscript{\num{4e-4}} \\
    OFT\textsubscript{$r$=32} & 1.50 & 7.35\textsubscript{0.07} & 28.52\textsubscript{0.46} & 66.04\textsubscript{0.29} & 41.27\textsubscript{0.54} & 0.165\textsubscript{\num{5e-4}} & 0.281\textsubscript{\num{7e-4}} \\
    OFT\textsubscript{$r$=4}\textsuperscript{$\S$} & 16.3 & 7.07\textsubscript{NA} & 27.06\textsubscript{NA} & 62.42\textsubscript{NA} & 40.09\textsubscript{NA} & NA & NA \\
    BOFT\textsubscript{($m$=2, $r$=32)} & 2.41 & 7.64\textsubscript{0.13} & 28.45\textsubscript{0.67} & 66.19\textsubscript{0.47} & 41.44\textsubscript{1.01} & 0.161\textsubscript{\num{4e-4}} & 0.276\textsubscript{\num{6e-4}} \\
    BOFT\textsubscript{($m$=4, $r$=8)}\textsuperscript{$\dag$} & 20.76 & 5.67\textsubscript{NA} & 28.83\textsubscript{NA} & 67.74\textsubscript{NA} & 41.24\textsubscript{NA} & NA & NA \\
    \midrule
    TLoRA\textsuperscript{*}\textsubscript{(2, 4)} & 0.94 & {\bftab{5.32}}\textsubscript{0.10} & {\bftab{29.23}}\textsubscript{0.43} & {\bftab{69.21}}\textsubscript{0.17} & {\bftab{41.88}}\textsubscript{0.84} & {\bftab{0.203}}\textsubscript{\num{3e-4}} & {\bftab{0.337}}\textsubscript{\num{5e-4}} \\
    TLoRA\textsuperscript{*}\textsubscript{(2, 2)} & 0.54 & 5.57\textsubscript{0.09} & 28.29\textsubscript{0.47} & 67.09\textsubscript{0.37} & 40.64\textsubscript{0.52} & 0.160\textsubscript{\num{2e-4}} & 0.274\textsubscript{\num{4e-4}} \\
    TLoRA\textsubscript{(2, 8)} & 0.60 & 5.78\textsubscript{0.09} & 27.42\textsubscript{0.38} & 66.95\textsubscript{0.21} & 38.99\textsubscript{0.49} & 0.178\textsubscript{\num{3e-4}} & 0.301\textsubscript{\num{5e-4}} \\
    TLoRA\textsubscript{(2, 6)} & 0.40 & 5.84\textsubscript{0.09} & 27.03\textsubscript{0.28} & 65.15\textsubscript{0.31} & 38.76\textsubscript{0.33} & 0.184\textsubscript{\num{4e-4}} & 0.310\textsubscript{\num{6e-4}} \\
    \bottomrule
    \bottomrule
    \end{tabular}
    \vspace{-1em}
\end{table*}


%% file: contents/5_conclusion.tex
\section{Conclusion and discussion}

In this work, we investigate the PEFT problem for text-to-image (T2I) models.
Our objective is to address the issues of LoRA regarding its low-rank assumption and inflexible parameter budget choices.
Our proposed PEFT method incorporates two adaptation parts, the transform and residual adaptations.
We adopt efficient and flexible tensor decomposition (TD) structures for these parts to meet different requirements.
Empirically, this combination of transforms and residuals shows advantages in approximation and fine-tuning T2I models for different tasks.

While we focus on T2I models, the proposed method is general for PEFT of other tasks. Moreover, TD is particularly suitable for convolutional layers, whose weights are naturally multi-way arrays.
In the future, we plan to test our method on other foundation models, including large-language models and vision transformers.

\section*{Acknowledgements}

We would like to thank Akio Hayakawa for his constructive feedback on an earlier version of this manuscript. We also thank the AC and anonymous reviewers for their valuable suggestions and comments. Qibin Zhao was supported by JSPS KAKENHI Grant Number JP23K28109.

%% file: contents/appendix.tex
\clearpage
\setcounter{page}{1}
\maketitlesupplementary
\addtocontents{toc}{\protect\setcounter{tocdepth}{2}}
\appendix
\tableofcontents



\section{Details and more results in \cref{sec:motivation}}\label{app-sec:motivation}

In this section, we present the detailed settings for experiments of approximating fine-tuned weights in \cref{sec:motivation}. Moreover, we provide results of more layers in the SDXL(-Inpaint) models, and models from other domains, e.g., large language models.

First, we present the experimental settings.
For results in \cref{fig:motivatoin_param_sdxl_inpaint}, the pre-trained weight $\mW_0$ is taken from the pre-trained SDXL model and the key for the layer is \texttt{mid\_block.attentions.0.transformer\_blocks\linebreak.0.attn1.to\_k}. The shape of $\mW_0$ is $1280 \times 1280$.
The detailed settings for three cases in \cref{fig:motivatoin_param_sdxl_inpaint} are as follows.
\begin{enumerate}
    \item \textbf{Additive low-rank difference.} The target weight is computed by $\mW_* = \mW_0 + \sigma \mB_* \mA_*$, where the rank of $\mB_* \mA_*$ is 128. Each element of $\mB_*$ and $\mA_*$ is drawn i.i.d. from $\gN(0, 1)$. After $\mB_*$ and $\mA_*$ are generated, we compute $\sigma$ to scale this difference part so that $\mW_* - \mW_0$ has the same standard deviation with the case where $\mW_*$ is the true fully fine-tuned weight from SDXL-Inpaint.
    \item \textbf{Orthogonal rotation.} The target weight is computed by $\mW_* = \mW_0 \mT_*$, where $\mT_*$ is a random orthogonal matrix. To generate this orthogonal matrix, we firstly generate a random matrix $\mX = \mI_{1280} + \mF$, where each element of $\mF$ is drawn i.i.d. from $\gN(0, 0.05^2)$. Then, $\mT_*$ is computed by the QR decomposition of $\mX$.
    \item \textbf{True fully fine-tuned weight.} The target weight $\mW_*$ is taken from the SDXL-Inpaint model \cite{SDXL-inpaint} in the same layer with $\mW_0$.
\end{enumerate}

All approximation methods are optimized by minimizing the Mean Squared Error (MSE) with Adam optimizer using learning rate $\num{1e-3}$ for 500 epochs, which is sufficient for them to converge. The settings of these methods are as follows. 
\begin{enumerate}[label=\Roman*.]
    \item \textbf{OFT.} We take the block diagonal structure and Cayley transform as in \cite{qiu2023controlling}. We vary the block size to get different parameter sizes.
    \item \textbf{LoRA.} We vary the rank to get different parameter sizes.
    \item \textbf{TR.} We use the TR form defined in \cref{eq:ttv-def} to replace the low-rank matrix decomposition in \cref{eq:lora-original}. The size of the pre-trained weight is $1280 \times 1280$. We reshape it into $1280 = 8 \times 8 \times 4 \times 5$ and choose different TR ranks to get different parameter sizes.
    \item \textbf{OFT + LoRA.} For OFT, we set the block size to 5, and combine it with LoRA of different ranks.
    \item \textbf{TRM + LoRA.} We choose TRM with rank 1, and combine it with LoRA of different ranks.
    \item \textbf{TRM + TR.} We choose TRM with rank 1, and combine it with TR of different ranks.
\end{enumerate}

Then, we provide results of different layers in the SDXL(-Inpaints) models in \cref{fig:sdxl-weight-app}.
To investigate the potential application of our method in large language models,
we also conduct similar investigations in Llama2 7B and Llama2-chat 7B models \cite{touvron2023llama}, as shown in \cref{fig:llama-weight}.
The size of SDXL weights is $1280 \times 1280$, while the size of the Llama2 7B weights is $4096 \times 4096$.
For both models, we have similar observations as in \cref{sec:motivation}.
Moreover, in the SDXL model, we find the effect of adding the transform is more significant for \emph{key} and \emph{query} weights.
It would be interesting to analyze this effect more theoretically in the future.
Besides, compared to SDXL, the TR structure appears to be more effective in the Llama2 models.
This may be because that the tensor decomposition is more suitable for compression of weight matrices with larger sizes.
A future direction would be exploration of our method in fine-tuning large language models.

\input{contents/fig_and_tab/sdxl_weights_app}

\input{contents/fig_and_tab/llama_weights}

\section{Proposed model}\label{app-sec:model}

\subsection{Expressiveness of TRM}

In this subsection, we present details of the simulation study in \cref{fig:boft-vs-ttm}, \cref{sec:ttm}.
To empirically compare TRM with Butterfly matrices in BOFT, we randomly generate orthogonal matrices and evaluate the approximation error. In specific, we generate a random matrix $\mX$ of shape $512 \times 512$, where each element is i.i.d. from unit Normal distribution $\gN(0, 1)$. Then, the target orthogonal matrix $\mT_*$ is obtained from the QR decomposition of $\mX$. We approximate $\mT_*$ using BOFT and TRM by minimizing the MSE with Adam optimizer.

For BOFT, we test 2, 4, and 16 Butterfly factors, as in \cite{liu2024parameterefficient}. For each setting, we test the number of diagonal blocks ranging from 1 to 9, 1 to 8, and 1 to 6, denoted as BOFT(1:9, 2), BOFT(1:8, 4), and BOFT(1:6, 16) respectively. For TRM, we test different tensor shape, namely, $4 \times 4 \times 4 \times 8$, $8 \times 8 \times 8$, and $16 \times 32$. Then, we set different TR ranks to obtain results of different parameter sizes. In \cref{fig:boft-vs-ttm}, we find the results of TRM are robust to the choices of tensor sizes, and are consistently better than BOFT. Besides, even there is no orthogonal constraints for TRM, it can approximate orthogonal matrices well.

\subsection{Proof of \cref{prop:init,prop:orthogonal}}

In this subsection, we provide the derivation of \cref{prop:init,prop:orthogonal}.

\paragraph{Proof of \cref{prop:init}.}

According to the definition of TRM in \cref{eq:ttm-def}, we have
\begin{equation*}
\begin{multlined}[0.9\linewidth]
    \mT[\overline{i_1 \cdots i_D}, \overline{j_1 \cdots j_D}] = \\ \tr(\tA^1[i_1, j_1, :, :] \cdots \tA^D[i_D, j_D, :, :]).
\end{multlined}
\end{equation*}

Following the initialization in \cref{prop:init}, for $d = 1, \dots, D$, if $i_d = j_d$, all the elements of $\tA^d[i_d, j_d, :, :]$ equal to $1 / R$, else zero. Then we can discuss the diagonal and non-diagonal elements of $\mT$ separately.

\begin{enumerate}
\item For non-diagonal elements, i.e., $\overline{i_1 \cdots i_D} \neq \overline{j_1 \cdots j_D}$, there is at least one sub-index $i_d \neq j_d$. Therefore, $\mT[\overline{i_1 \cdots i_D}, \overline{j_1 \cdots j_D}] = 0$, since $\tA^d[i_d, j_d, :, :] = \vzero$ if $i_d \neq j_d$.

\item For diagonal elements, i.e., $\overline{i_1 \cdots i_D} = \overline{j_1 \cdots j_D}$, we have $i_d = j_d, \forall d = 1, \dots, D$. Now the core tensors become $\tA^d[i_d, j_d, :, :] = \vone_{R \times R} / R, \forall d = 1, \dots, D$ and $i_d, j_d = 1, \dots, I_d$, where $\vone_{R \times R}$ is a matrix of shape $R \times R$ with all elements being one.
Therefore, $\mT[\overline{i_1 \cdots i_D}, \overline{j_1 \cdots j_D}] = 1$, since
\[
\tr \left( \frac{\vone_{R \times R}}{R} \cdots \frac{\vone_{R \times R}}{R} \right) = 1.
\]
\end{enumerate}

Therefore, $\mT$ is an identity matrix.

\paragraph{Proof of \cref{prop:orthogonal}.}

For simplicity, in this proof, we denote $\tA^d[i_d, j_d] = \tA^d[i_d, j_d, :, :]$ and $\tB^d[i_d, j_d] = \tB^d[i_d, j_d, :, :]$.
Suppose $\mX$ has shape $I \times K$ with $I = \prod_{d=1}^D I_d$ and $K = \prod_{d=1}^D K_d$, and $\mY$ has shape $J \times K$ with $J = \prod_{d=1}^D J_d$.
According to the definition of TRM in \cref{eq:ttm-def}, we can compute the product $\mZ = \mX \mY^\intercal$ as follows,
\begin{equation*}
\begin{aligned}
&\mZ[\overline{i_1 \cdots i_D}, \overline{j_1 \cdots j_D}] \\
& = \sum^{K_1, \dots, K_D}_{k_1, \dots, k_D=1} \mX[\overline{i_1 \cdots i_D}, \overline{k_1 \cdots k_D}] \mY[\overline{j_1 \cdots j_D}, \overline{k_1 \cdots k_D}] \\
& \quad = \begin{multlined}[t][0.8\linewidth]
    \sum^{K_1, \dots, K_D}_{k_1, \dots, k_D=1} \tr(\tA^1[i_1, k_1] \cdots \tA^D[i_D, k_D]) \\
        \cdot \tr(\tB^1[j_1, k_1] \cdots \tB^D[j_D, k_D])
    \end{multlined} \\
& \quad = \begin{multlined}[t][0.8\linewidth]
    \sum^{K_1, \dots, K_D}_{k_1, \dots, k_D=1} \tr\left\{ (\tA^1[i_1, k_1] \cdots \tA^D[i_D, k_D]) \right. \\
        \left. \otimes (\tB^1[j_1, k_1] \cdots \tB^D[j_D, k_D]) \right\}
    \end{multlined} \\
& \quad = \begin{multlined}[t][0.8\linewidth]
    \sum^{K_1, \dots, K_D}_{k_1, \dots, k_D=1} \tr\left\{ (\tA^1[i_1, k_1] \otimes \tB^1[j_1, k_1]) \right. \\
        \left. \cdots (\tA^D[i_D, k_D] \otimes \tB^D[j_D, k_D]) \right\}
    \end{multlined} \\
& \quad = \begin{multlined}[t][0.8\linewidth]
    \tr \Biggl\{ \sum^{K_1, \dots, K_D}_{k_1, \dots, k_D=1} \Bigl[ (\tA^1[i_1, k_1] \otimes \tB^1[j_1, k_1]) \\
        \cdots (\tA^D[i_D, k_D] \otimes \tB^D[j_D, k_D]) \Bigr] \Biggr\}
    \end{multlined} \\
& \quad = \begin{multlined}[t][0.8\linewidth]
        \tr \Biggl\{ \sum^{K_1}_{k_1=1} (\tA^1[i_1, k_1] \otimes \tB^1[j_1, k_1]) \\
        \cdots \sum^{K_D}_{k_D} (\tA^D[i_D, k_D] \otimes \tB^D[j_D, k_D]) \Biggr\},
    \end{multlined}
\end{aligned}
\end{equation*}
which follows a TR format.
Therefore $\mX \mY^\intercal = \text{TRM}(\tC^{1:D})$, where each core tensor $\tC^d[i_d, j_d, :, :] = \sum_{l_d} (\tA^d[i_d, l_d, :, :] \otimes \tB^d[j_d, l_d, :, :])$.

To make $\mX$ orthogonal, i.e., $\mX \mX^\intercal = \mI$, we can regularize $\tC^{1:D}$ according to the initialization scheme in \cref{prop:init}.

\section{Experiments}\label{app-sec:exp}

In this section, we provide experimental details and more results.
All the experiments are conducted on single Nvidia H100 or A100 GPU with 80GB memory.
The code is available at \url{https://github.com/taozerui/tlora_diffusion}.

\subsection{Experimental details}

\paragraph{Datasets.}

For the subject-driven generation, we use the DreamBooth dataset \cite{ruiz2023dreambooth}, which includes 30 subjects from 15 different classes.
For text prompts, we follow the setting in \citet{lee2024direct,zhang2024spectrum}. For each subject, there are several images with 10 testing text prompts. The dataset is available at the Github repository\footnote{\url{https://github.com/phymhan/SODA-Diffusion}} of \citet{zhang2024spectrum}.

For the controllable generation task, we consider three tasks and two datasets.
The settings basically follow \citet{qiu2023controlling}.
For the Landmark to Image (L2I) task, we use the CelebA-HQ \cite{xia2021tedigan} dataset. The whole dataset consists of $30k$ images of faces, captions generated by BLIP \cite{li2022blip}, and face landmarks detected by the \texttt{face-alignment} library \cite{bulat2017far}. The test set contains 2987 samples.
For the Segmentation to Image (S2I) task, we use the ADE20K 2017 dataset \cite{zhou2017scene}, which consists of $20k$ training images, segmentations and captions generated by BLIP. The test set contains 2000 samples.
For the Canny to Image (C2I) task, we also use the ADE20K dataset, where the canny edges are detected using the same detector as in \citet{zhang2023adding}.

\paragraph{Baselines.}

We choose the following baselines, which are highly related to our work and competitive methods.
\begin{itemize}
    \item LoRA \cite{hu2022lora}, which is the original method.
    \item DoRA \cite{liu2024dora}, which is a popular extension of LoRA. Moreover, it shares some similarities with our method, as discussed in \cref{sec:connection}.
    \item OFT \cite{qiu2023controlling}, which applies orthogonal transforms for fine-tuning. It uses block diagonal transforms for parameter efficiency.
    \item BOFT \cite{liu2024parameterefficient}, which extends OFT by using Butterfly matrices to construct dense transforms.
    \item ETHER \cite{biniether}, which adopts Householder reflection to parameterize orthogonal transforms. In particular, we adopt the implementation called ETHER+, which relaxes the orthogonal constraint and applies transforms on the left and right sides of the pre-trained weight. It is more flexible and has shown better results in \citet{biniether}.
    \item LoRETTA \cite{yang2024loretta}, which is an extension of LoRA. It further factorized LoRA matrices using TT decomposition for parameter efficiency.
\end{itemize}

LoRA, DoRA, OFT, and BOFT are provided in the \texttt{PEFT} library \cite{peft}.
For ETHER\footnote{\url{https://github.com/mwbini/ether}} and LoRETTA\footnote{\url{https://github.com/yifanycc/loretta}}, we apply their official implementations in our experiments.

\input{contents/fig_and_tab/db_hyper}

\paragraph{General settings.}

For all the baselines and our model, we inject the trainable parameters into the attention modules on \emph{key}, \emph{value}, \emph{query}, and \emph{output} layers. This setting is consistent with previous works \cite{qiu2023controlling,liu2024parameterefficient,biniether} for a fair comparison.

For the subject-driven generation, we fine-tune the SDXL \cite{podell2024sdxl} model using the Direct Consistency Optimization \cite[DCO,][]{lee2024direct} algorithm. Following previous works \cite{ruiz2023dreambooth,lee2024direct}, we set the batch size to 1 and use the AdamW optimizer with constant learning rates.
For all methods, learning rates are tuned from \{\num{5e-4}, \num{1e-4}, \num{5e-5}\}.
We train the model for 20 epochs for each individual subject.

For the controllable generation, we follow the implementation of ControlNet \cite{zhang2023adding}. It contains a shallow 8-layer CNN to encode the control signals. For a fair comparison and being consistent with previous works \cite{qiu2023controlling,liu2024parameterefficient,biniether}, we report the number of training parameters for adaptation parts of all methods. The optimizer is AdamW. For the CelebA-HQ dataset in the L2I task, the batch size is 16 and we fine-tune the model for 22 epochs. For the ADE20K dataset in the S2I and C2I tasks, the batch size is 8 and we fine-tune the model for 20 epochs.
For all methods, learning rates are tuned from \{\num{1e-3}, \num{5e-4}, \num{1e-4}, \num{5e-5}\}.
We do some preliminary learning rate search on the L2I and S2I tasks, and find the optimal learning rate of each method is similar for these two tasks. Due to the computational cost, we use the same learning rate for each method on three tasks.
Moreover, as the training of LoRETTA is unstable, we additionally adopt a smaller learning rate \num{1e-5} for it.

\paragraph{Hyper-parameters.}

The hyper-parameters for the subject-driven generation is provided in \cref{tab:db-hyper}.

The hyper-parameters for the controllable generation is provided in \cref{tab:control-hyper}. For this experiment, we find adding the identity regularization $\gR_I$ sometimes helps the performance. We apply a scale $\lambda$ before adding the regularization on the original loss. The scale $\lambda$ is chosen from $\{0, \num{1e-3} \}$. Moreover, for tensorization of large matrices, we choose the following setting, where the keys are original dimensions and values are dimensions of sub-indices. We do not test other tensorization shapes.

\begin{lstlisting}[language=Python]
TENSOR_SHAPE_DICT = {
    '320': [4, 8, 10], '640': [8, 8, 10],
    '768': [8, 8, 12], '1280': [8, 10, 16],
    '2048': [8, 16, 16], '2560': [10, 16, 16],
    '5120': [20, 16, 16], '10240': [32, 20, 16],
}
\end{lstlisting}

\paragraph{Evaluation.}

The evaluation of the subject-driven generation is described in \cref{sec:exp-db}.

For the L2I task, we use the same face landmark detector provided in the \texttt{face-alignment} library \cite{bulat2017far} to detect landmarks from generated images. The Mean Squared Error (MSE) between ground truth and detected landmarks is reported. For the S2I task, we adopt the SegFormer B4 model \cite{xie2021segformer} pre-trained on the ADE20K dataset for segmentation of generated images. The model is downloaded from HuggingFace\footnote{\url{https://huggingface.co/nvidia/segformer-b4-finetuned-ade-512-512}}.
Then, the mean Intersection over Union (mIoU), all ACC (aACC) and mean ACC (mACC) metrics are reported.
For C2I, we compute the canny edges of generate images using the same detector as in \cite{zhang2023adding}. Then,
we evaluate the IoU and F1 score of the Canny edges of generated images.
For each test sample, we generate six images to report the mean and standard deviation.

\subsection{Computational cost}

Our method is computationally efficient, since the tensor decomposition structure  consists of several small linear layers \cref{eq:ttm-def,eq:ttv-def}.
The main computational overhead comes from multiplying pretrained weights with transform matrices, which is inherent to related methods such as ETHER, OFT and BOFT. Training times measured on an NVIDIA A100 GPU for controllable generation tasks demonstrate competitive computational efficiency, as shown in \cref{app-tab:time}.
\begin{table}[t]
\vspace{-1em}
\caption{Computing time.}\label{app-tab:time}
    \centering
    \setlength{\tabcolsep}{3pt}
    \resizebox{0.99\linewidth}{!}{
    \begin{tabular}{lccccccc}
    \toprule
    \textbf{} & LoRA & DoRA & ETHER+ & BOFT & OFT & TLoRA$^*$(2,4) & TLoRA(2,8) \\
    \midrule
    Iter./Sec.$\uparrow$      & 0.58 & 0.55 & 0.30 & 0.35 & 0.34 & 0.37 & 0.34 \\
    \bottomrule
    \end{tabular}
    }
\end{table}

\begin{figure*}[h]
    \centering
    \includegraphics[width=0.9\linewidth]{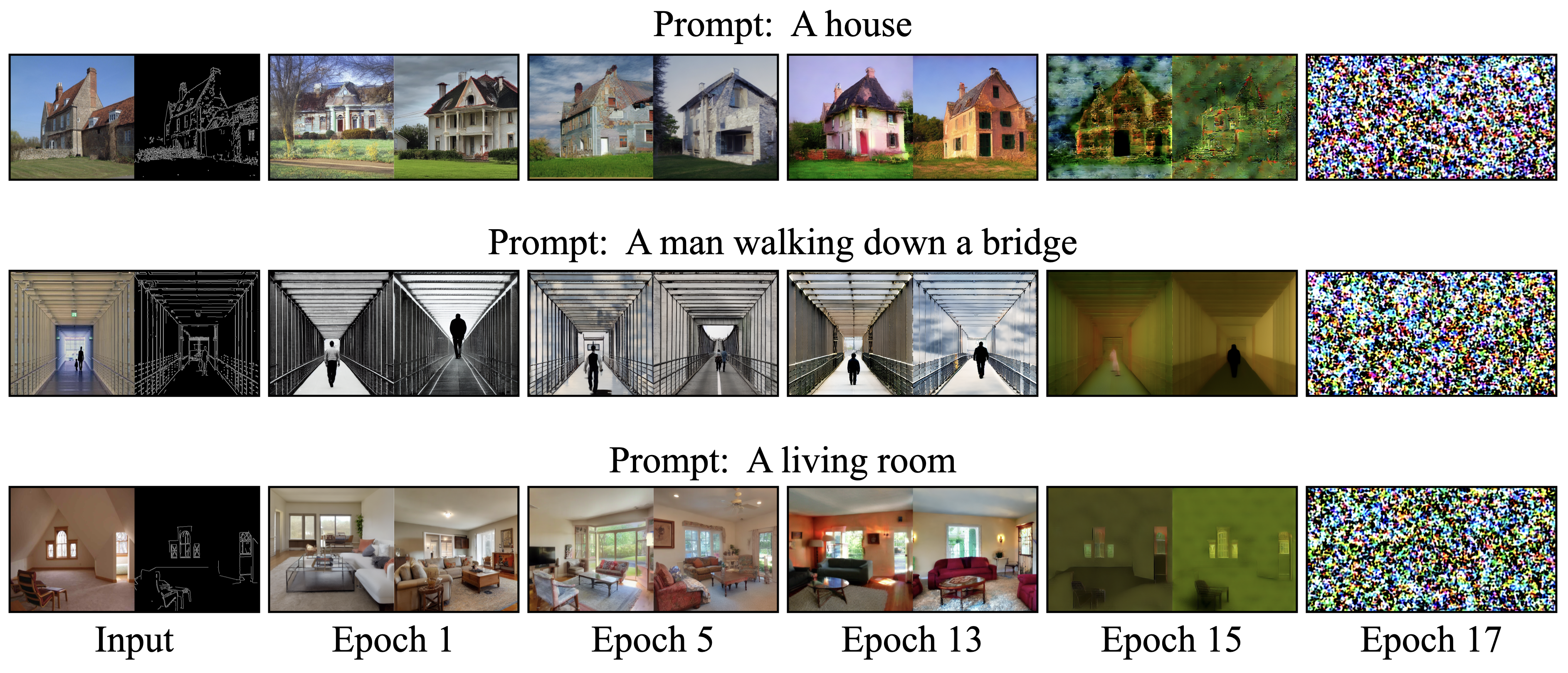}
    \caption{Training process of LoRETTA on the C2I task.}
    \label{fig:loretta-epochs}
\end{figure*}

\subsection{Failure of LoRETTA}

When applying the LoRETTA \cite{yang2024loretta} on controllable generation tasks, we find the training is unstable. It either generates unrealistic images or diverges. We test learning rates from \{\num{1e-3}, \num{5e-4}, \num{1e-4}, \num{5e-5}, \num{1e-5}\}. For large learning rates, it diverges quickly, while for small learning rates, it does not learn the control signals well and the image quality is low.
To illustrate this, we showcase the training process of LoRETTA on the C2I task in \cref{fig:loretta-epochs}.
This may be because of its non-zero initialization, which destroys the information from the pre-trained model.
As a comparison, our method, which also adopts TR residual adaptation, works well for these datasets.

\subsection{Additional visualization results}

\begin{figure*}[t]
    \centering
    \includegraphics[width=0.95\linewidth]{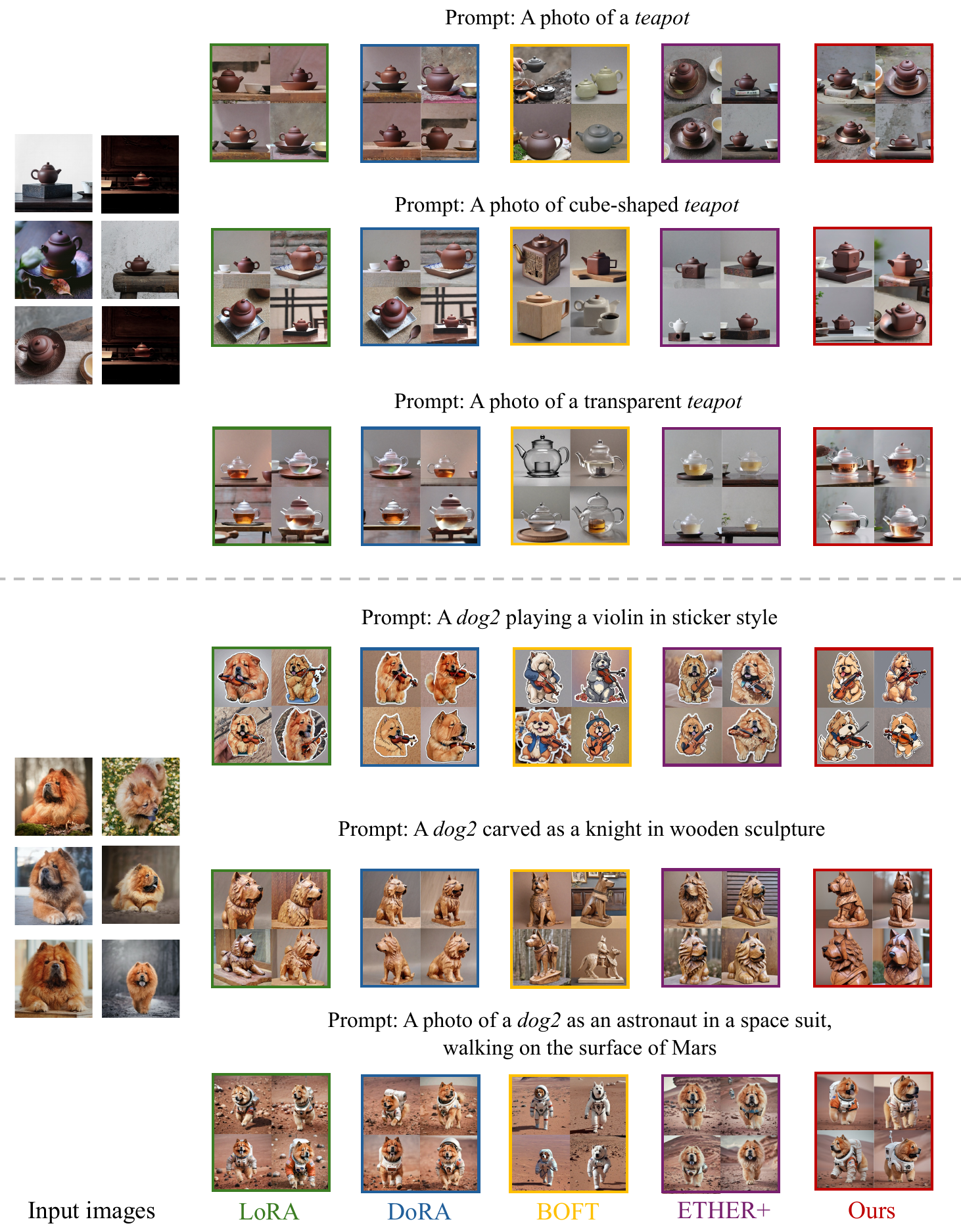}
    \caption{Qualitative comparison of the subject-driven generation results.}\label{fig:appendix-db1}
\end{figure*}

\begin{figure*}[t]
    \centering
    \includegraphics[width=0.95\linewidth]{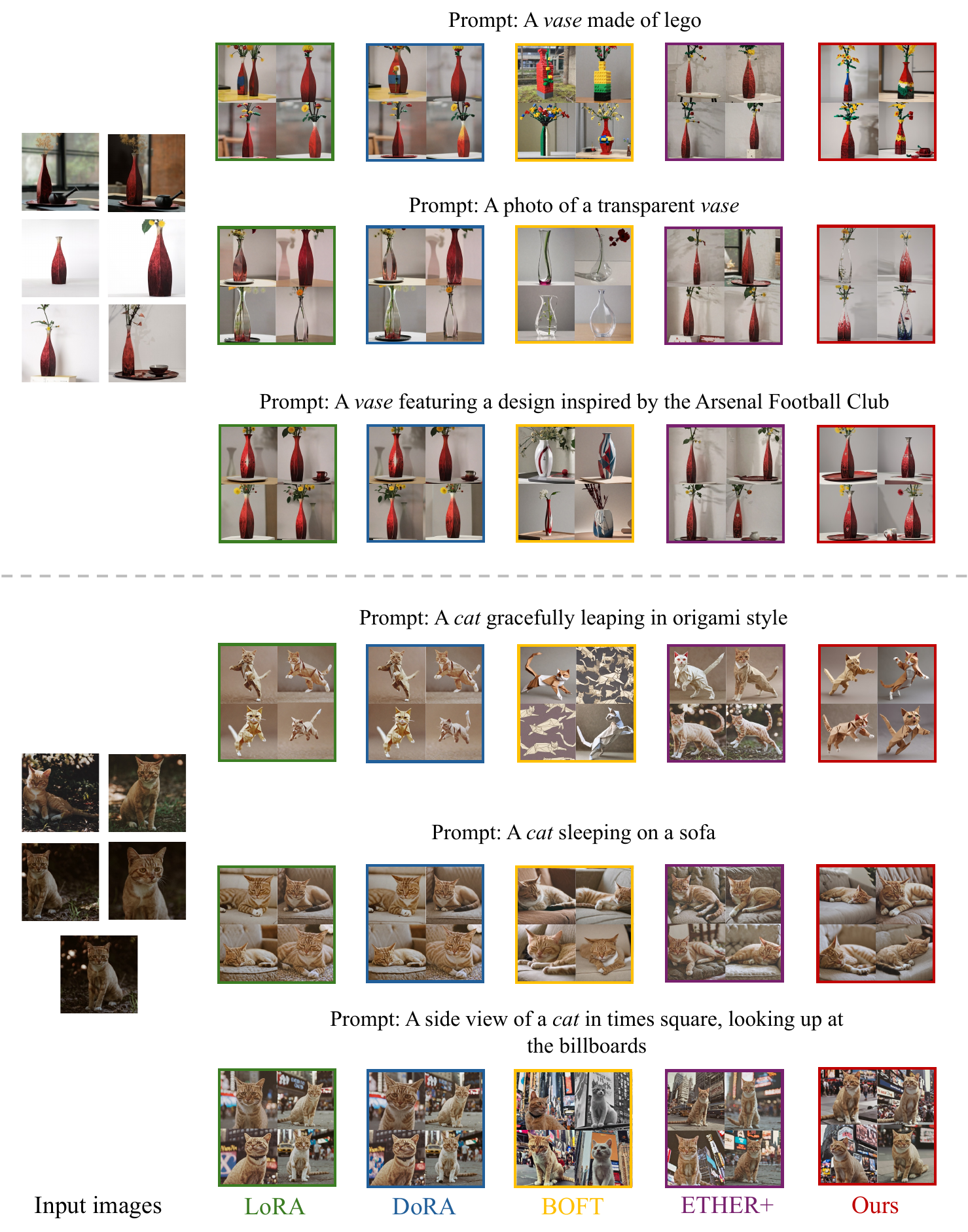}
    \caption{Qualitative comparison of the subject-driven generation results.}\label{fig:appendix-db2}
\end{figure*}

\begin{figure*}[t]
    \centering
    \includegraphics[width=0.92\linewidth]{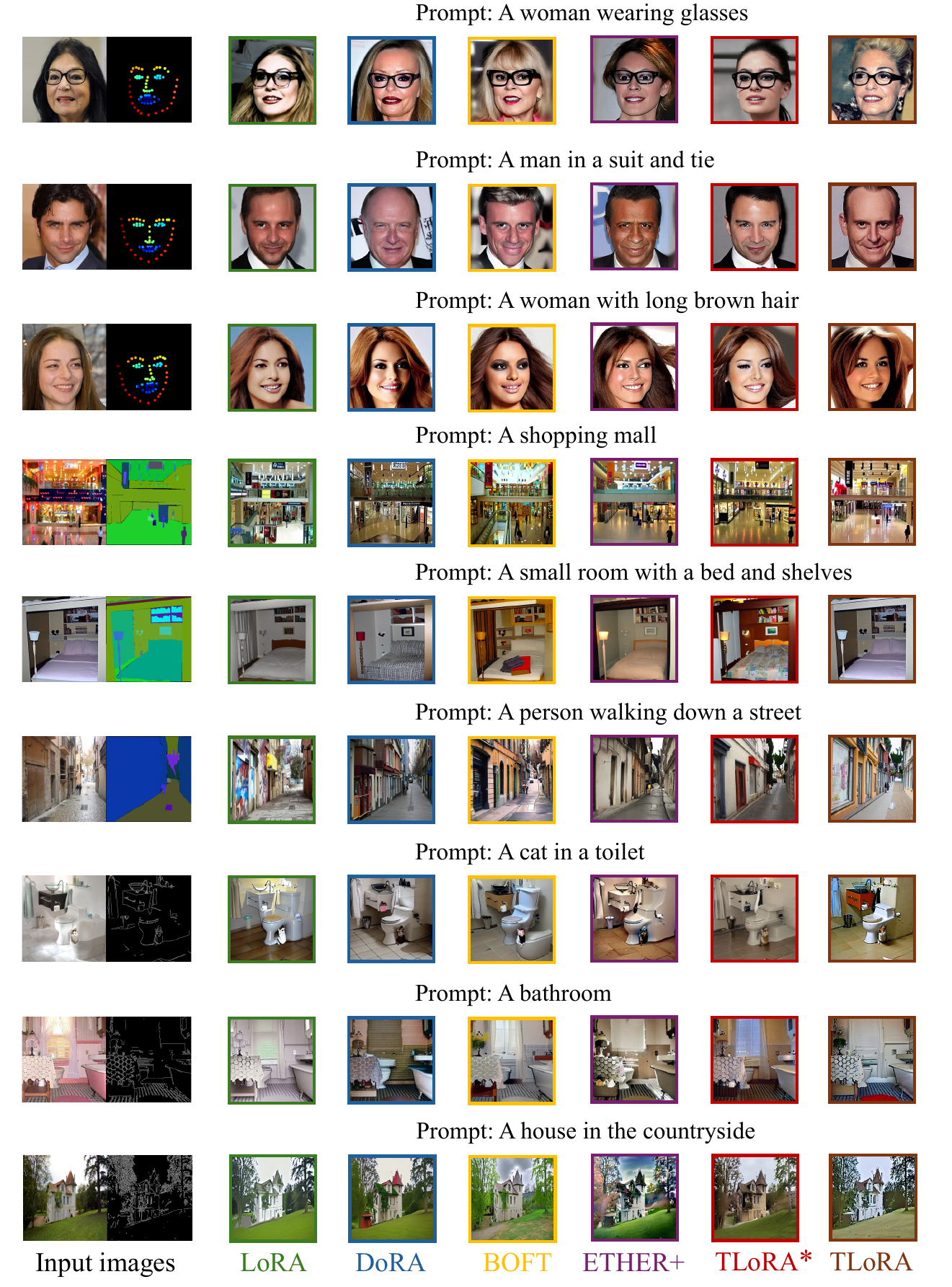}
    \caption{Qualitative comparison of the subject-driven generation results.}\label{fig:appendix-control}
\end{figure*}

We present more visualization results in \cref{fig:appendix-db1,fig:appendix-db2,fig:appendix-control}.

\section{Related work}\label{app-sec:related}

Due to the space limit, we present a more comprehensive review of related work here.

\paragraph{Text-to-image model personalization.}

Text-to-image generative models have shown exceptional results in image synthesis \cite{ramesh2022hierarchical,rombach2022high,saharia2022photorealistic,podell2024sdxl,esser2024scaling}.
Given the various pre-trained models available, many works aim to fine-tune these models for personalized datasets or tasks, such as subject-driven generation and controllable generation.
\citet{gal2023an} propose learning given subjects via textual inversion, while \citet{ruiz2023dreambooth} fine-tune the whole model.
ControlNet \cite{zhang2023adding} incorporates an additional network branch, which can learn datasets of paired control signals and images.
While \citet{ruiz2023dreambooth,zhang2023adding} have large numbers of trainable parameters, \citet{kumari2023multi} show that fine-tuning the attention layers alone is also effective for these tasks.
More recently, many works have focused on developing PEFT methods for these attention layers and have shown promising results \cite{han2023svdiff,qiu2023controlling,gu2024mix,yeh2024navigating,zhang2024spectrum,borse2024foura,xie2023difffit,NEURIPS2024_18c0102c,hu2025sara,chen2025para,zhuang2024time}. In particular, \citet{zhuang2024time} propose time-varying adapters that match the denoising process of diffusion models; this can be an interesting direction to further improve our method.
There are also training-free approaches \cite{rout2024rb}, which could be slow at inference.

\vspace{-1em}
\paragraph{Parameter-efficient fine-tuning.}

PEFT has become a hot topic with the emergence of large foundation models including text-to-image models and large language models.
Popular PEFT methods include Adapter \cite{houlsby2019parameter}, Prefix-tuning \cite{li2021prefix}, Prompt-tuning \cite{lester2021power}, low-rank adaptation \cite[LoRA,][]{hu2022lora}, and many of their variants.
Adapter adds additional layers after pretrained feed forward layers.
Prompt-tuning introduces learnable prompts for specific tasks.
LoRA has become the most popular PEFT method due to its simplicity and impressive performance \cite{fomenko2024note}.
Many variants of LoRA have been proposed
\cite{zhang2023adaptive,liu2024dora,qiu2023controlling,liu2024parameterefficient,nikdan2024rosa,kopiczkovera,hyeon-woo2022fedpara,NEURIPS2024_2c570b0f,NEURIPS2024_68716328,huang2025hira}. In particular, DoRA \cite{liu2024dora} proposes decomposing the pre-trained weight into magnitude and direction, where vanilla LoRA is applied to the direction. In this work, we show that DoRA can also be connected to our work by using a diagonal transform. OFT \cite{qiu2023controlling} applies a learnable orthogonal transform for adaptation. However, for parameter efficiency, OFT adopts block diagonal matrices, which are highly sparse. Subsequently, many methods aim to improve OFT by applying more efficient dense transform structures, including Butterfly matrix \cite{liu2024parameterefficient}, Given rotation \cite{maparameter}, Householder reflection \cite{biniether,yuan2024bridging} and Kronecker product \cite{zhang2024spectrum}. Our method adopts a similar idea of using a transform, but we design a different efficient dense matrix parameterization using tensor decomposition.
Some methods also share similarities with ours by using \emph{pre-defined and fixed} transforms on the pre-trained weights to project onto some low-rank spaces \cite{gao2024parameterefficient,borse2024foura,sehanobish2024structured}.
There are also works aim to design memory-efficient \emph{full-parameter} fine-tuning/pre-training optimizers \cite{zhao2024galore,NEURIPS2024_2c570b0f,NEURIPS2024_68716328}. However, they do not provide light-weight adapters that can be plugged into foundation models.

\vspace{-1em}
\paragraph{Tensor decomposition.}

TD is a classical tool in signal processing and machine learning \cite{cichocki2016tensor}.
In particular, tensor-train (TT) decomposition \cite{oseledets2011tensor} and its extension, tensor-ring (TR) decomposition \cite{zhao2016tensor}, have shown exceptional results in model compression,
including MLP \cite{novikov2015tensorizing}, CNN \cite{wang2018wide,garipov2016ultimate}, RNN/LSTM \cite{pan2019compressing,su2020convolutional,yang2017tensor,mehta2019scaling} and Transformer \cite{ma2019tensorized,pham2022tt}.
Recently, TDs have also been applied to fine-tuning tasks.
\citet{jie2023fact} parameterize the Adapter layers using a TT format and show ultra-parameter-efficiency in ViT adaptation. \citet{yang2024loretta} extend this idea to large language model PEFT, and apply the TT format to both Adapters and LoRA factors. Similarly, \citet{anjum2024tensor} propose to directly parameterize the adaptation using the TT format.
\citet{chen2024quanta} adopt a quantum-inspired tensor network, which is a generalization of the TT-Matrix (TTM) form.
While these works use similar TT/TR structures to our model, they do not apply the transform adaptation.
Moreover, we study a different initialization strategy for the TR factors, which would be more stable, as we show in our experiments.

%% file: contents/fig_and_tab/sdxl_weights_app.tex
\begin{figure*}[t]
    \centering
    \begin{subfigure}[b]{0.23\linewidth}
    \includegraphics[width=\linewidth]{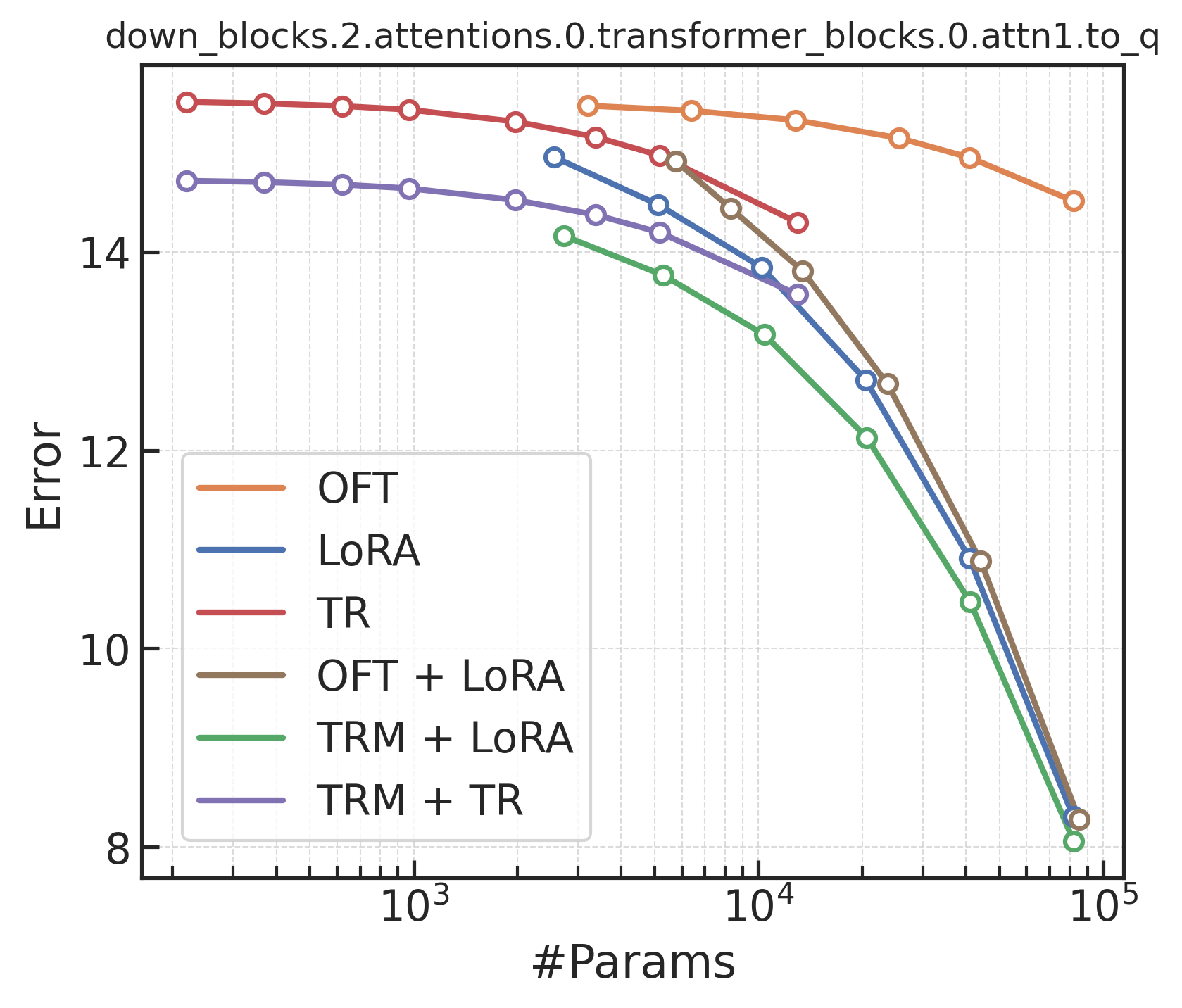}
    \end{subfigure}
    \hfill
    \begin{subfigure}[b]{0.23\linewidth}
    \includegraphics[width=\linewidth]{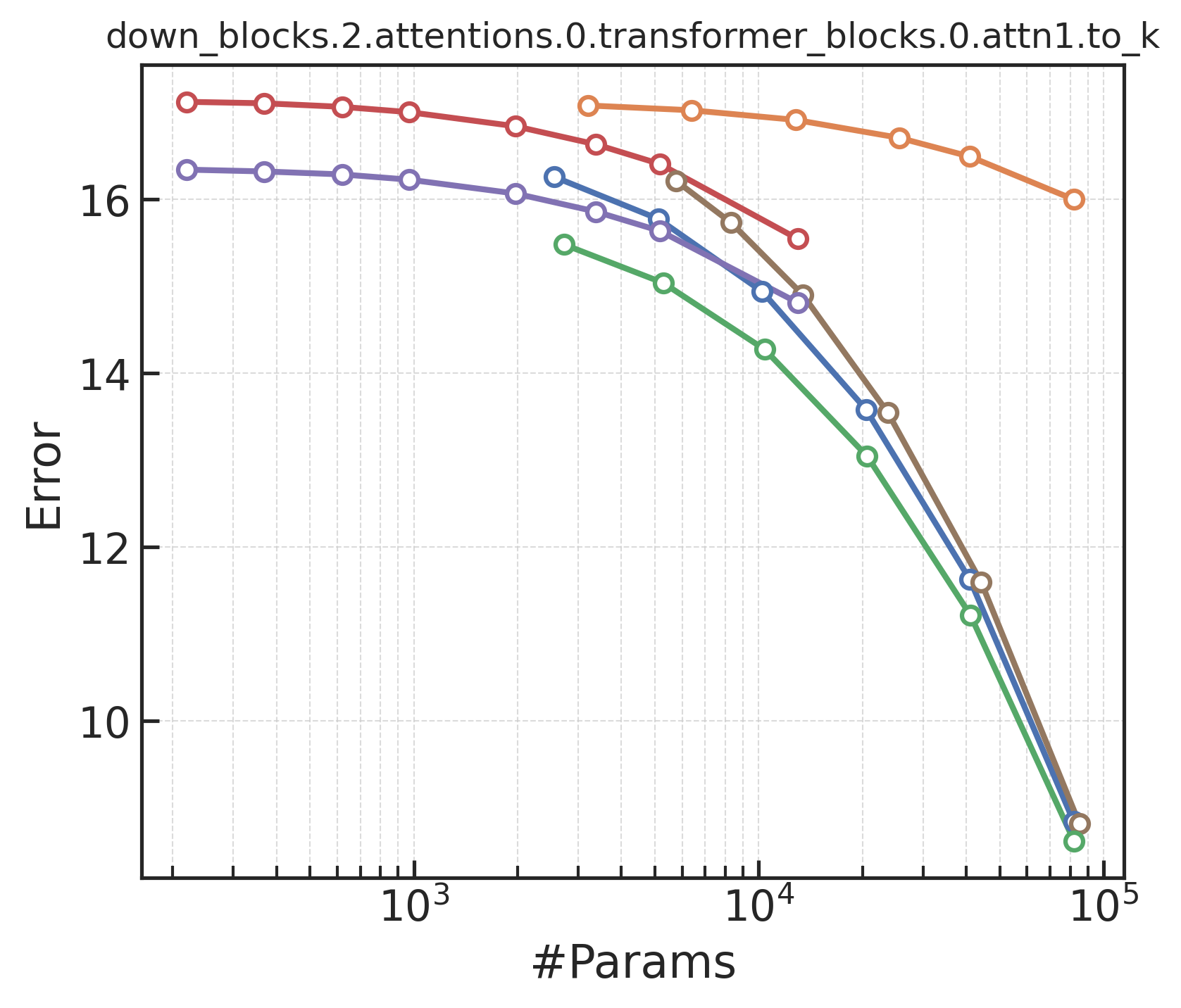}
    \end{subfigure}
    \hfill
    \begin{subfigure}[b]{0.23\linewidth}
    \includegraphics[width=\linewidth]{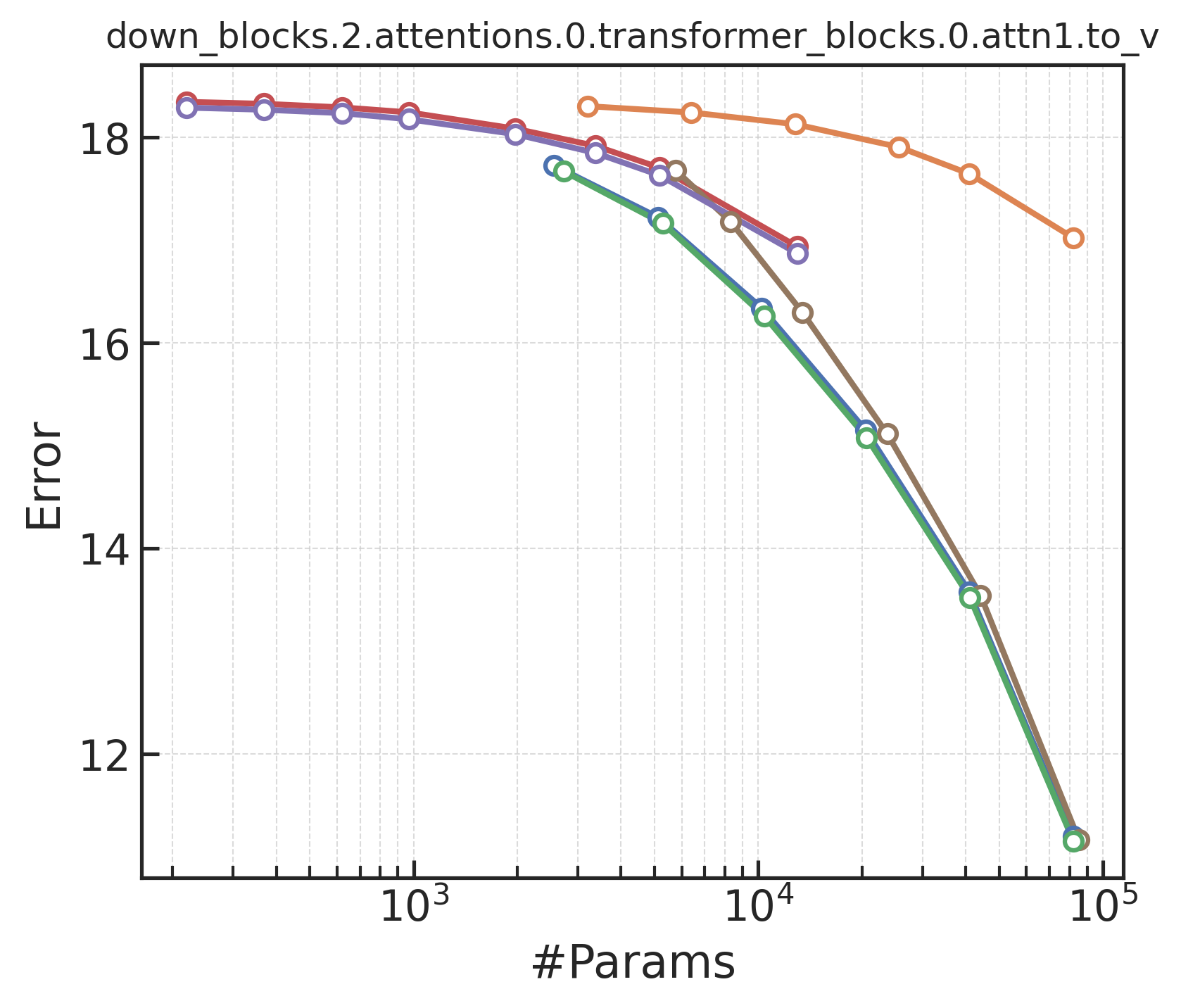}
    \end{subfigure}
    \hfill
    \begin{subfigure}[b]{0.23\linewidth}
    \includegraphics[width=\linewidth]{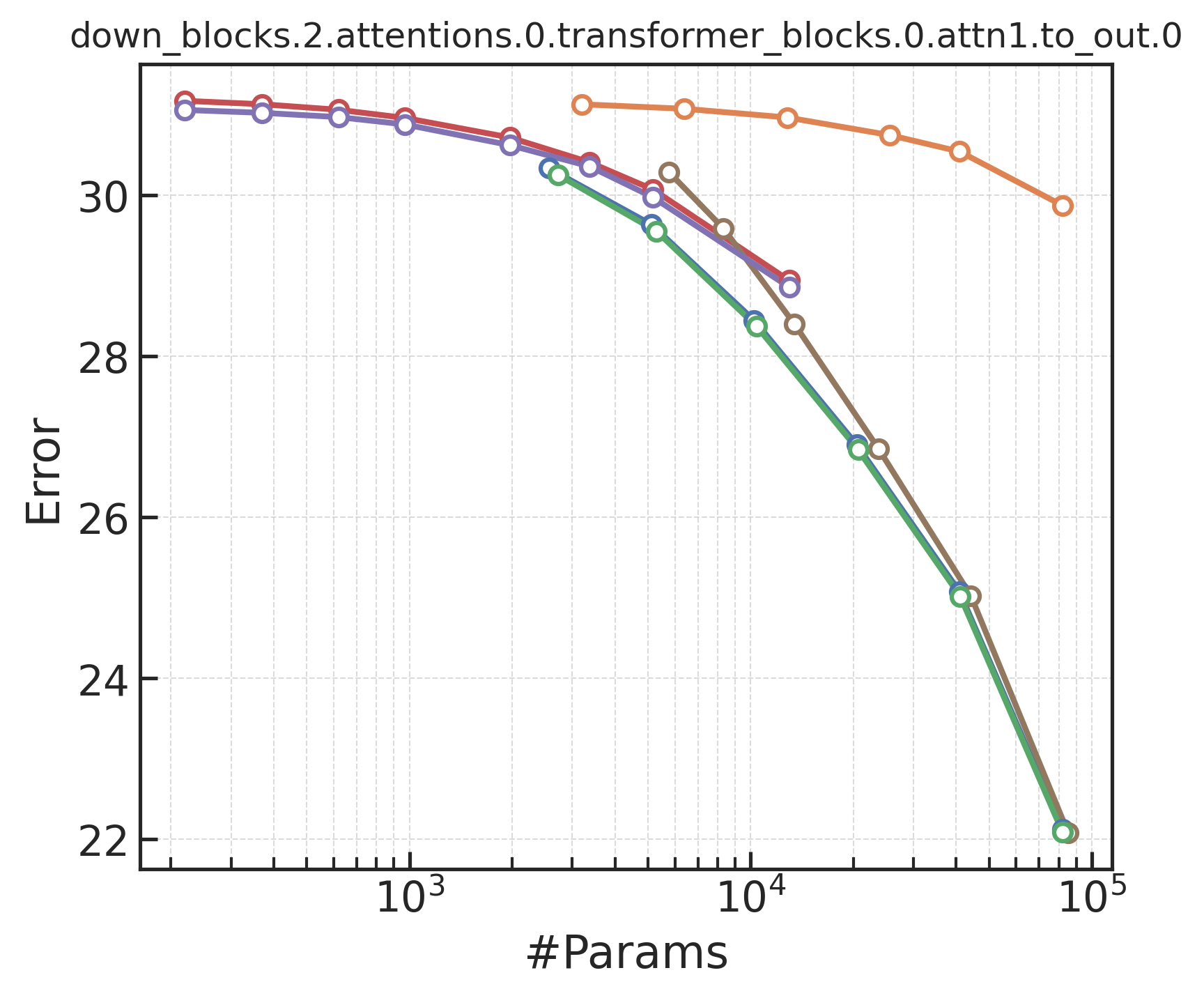}
    \end{subfigure}
    
    \begin{subfigure}[b]{0.23\linewidth}
    \includegraphics[width=\linewidth]{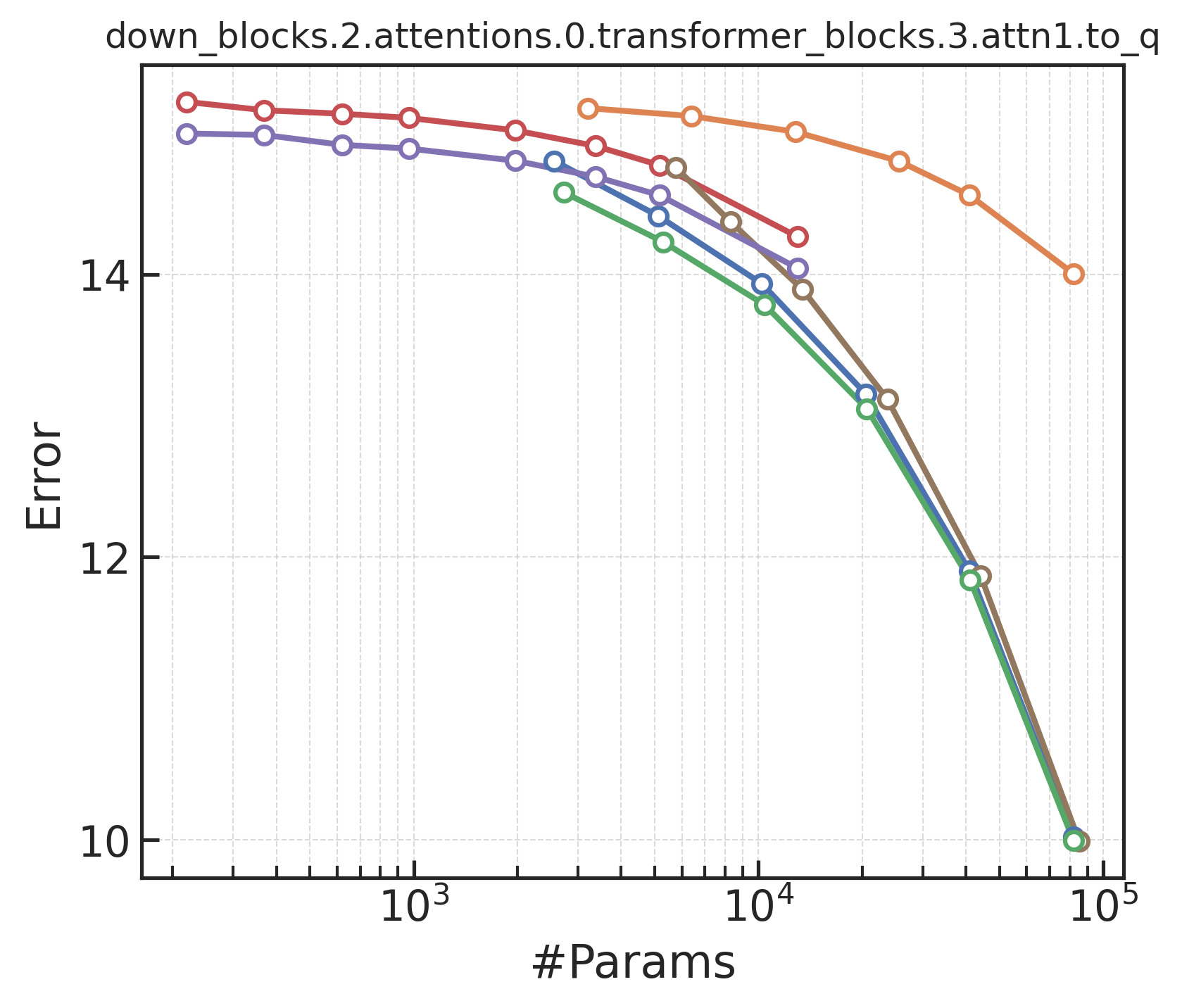}
    \end{subfigure}
    \hfill
    \begin{subfigure}[b]{0.23\linewidth}
    \includegraphics[width=\linewidth]{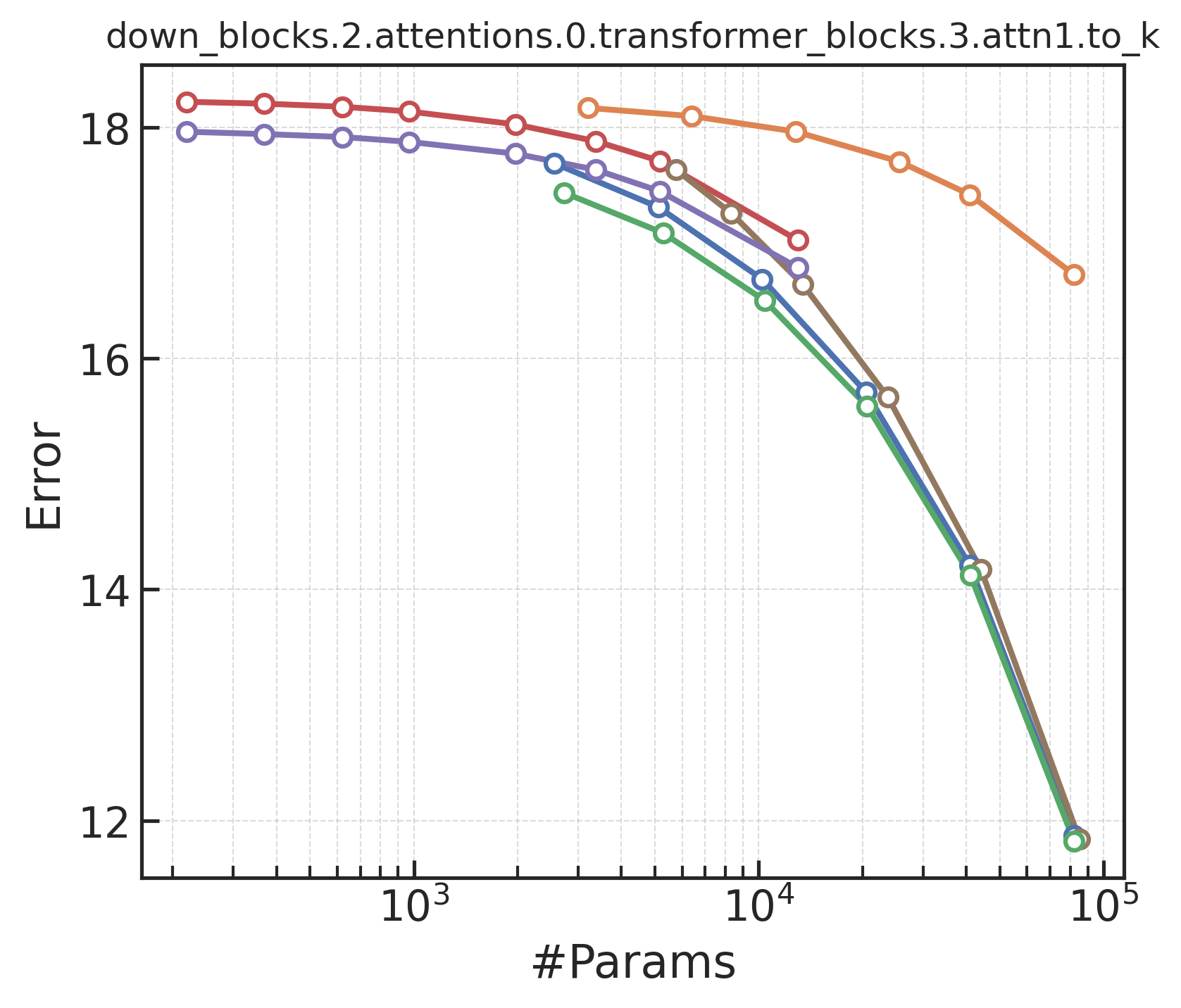}
    \end{subfigure}
    \hfill
    \begin{subfigure}[b]{0.23\linewidth}
    \includegraphics[width=\linewidth]{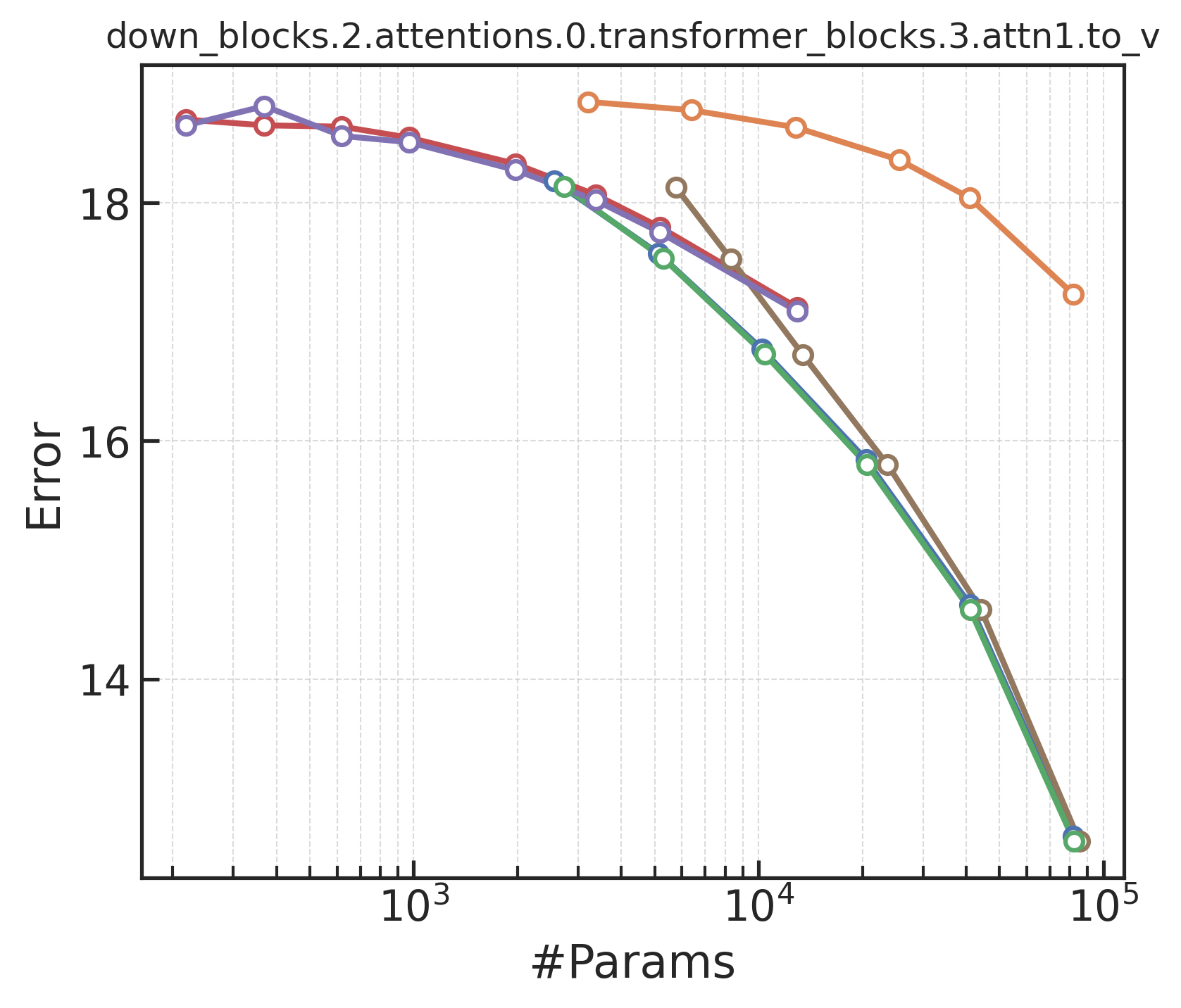}
    \end{subfigure}
    \hfill
    \begin{subfigure}[b]{0.23\linewidth}
    \includegraphics[width=\linewidth]{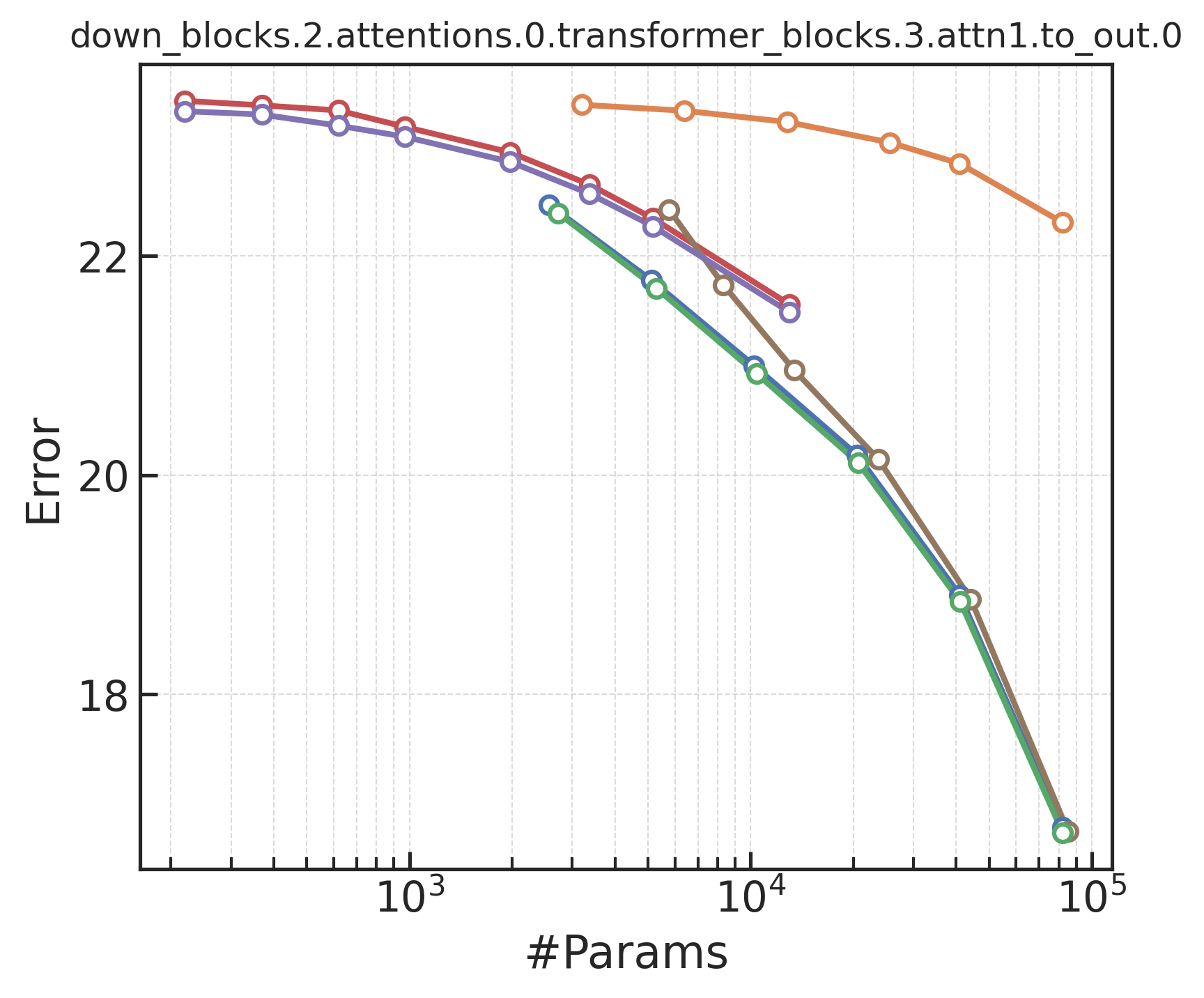}
    \end{subfigure}
    
    \begin{subfigure}[b]{0.23\linewidth}
    \includegraphics[width=\linewidth]{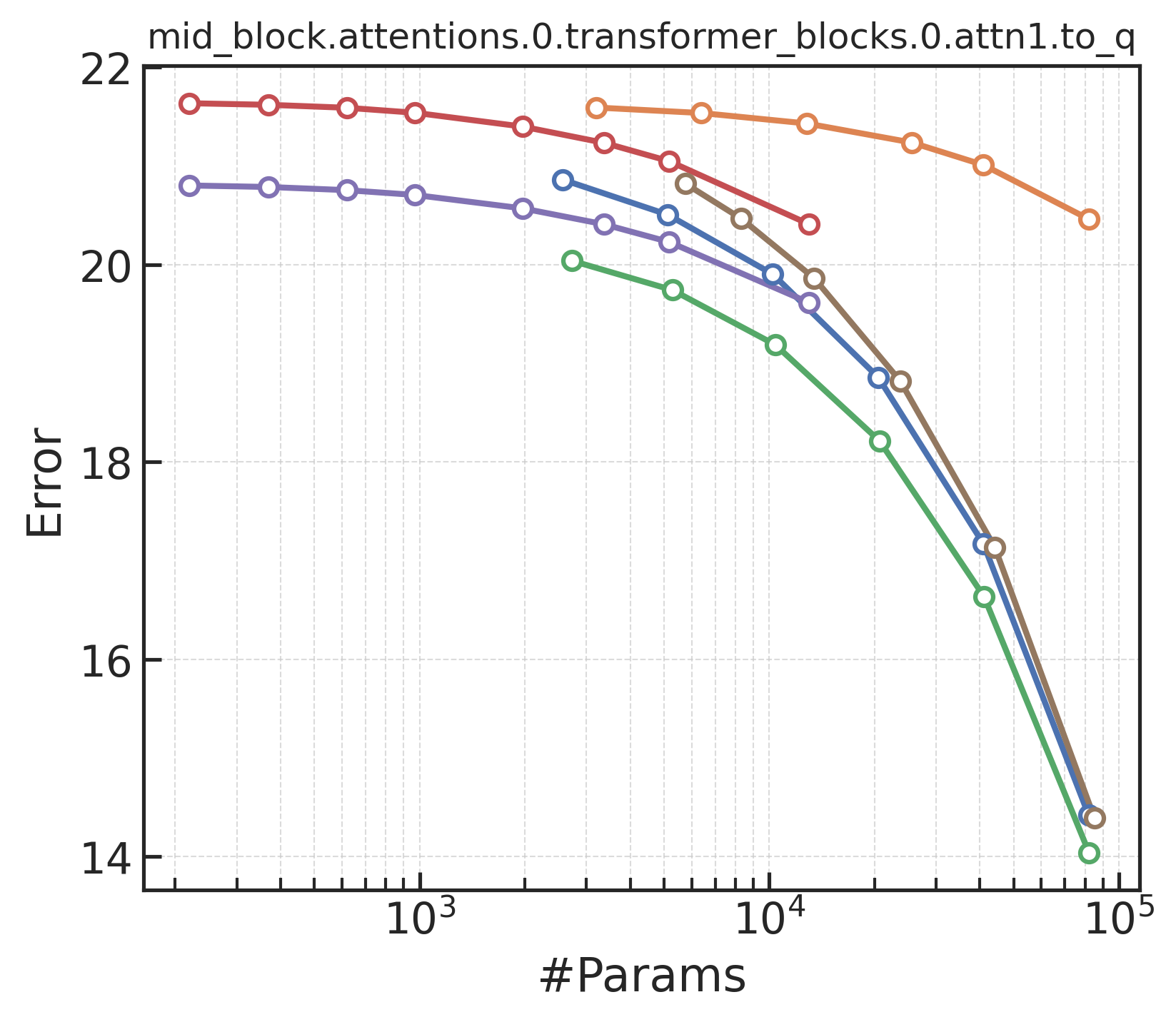}
    \end{subfigure}
    \hfill
    \begin{subfigure}[b]{0.23\linewidth}
    \includegraphics[width=\linewidth]{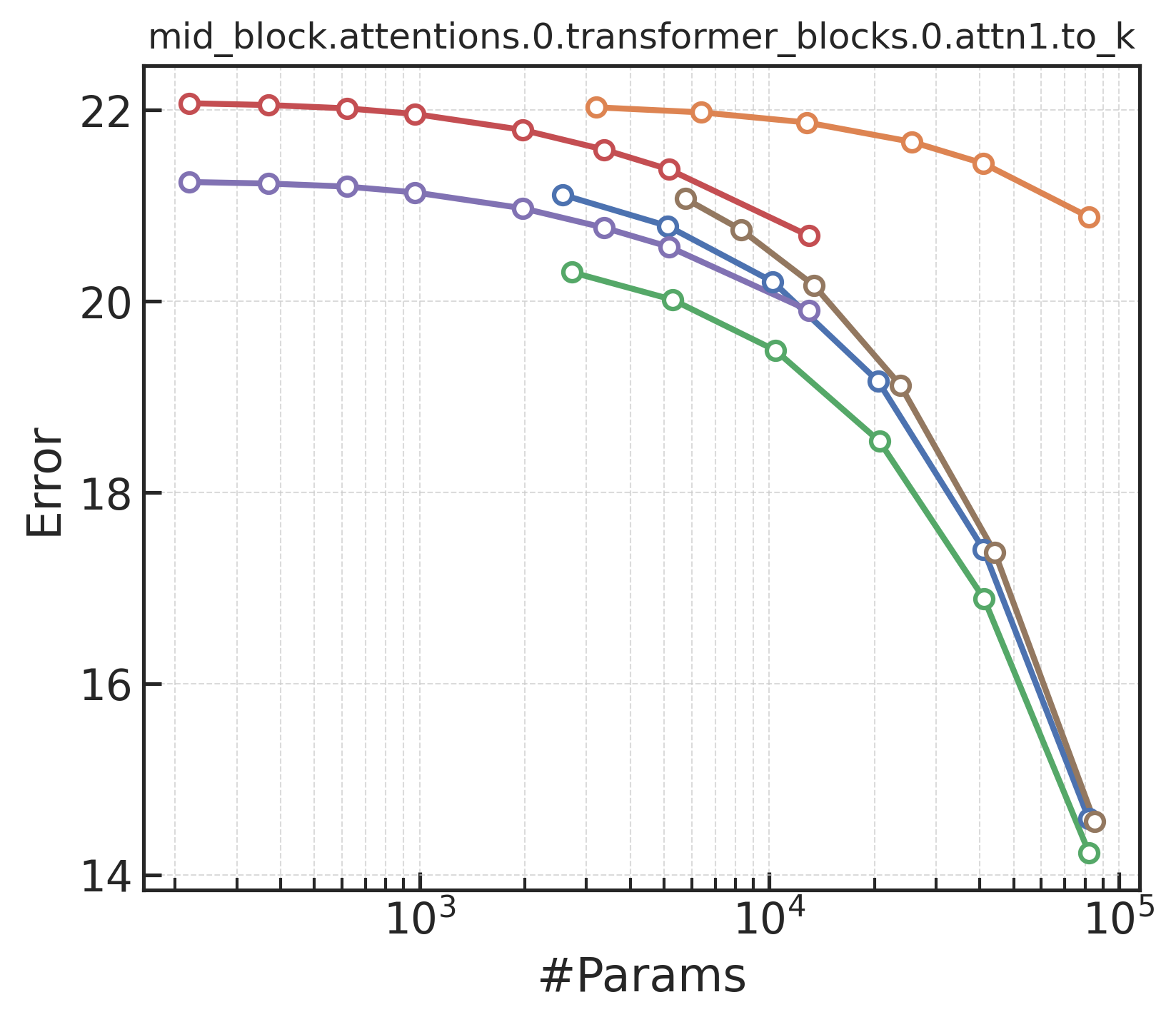}
    \end{subfigure}
    \hfill
    \begin{subfigure}[b]{0.23\linewidth}
    \includegraphics[width=\linewidth]{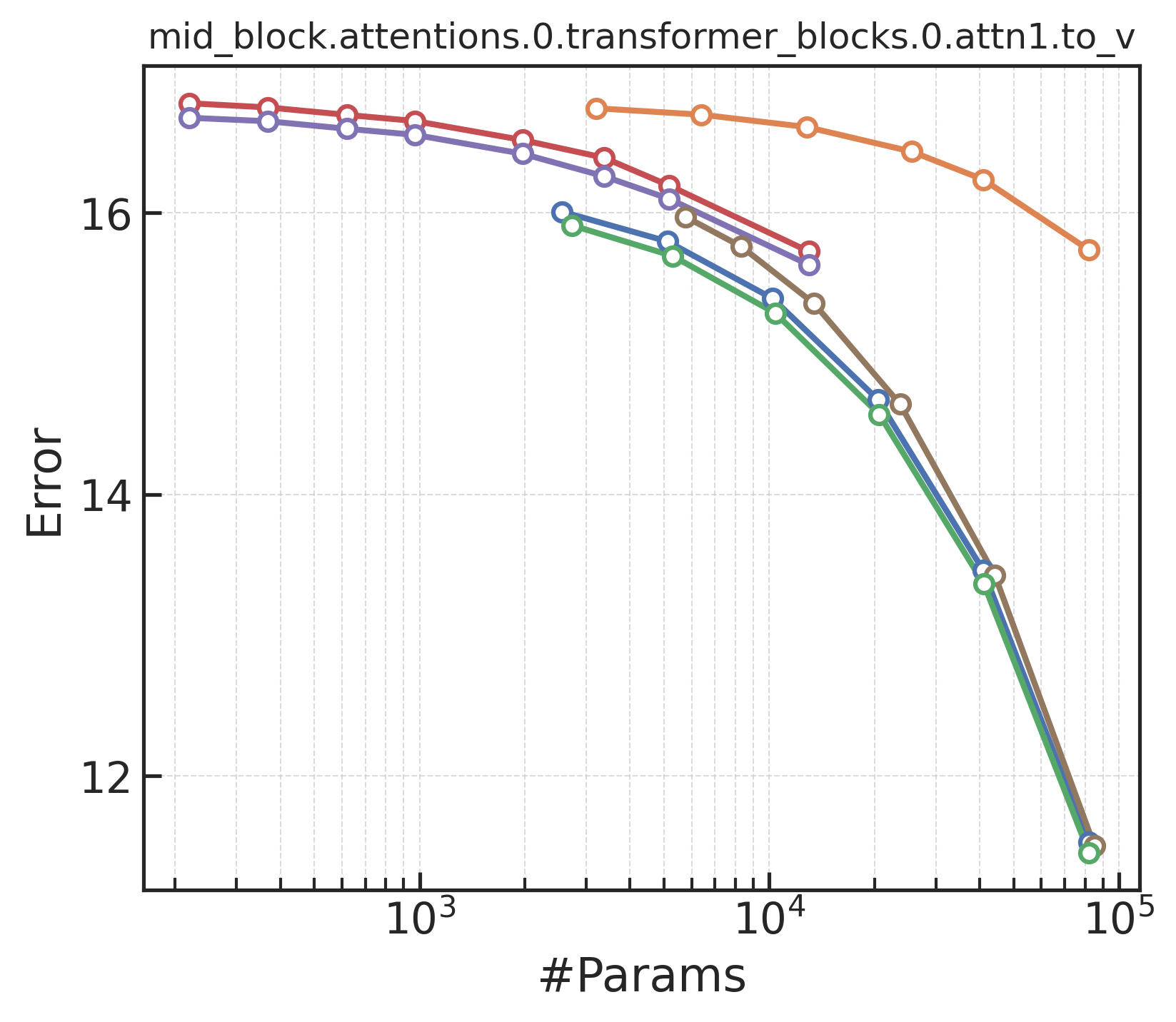}
    \end{subfigure}
    \hfill
    \begin{subfigure}[b]{0.23\linewidth}
    \includegraphics[width=\linewidth]{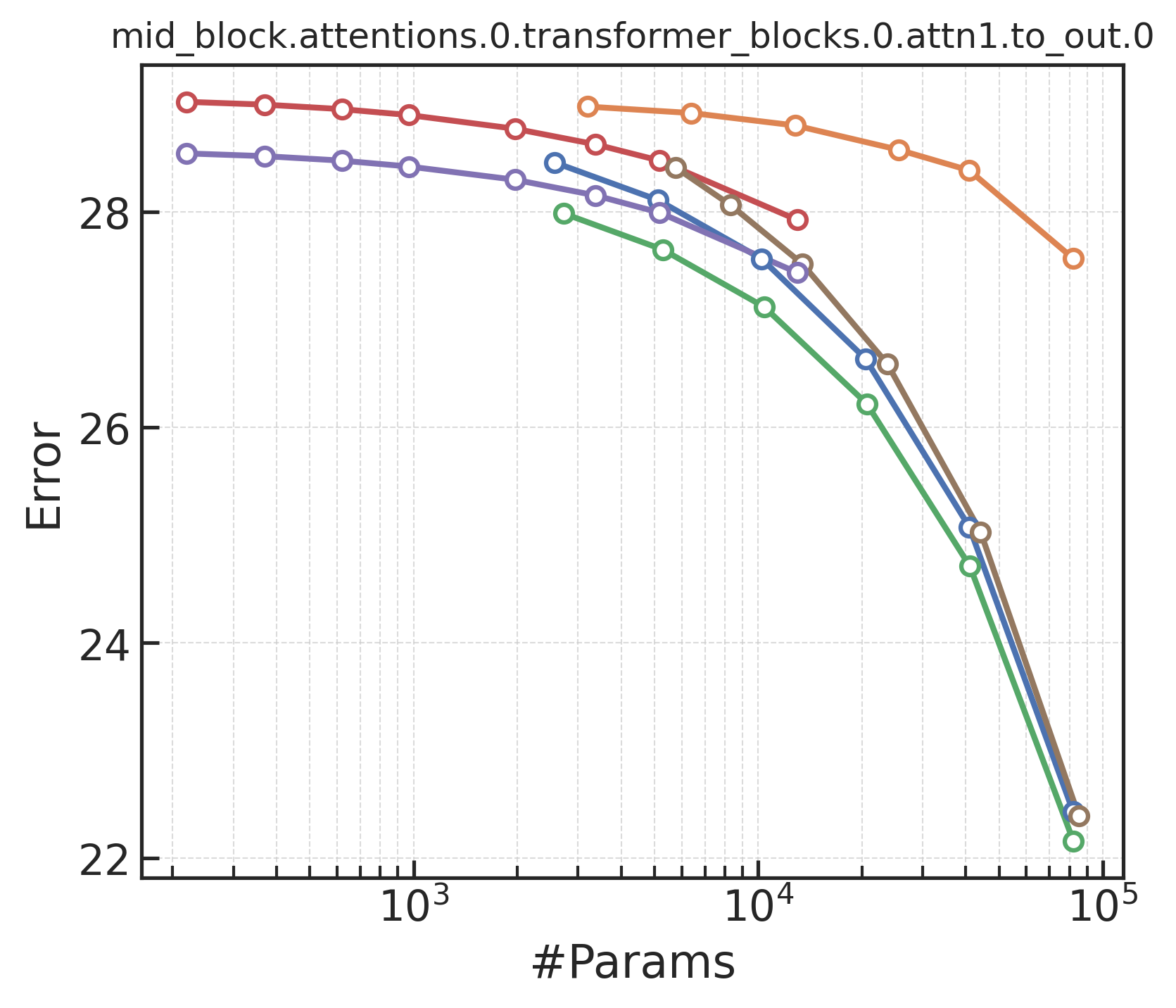}
    \end{subfigure}
    
    \begin{subfigure}[b]{0.23\linewidth}
    \includegraphics[width=\linewidth]{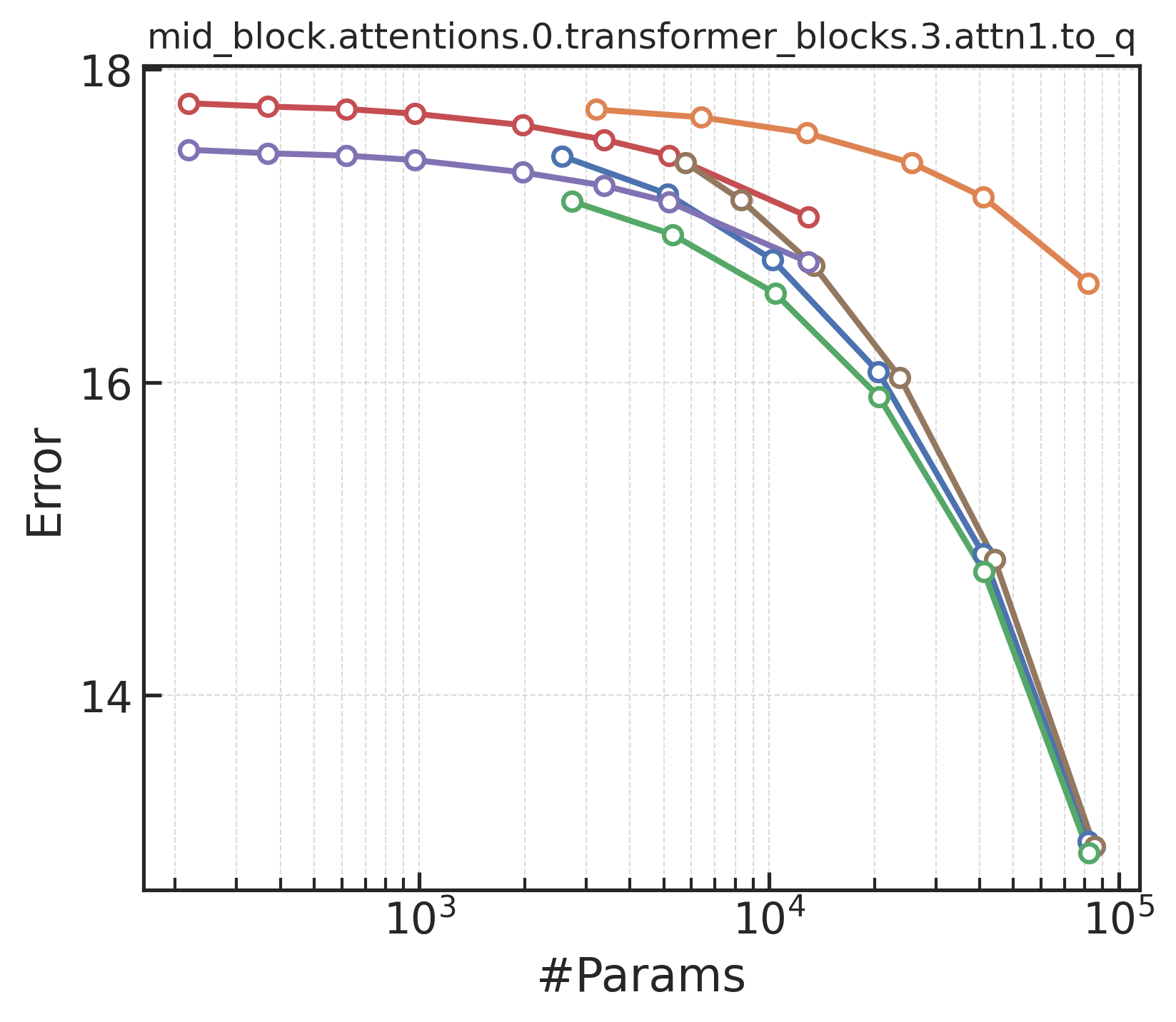}
    \end{subfigure}
    \hfill
    \begin{subfigure}[b]{0.23\linewidth}
    \includegraphics[width=\linewidth]{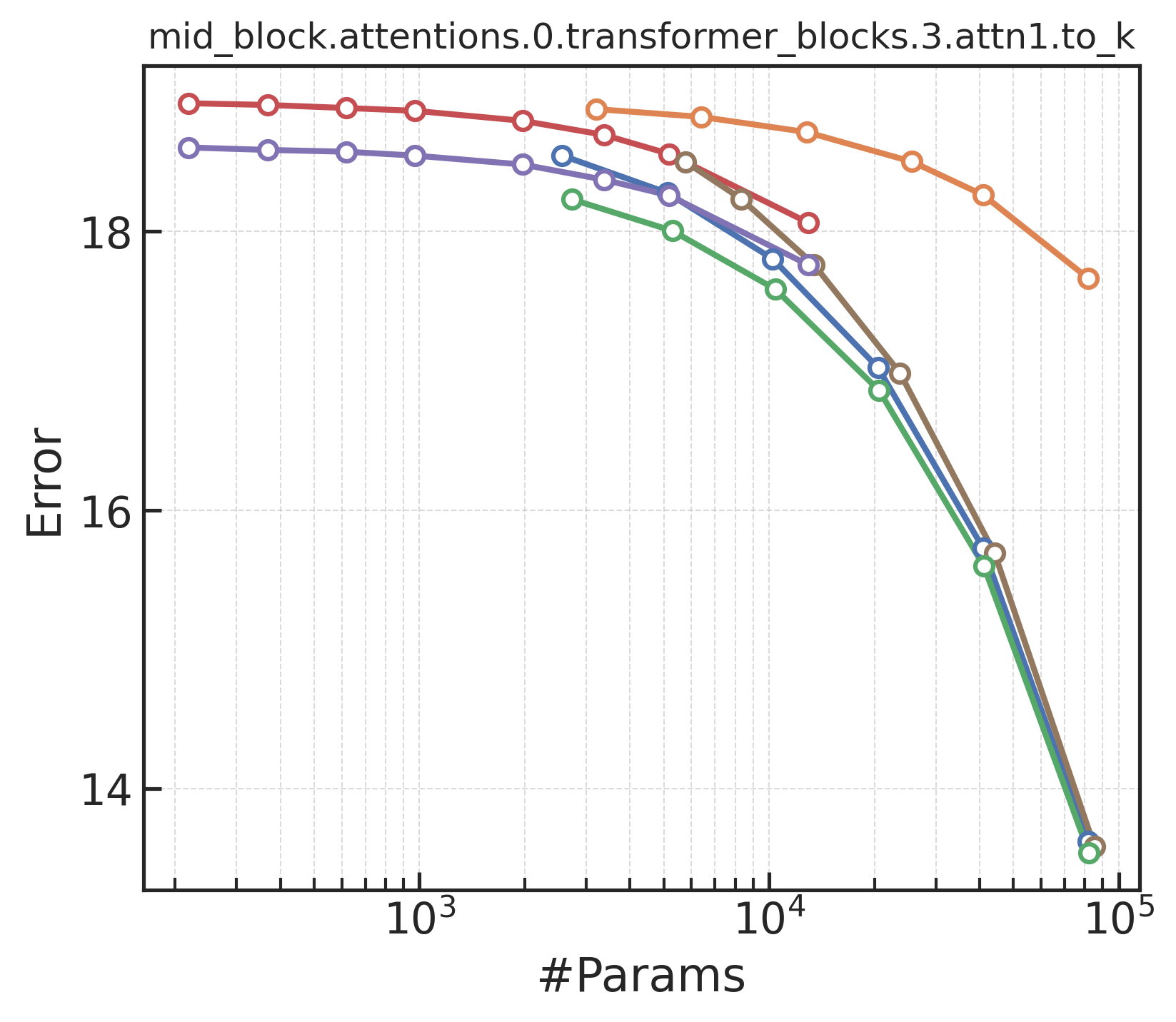}
    \end{subfigure}
    \hfill
    \begin{subfigure}[b]{0.23\linewidth}
    \includegraphics[width=\linewidth]{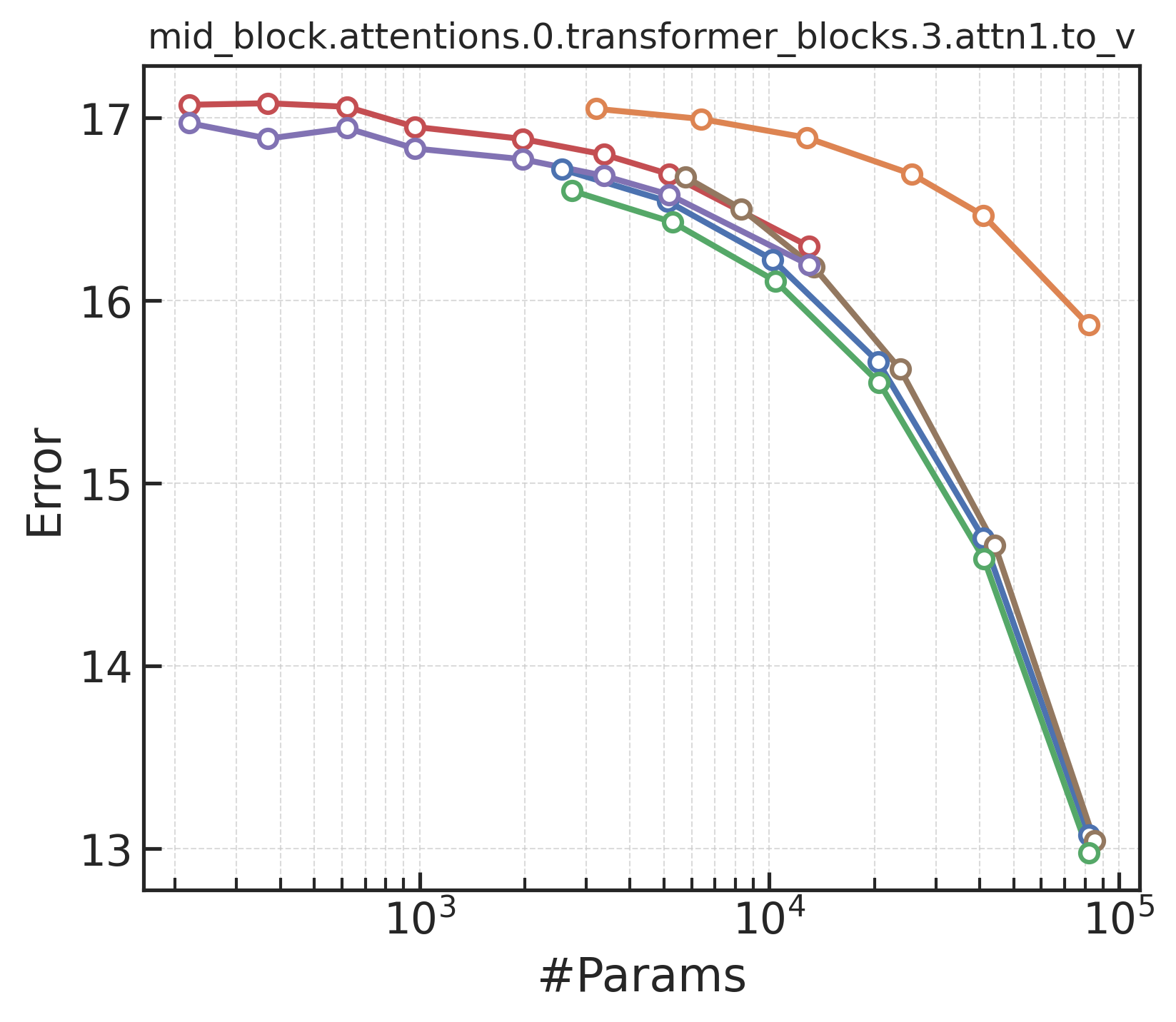}
    \end{subfigure}
    \hfill
    \begin{subfigure}[b]{0.23\linewidth}
    \includegraphics[width=\linewidth]{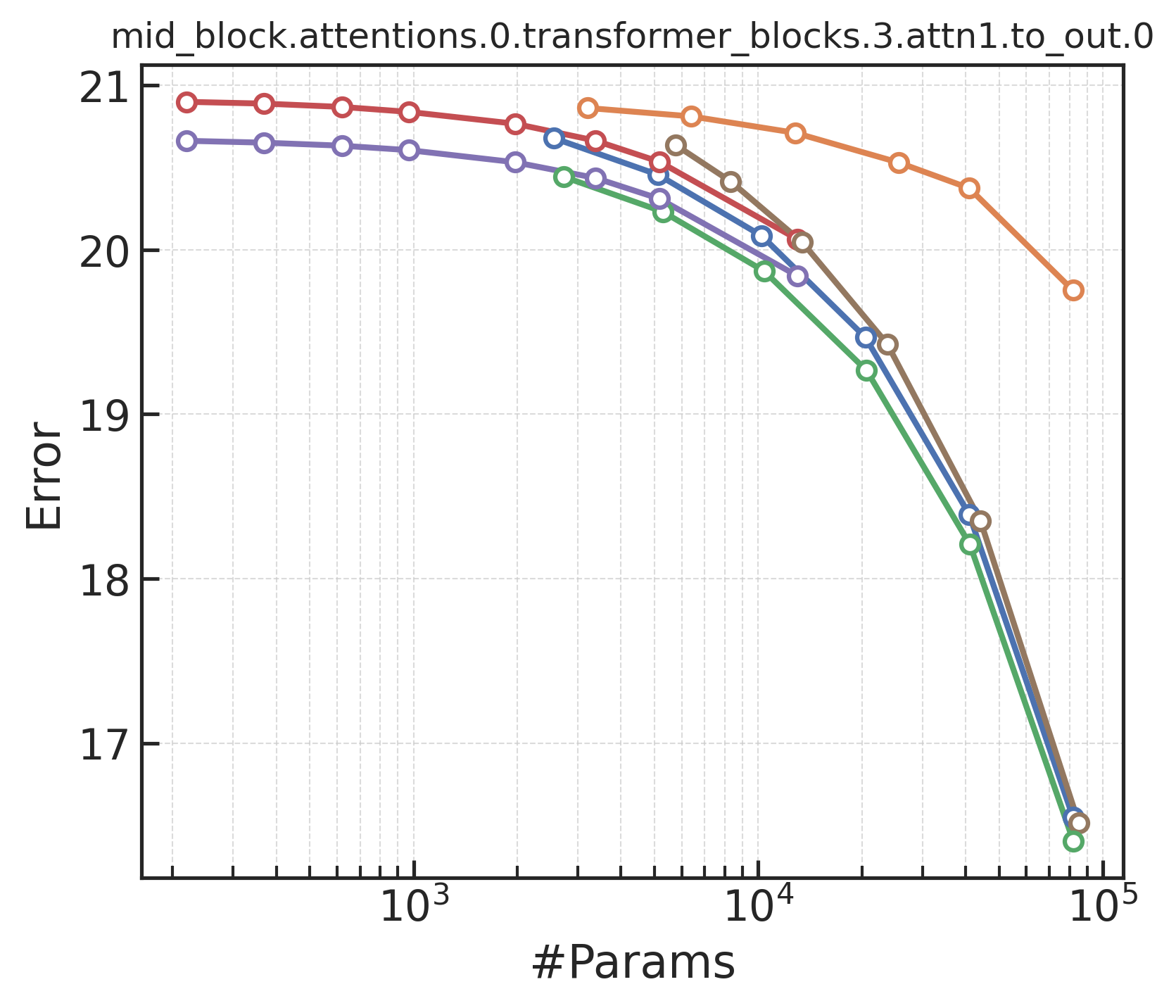}
    \end{subfigure}
    
    \begin{subfigure}[b]{0.23\linewidth}
    \includegraphics[width=\linewidth]{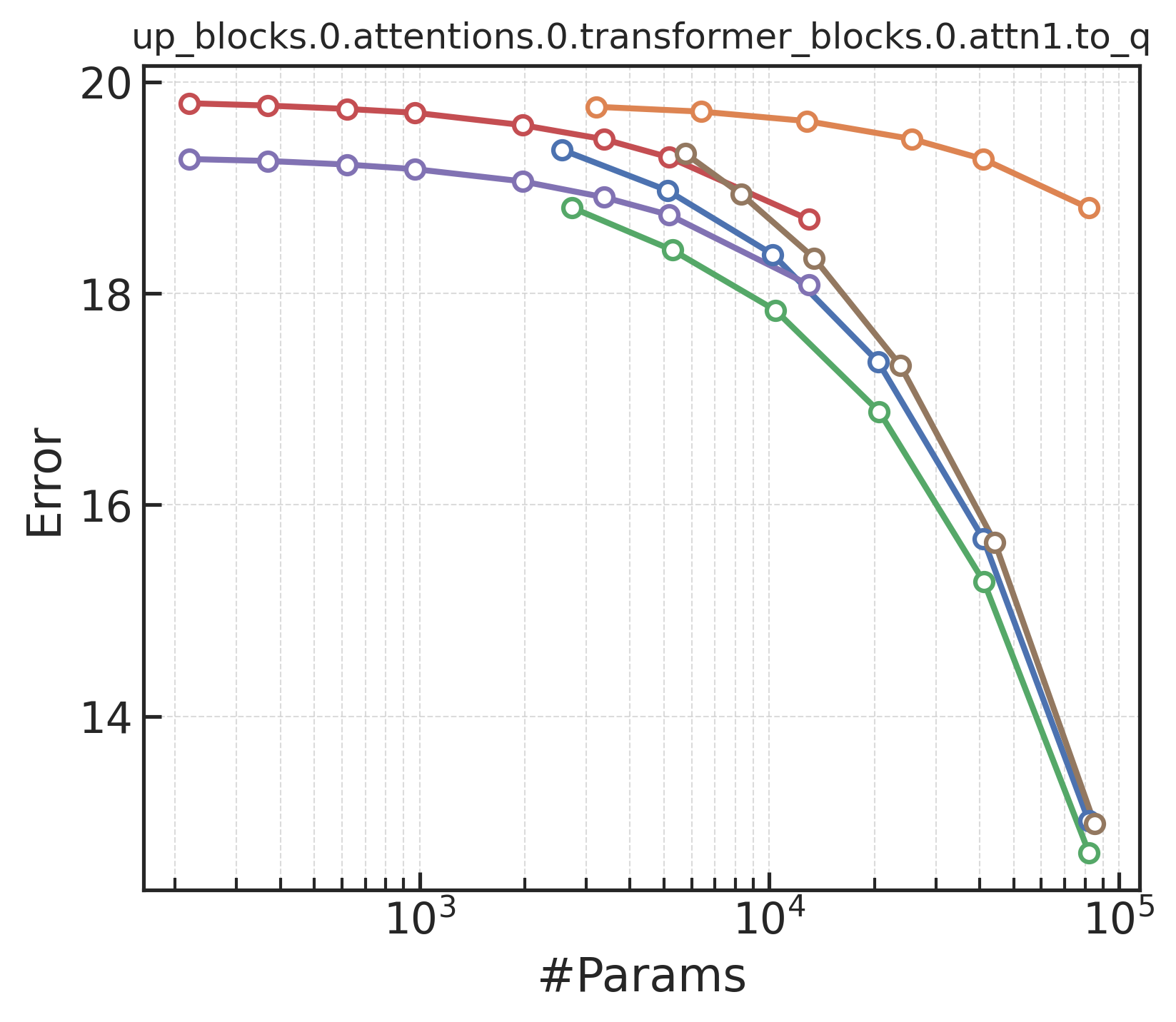}
    \end{subfigure}
    \hfill
    \begin{subfigure}[b]{0.23\linewidth}
    \includegraphics[width=\linewidth]{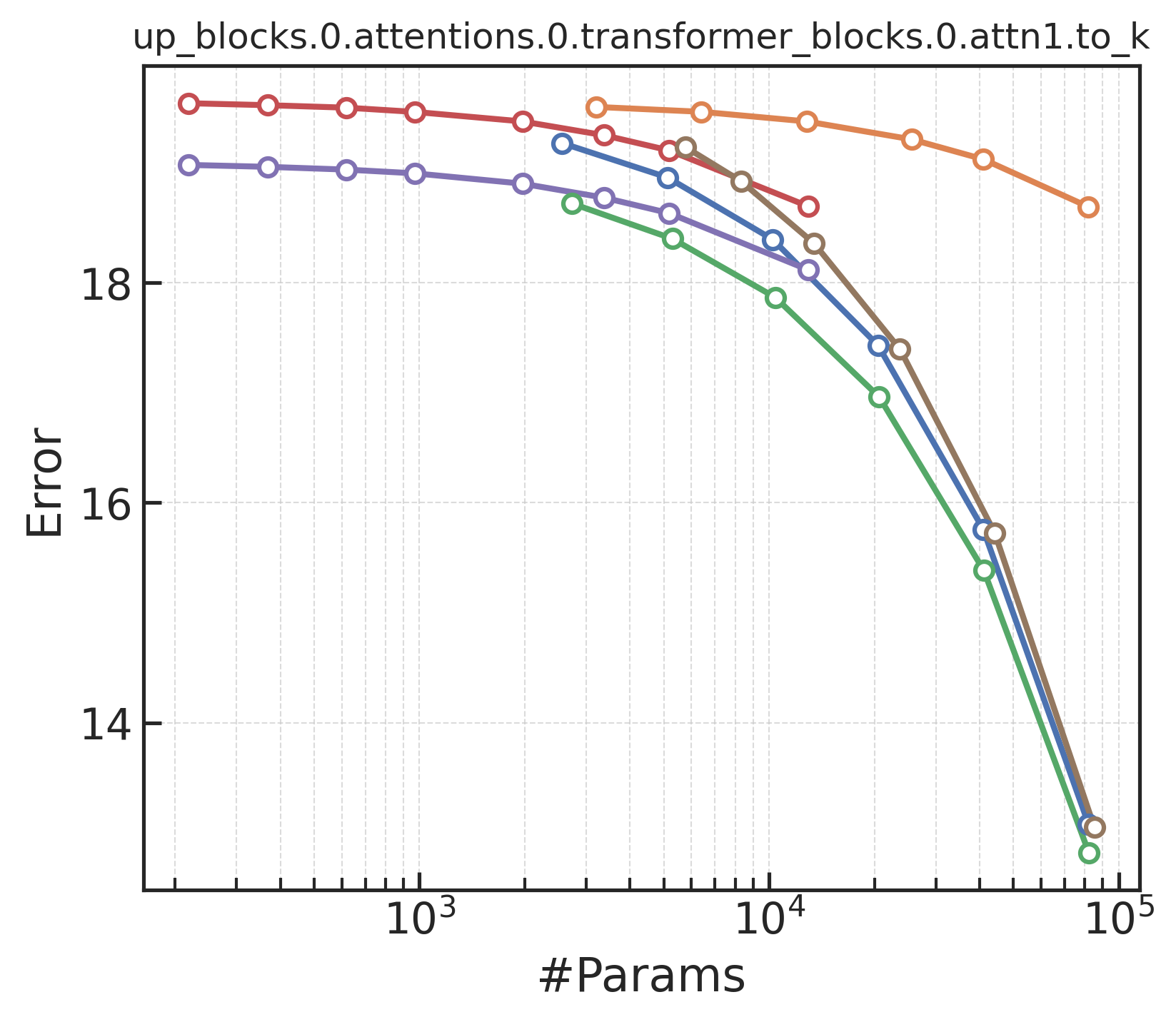}
    \end{subfigure}
    \hfill
    \begin{subfigure}[b]{0.23\linewidth}
    \includegraphics[width=\linewidth]{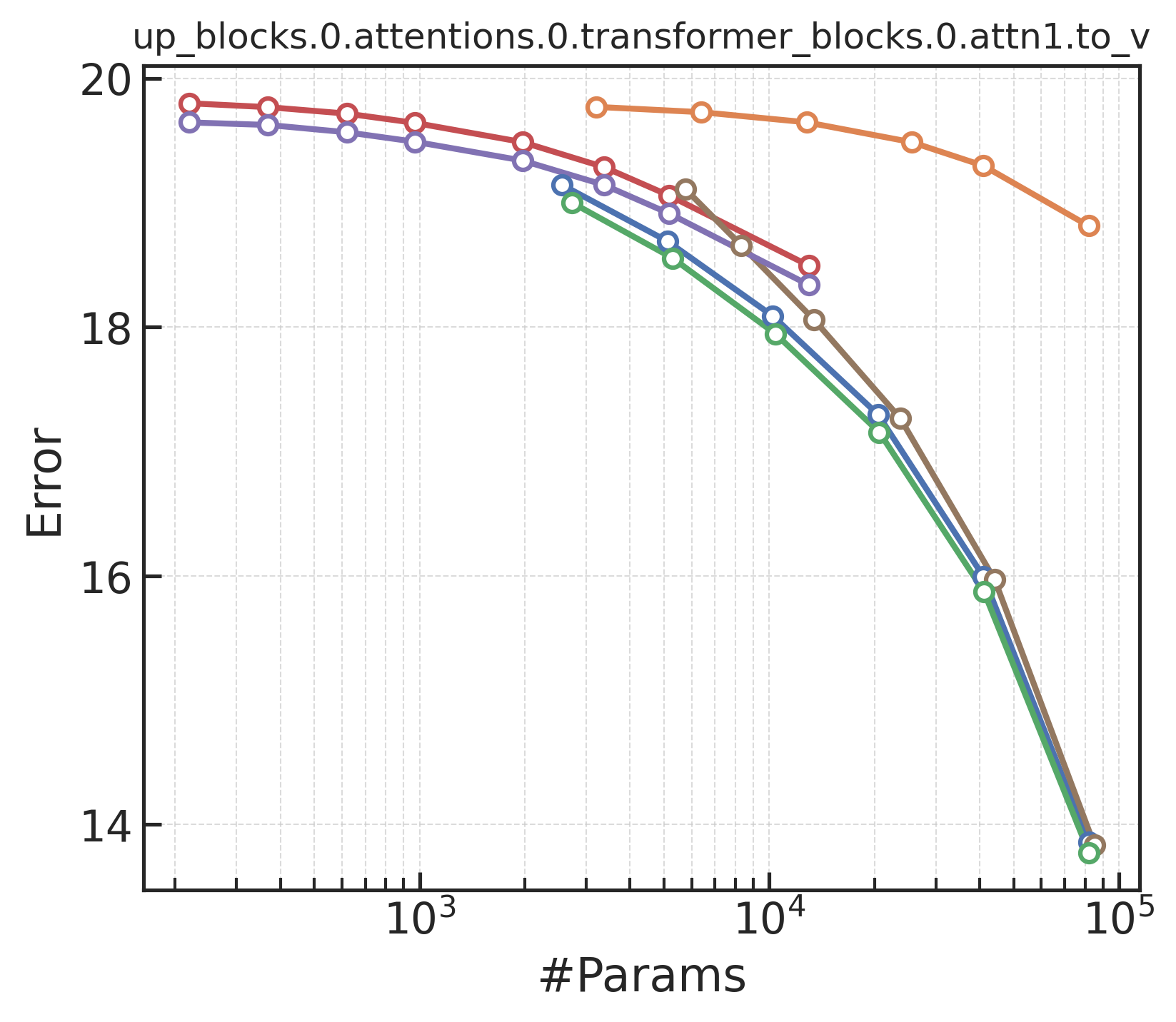}
    \end{subfigure}
    \hfill
    \begin{subfigure}[b]{0.23\linewidth}
    \includegraphics[width=\linewidth]{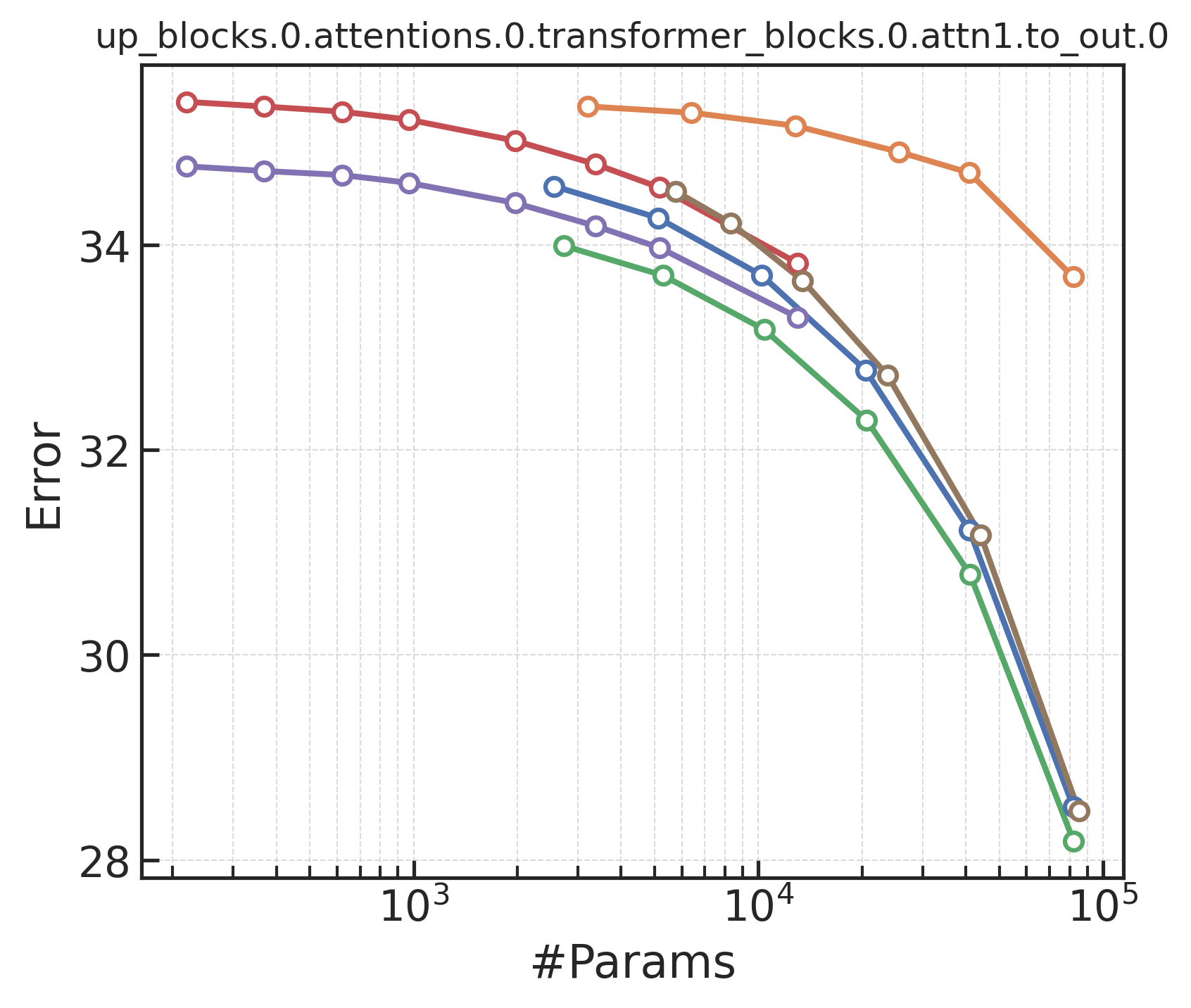}
    \end{subfigure}
    
    \begin{subfigure}[b]{0.23\linewidth}
    \includegraphics[width=\linewidth]{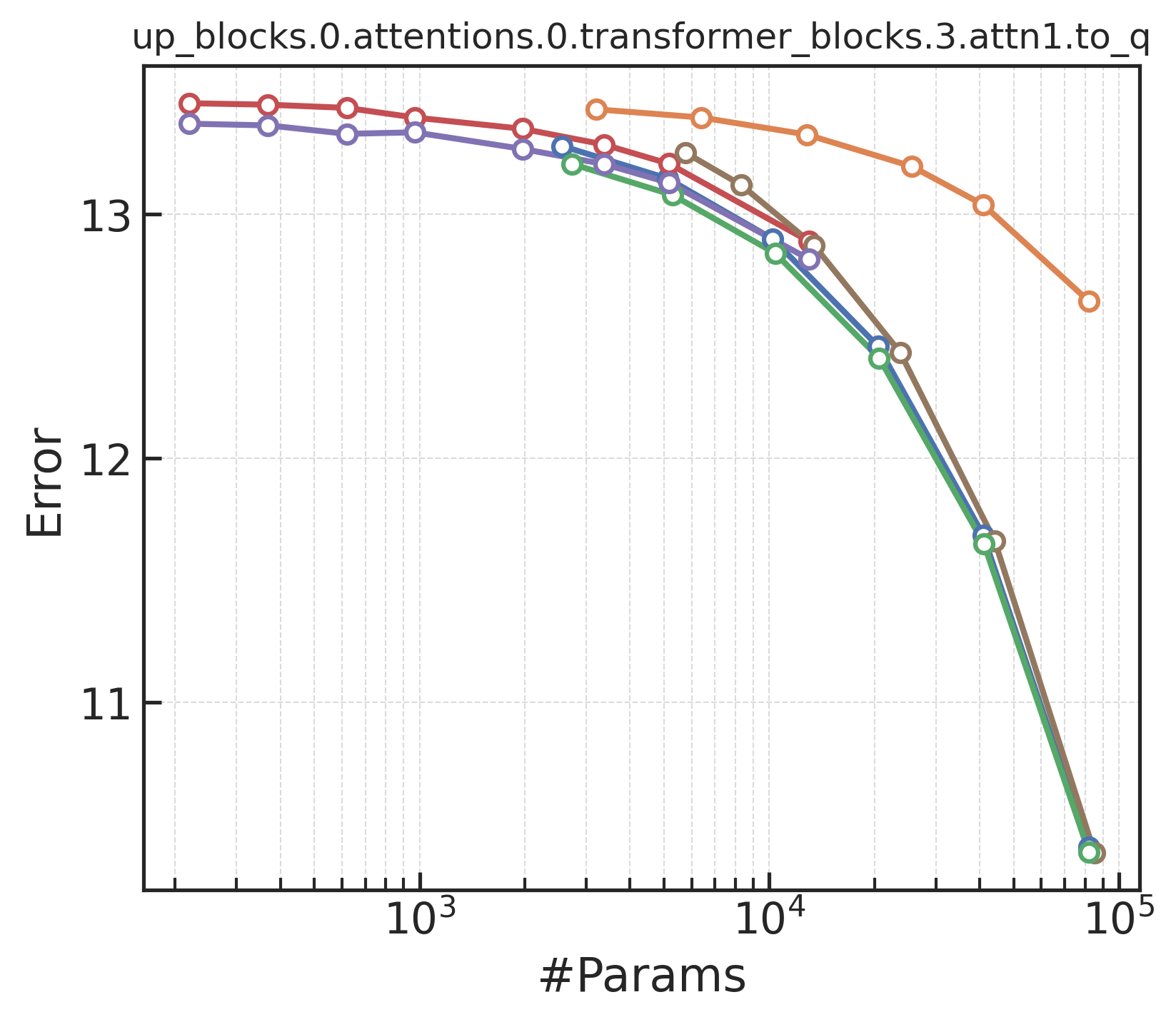}
    \end{subfigure}
    \hfill
    \begin{subfigure}[b]{0.23\linewidth}
    \includegraphics[width=\linewidth]{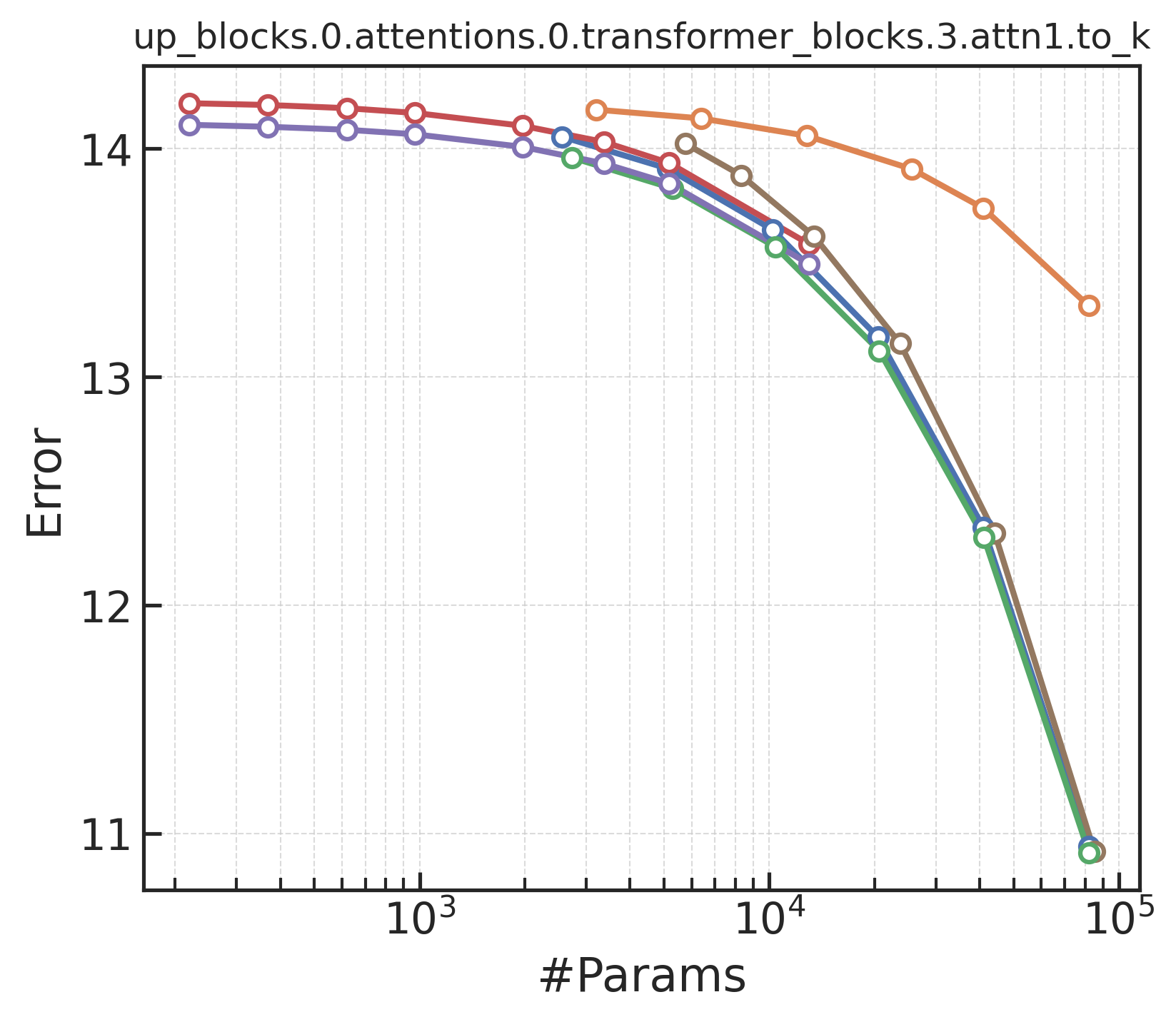}
    \end{subfigure}
    \hfill
    \begin{subfigure}[b]{0.23\linewidth}
    \includegraphics[width=\linewidth]{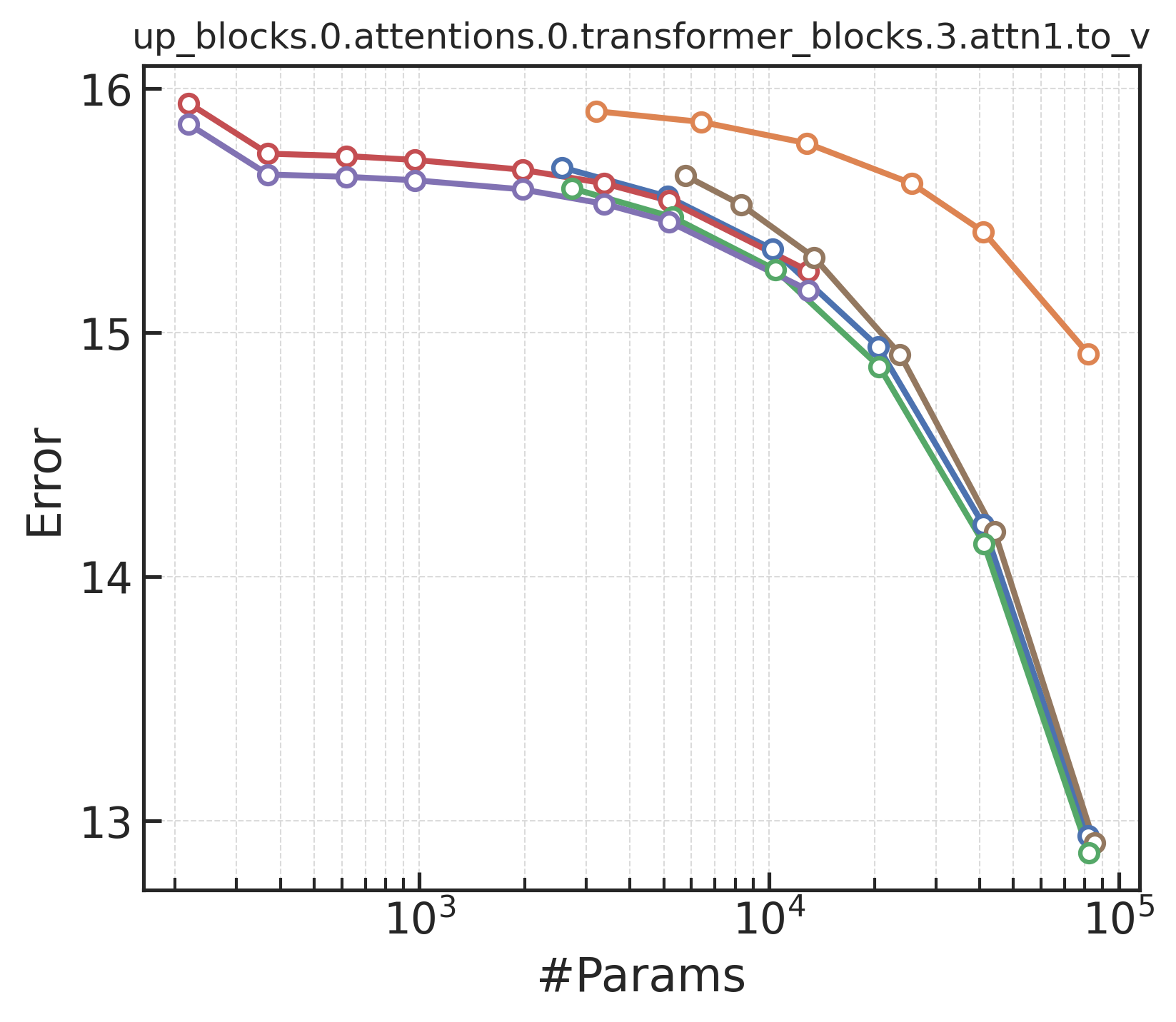}
    \end{subfigure}
    \hfill
    \begin{subfigure}[b]{0.23\linewidth}
    \includegraphics[width=\linewidth]{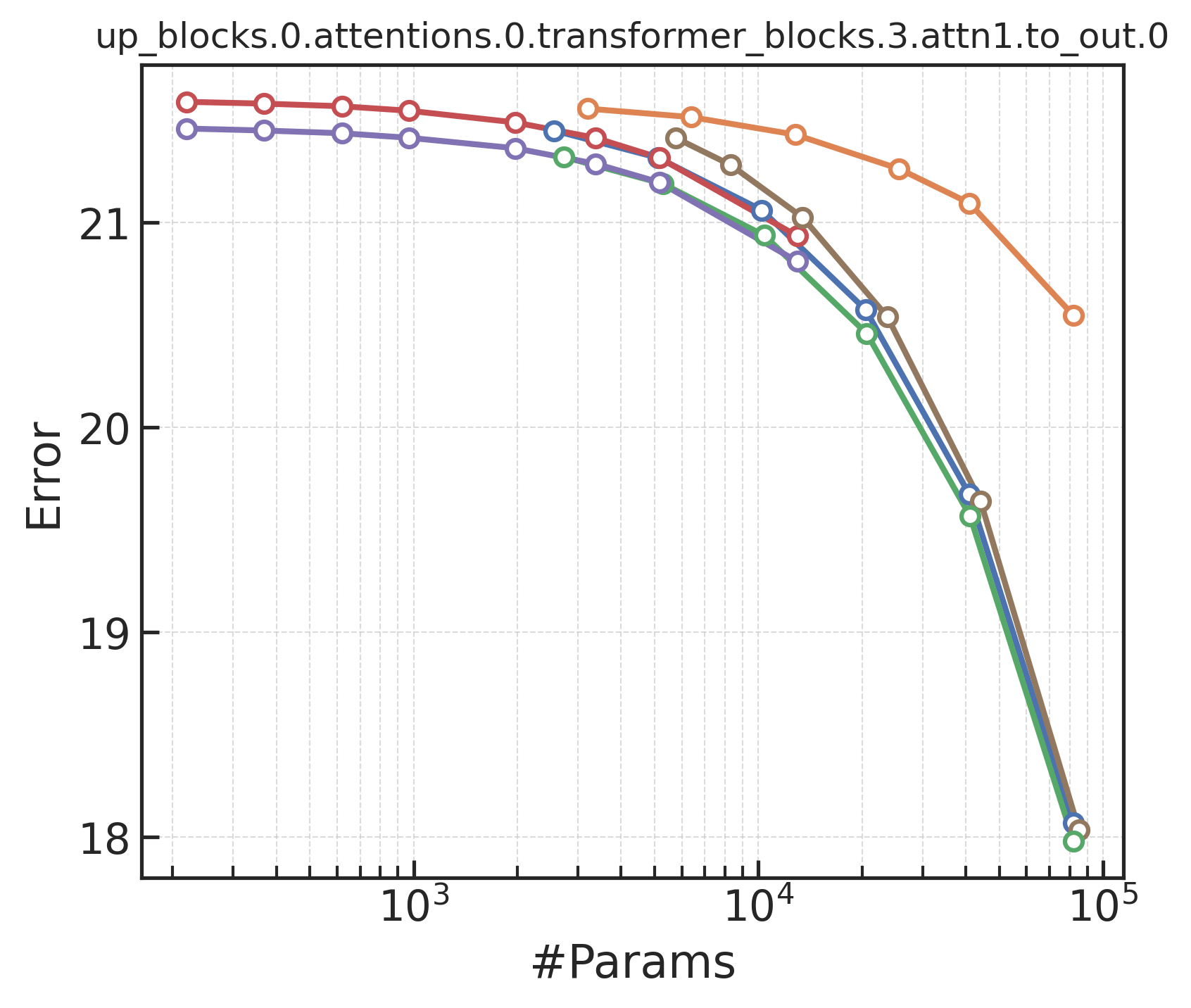}
    \end{subfigure}
    \caption{Simulation on different layers of the SDXL and SDXL-Inpaint models.}\label{fig:sdxl-weight-app}
\end{figure*}

%% file: contents/fig_and_tab/llama_weights.tex
\begin{figure*}[t]
    \centering
    \begin{subfigure}[b]{0.23\linewidth}
    \includegraphics[width=\linewidth]{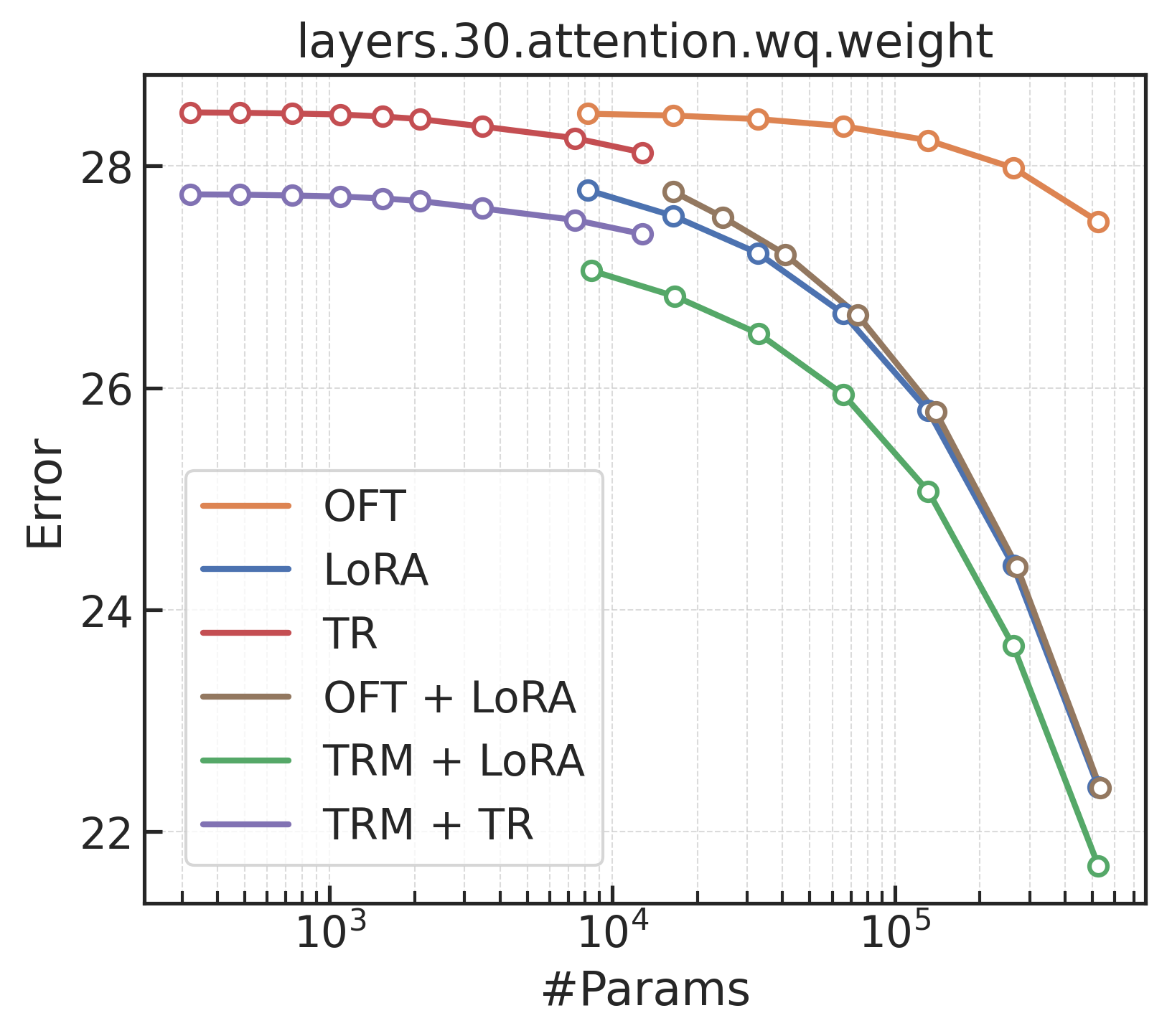}
    \end{subfigure}
    \hfill
    \begin{subfigure}[b]{0.23\linewidth}
    \includegraphics[width=\linewidth]{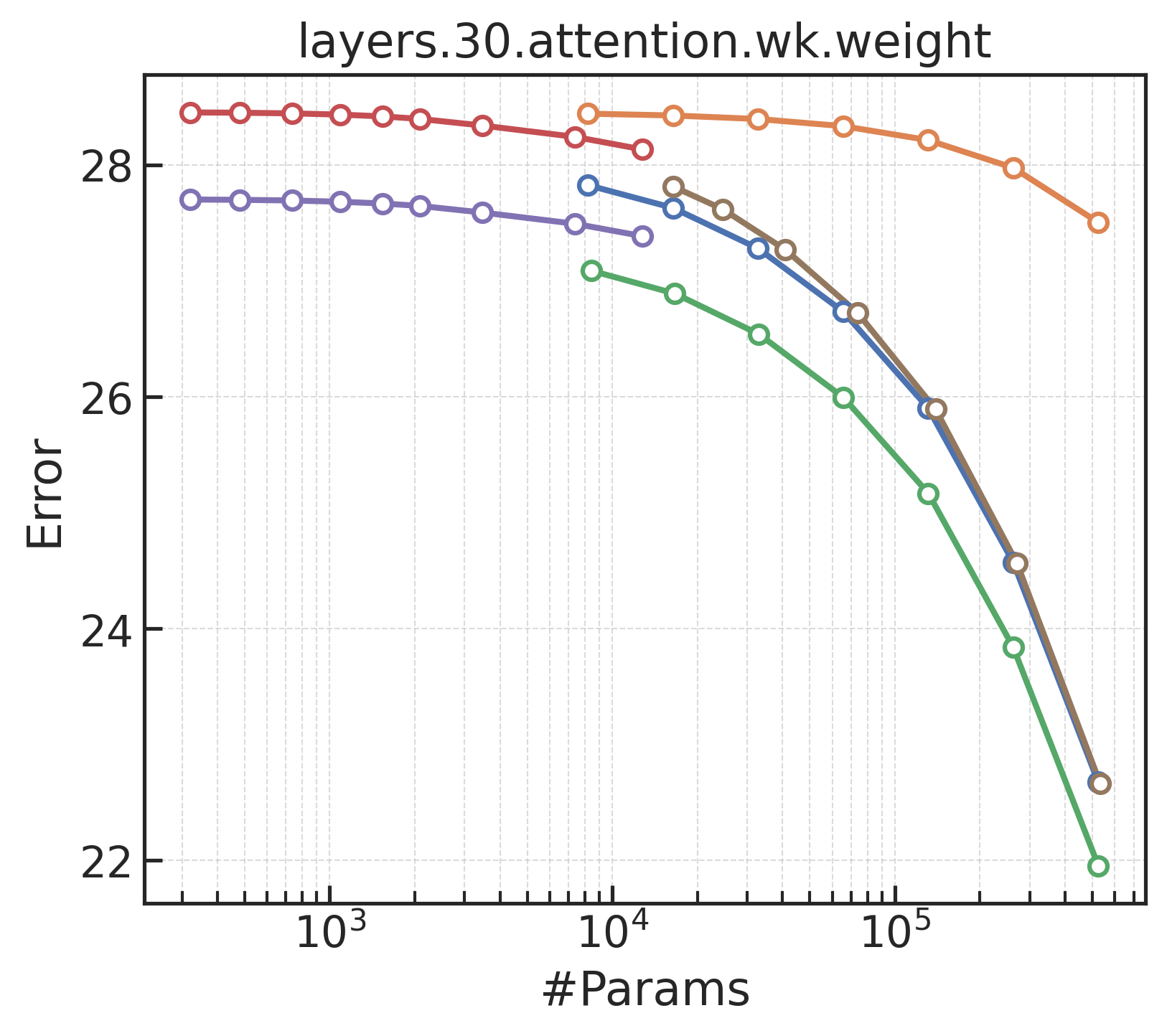}
    \end{subfigure}
    \hfill
    \begin{subfigure}[b]{0.23\linewidth}
    \includegraphics[width=\linewidth]{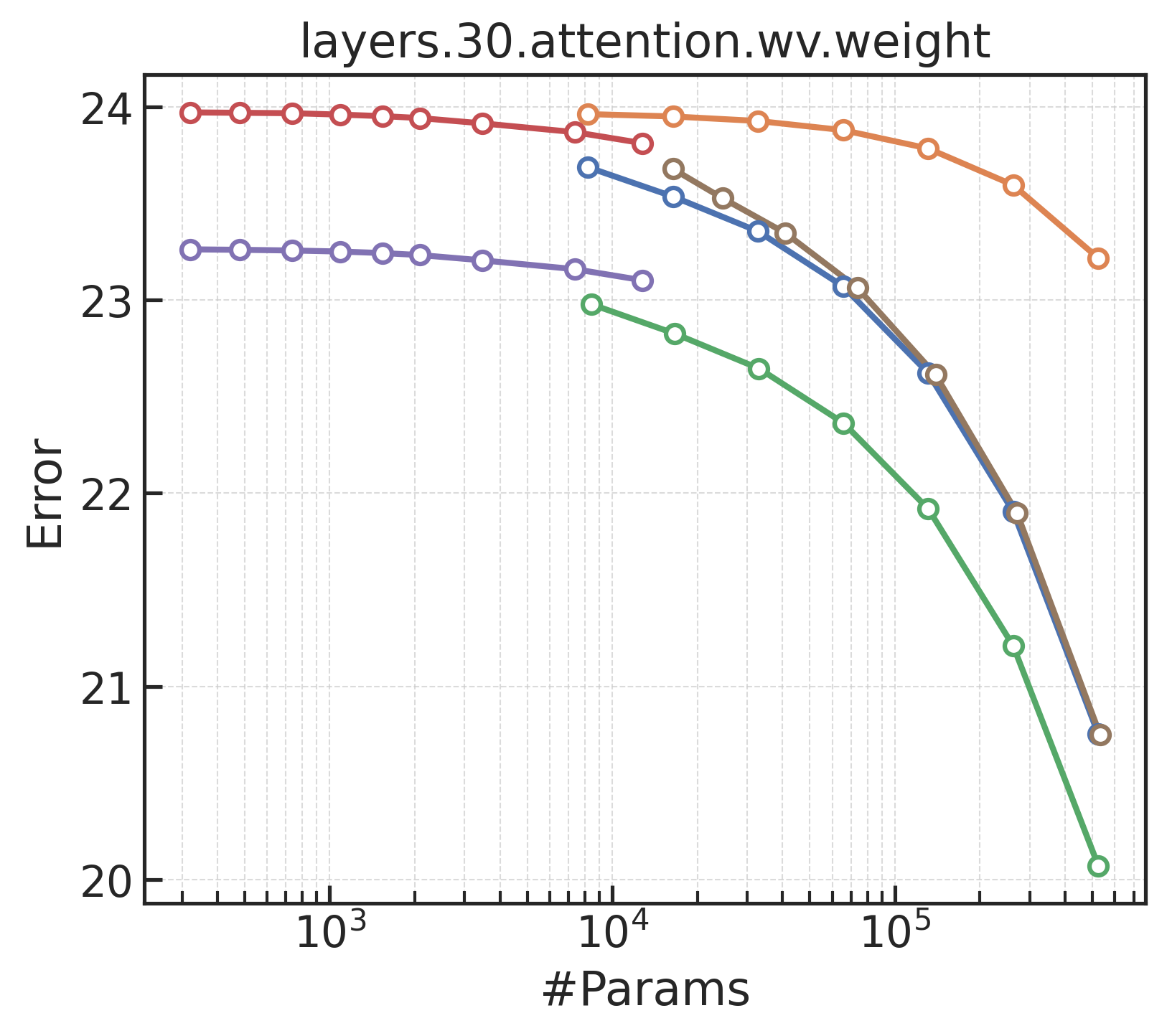}
    \end{subfigure}
    \hfill
    \begin{subfigure}[b]{0.23\linewidth}
    \includegraphics[width=\linewidth]{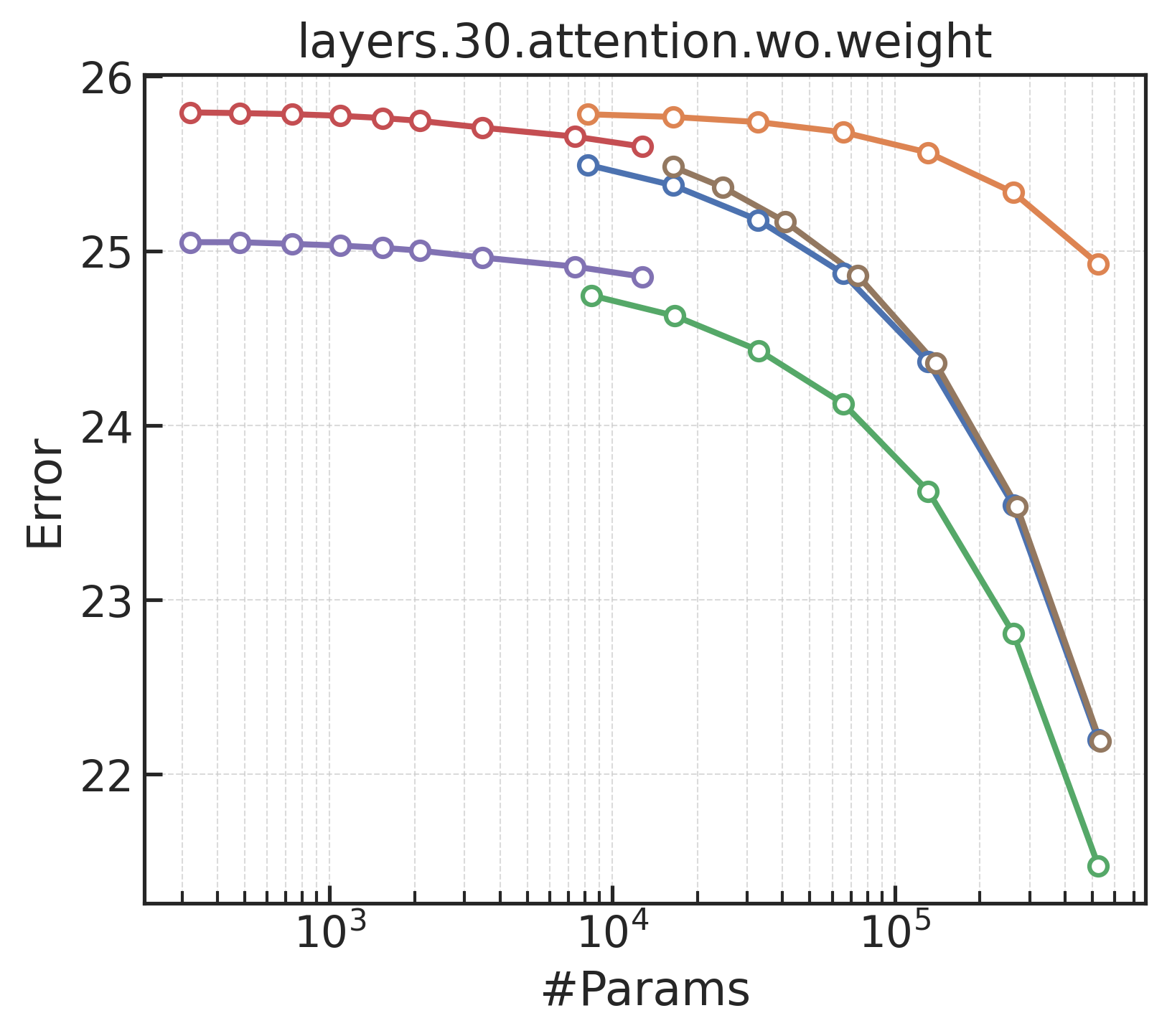}
    \end{subfigure}
    
    \begin{subfigure}[b]{0.23\linewidth}
    \includegraphics[width=\linewidth]{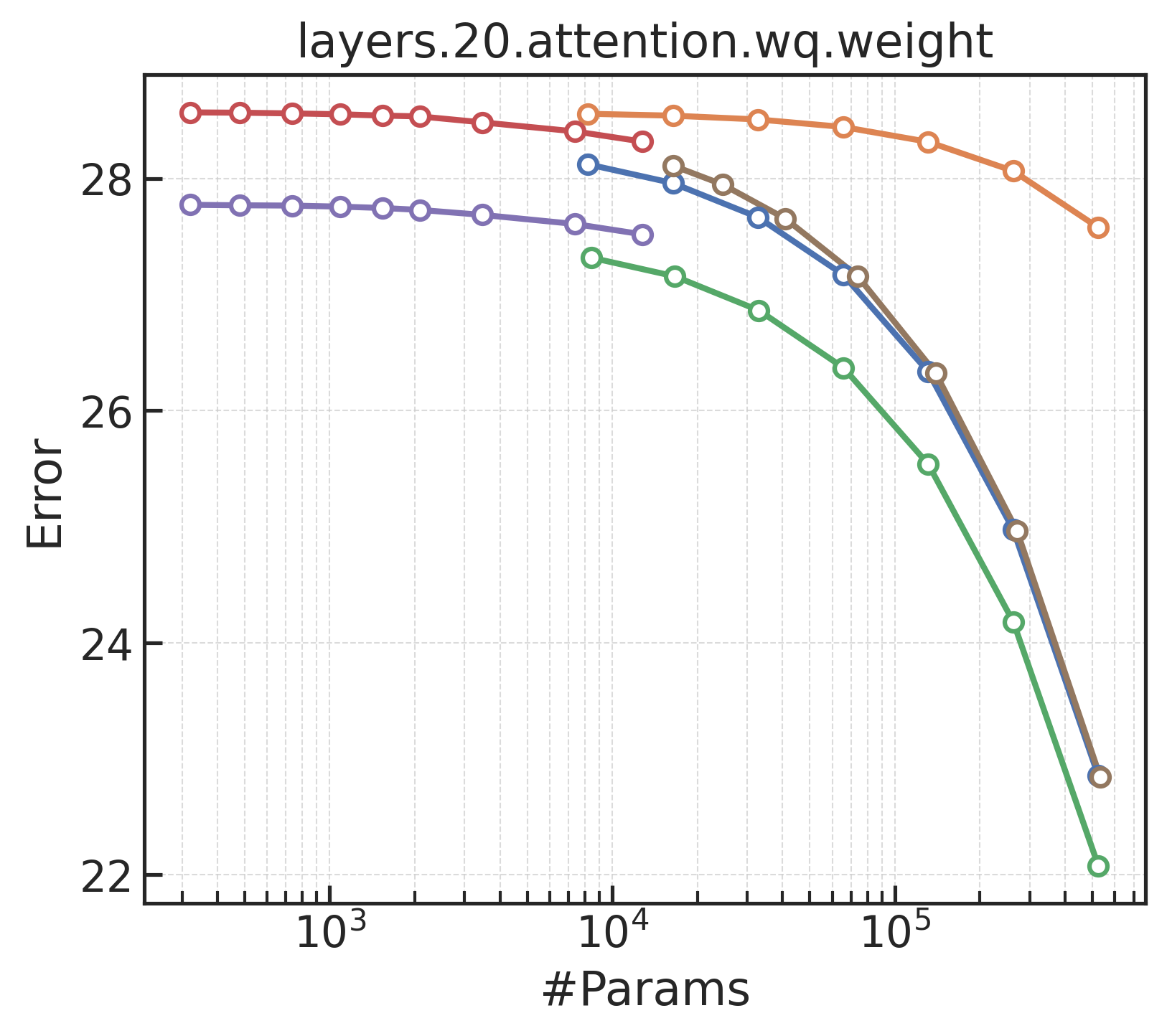}
    \end{subfigure}
    \hfill
    \begin{subfigure}[b]{0.23\linewidth}
    \includegraphics[width=\linewidth]{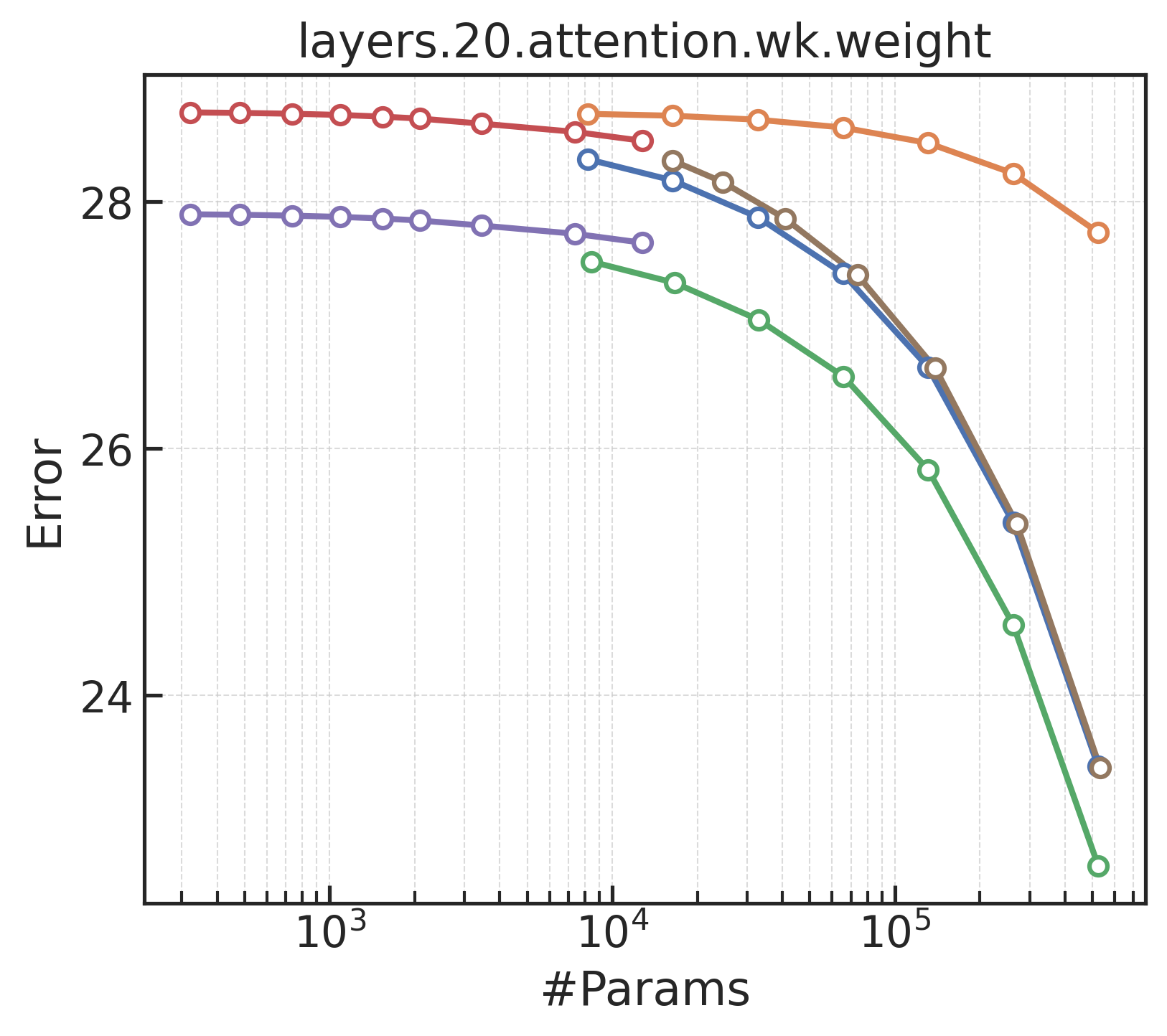}
    \end{subfigure}
    \hfill
    \begin{subfigure}[b]{0.23\linewidth}
    \includegraphics[width=\linewidth]{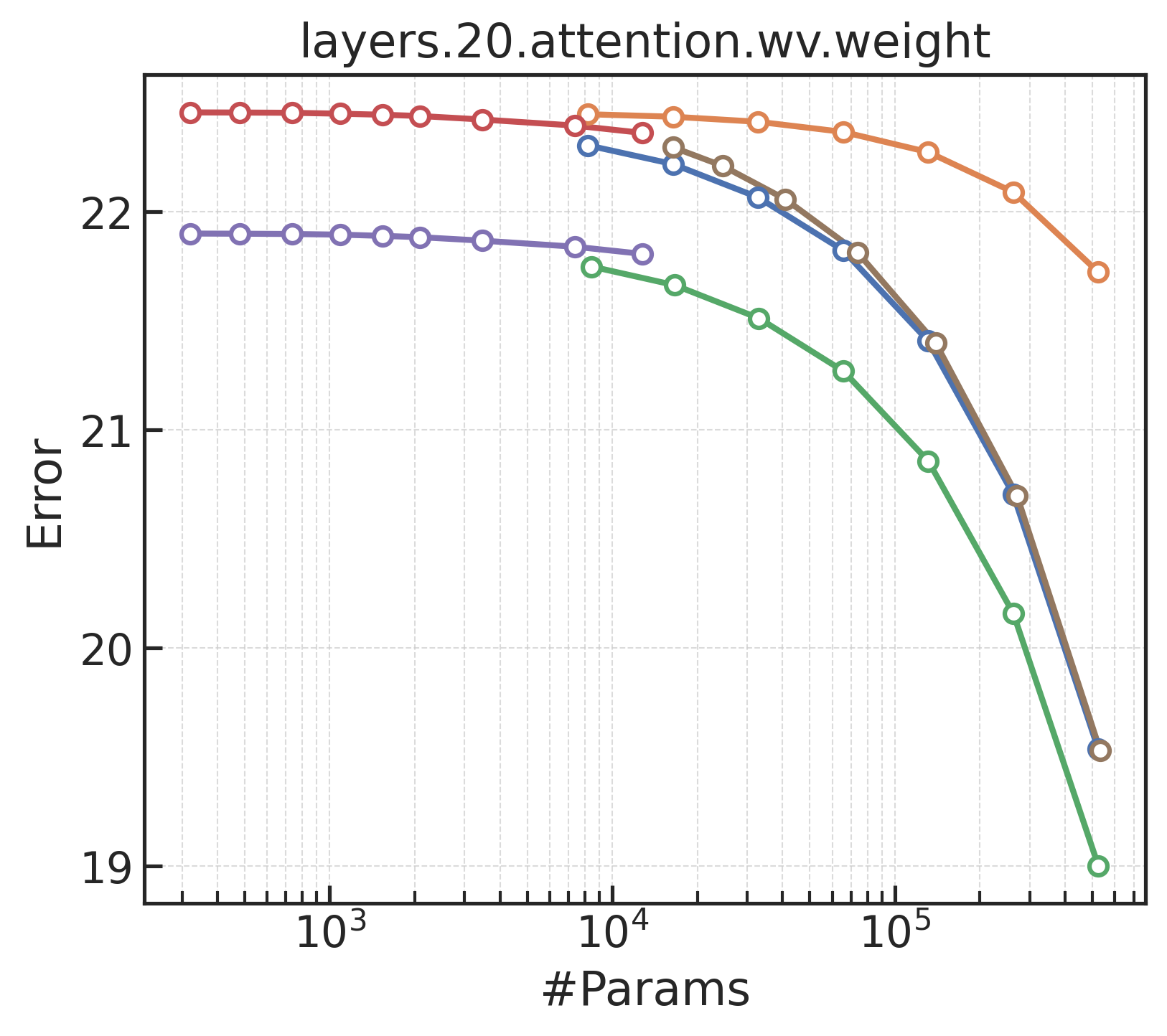}
    \end{subfigure}
    \hfill
    \begin{subfigure}[b]{0.23\linewidth}
    \includegraphics[width=\linewidth]{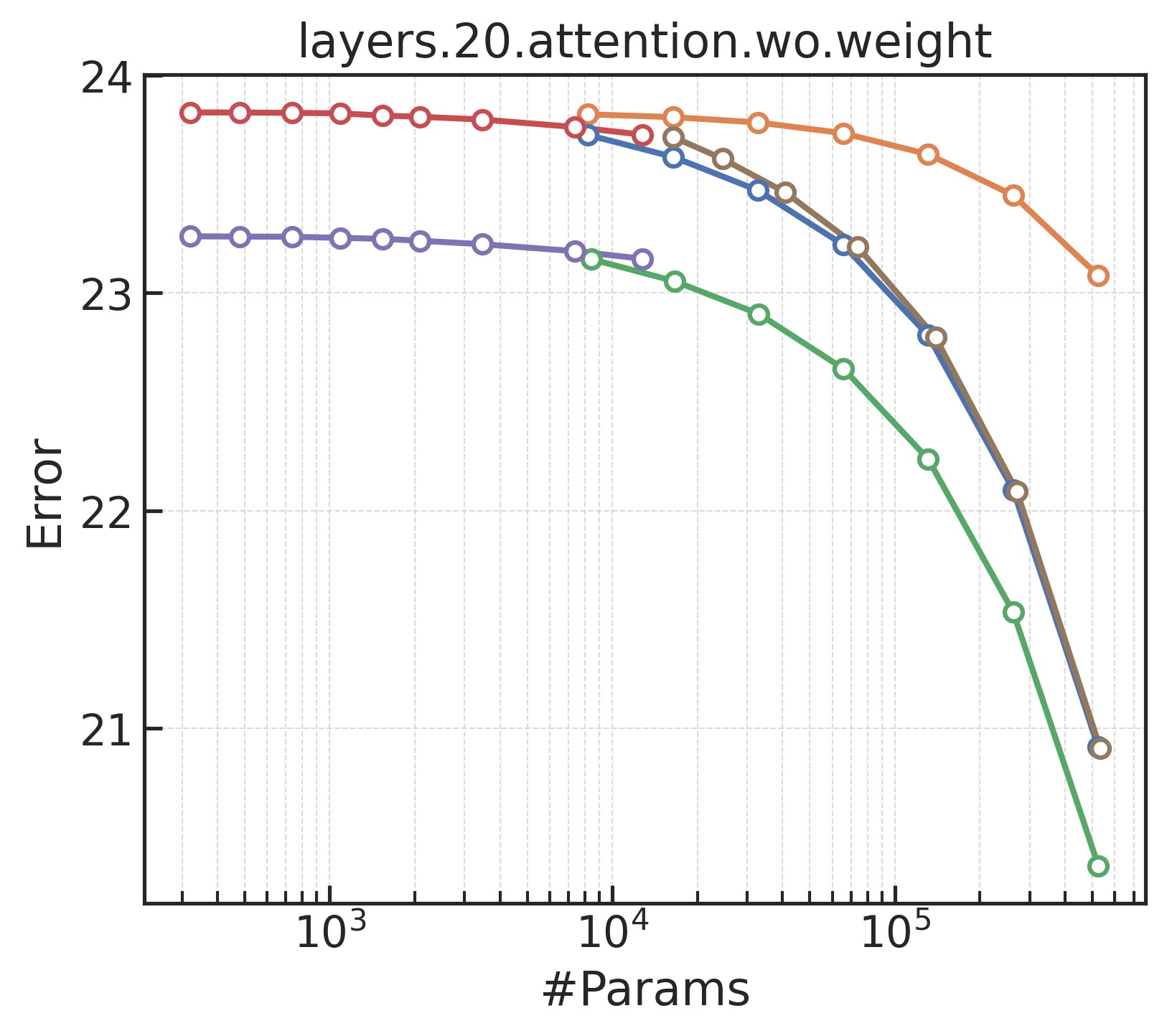}
    \end{subfigure}
    
    \begin{subfigure}[b]{0.23\linewidth}
    \includegraphics[width=\linewidth]{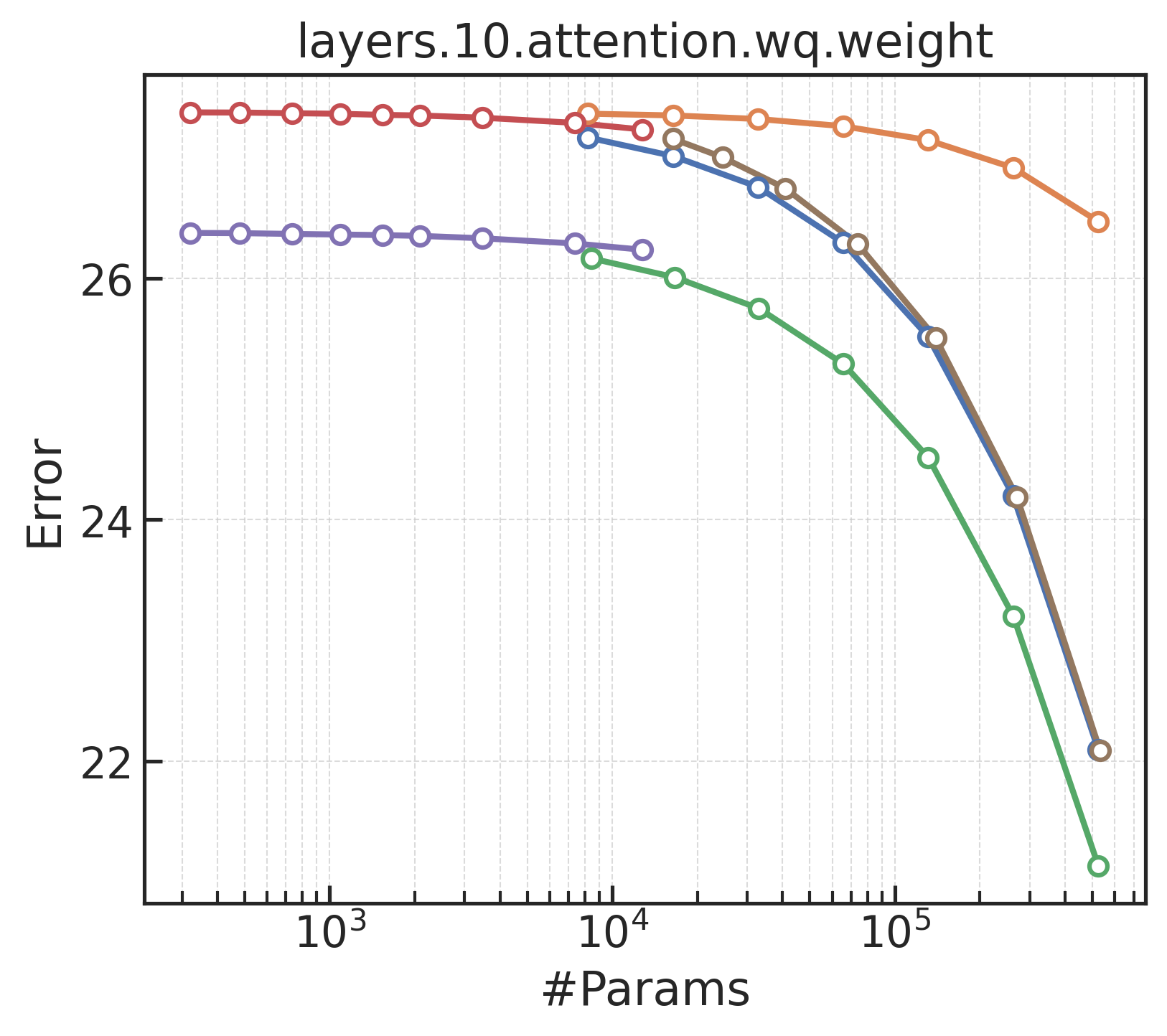}
    \end{subfigure}
    \hfill
    \begin{subfigure}[b]{0.23\linewidth}
    \includegraphics[width=\linewidth]{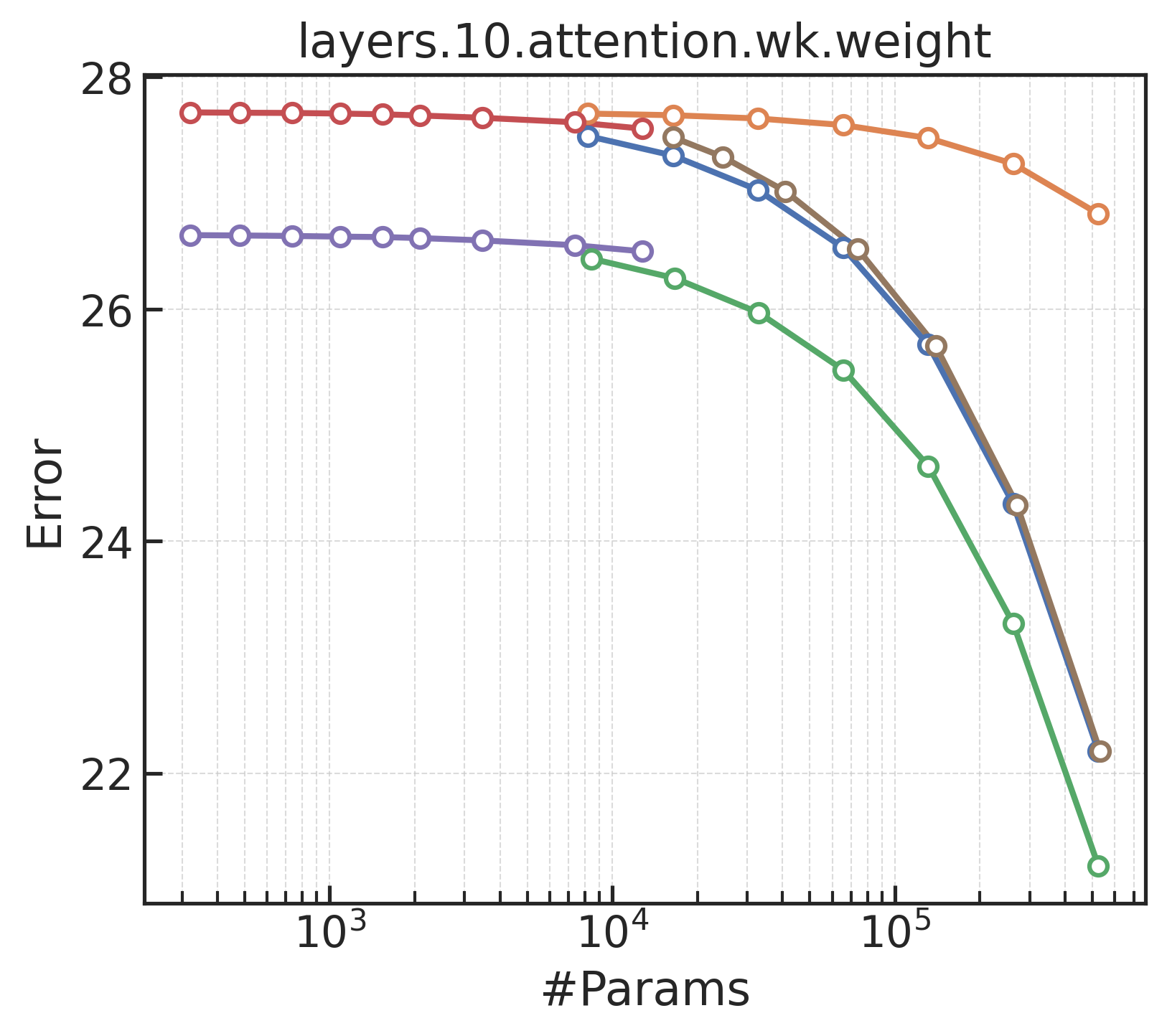}
    \end{subfigure}
    \hfill
    \begin{subfigure}[b]{0.23\linewidth}
    \includegraphics[width=\linewidth]{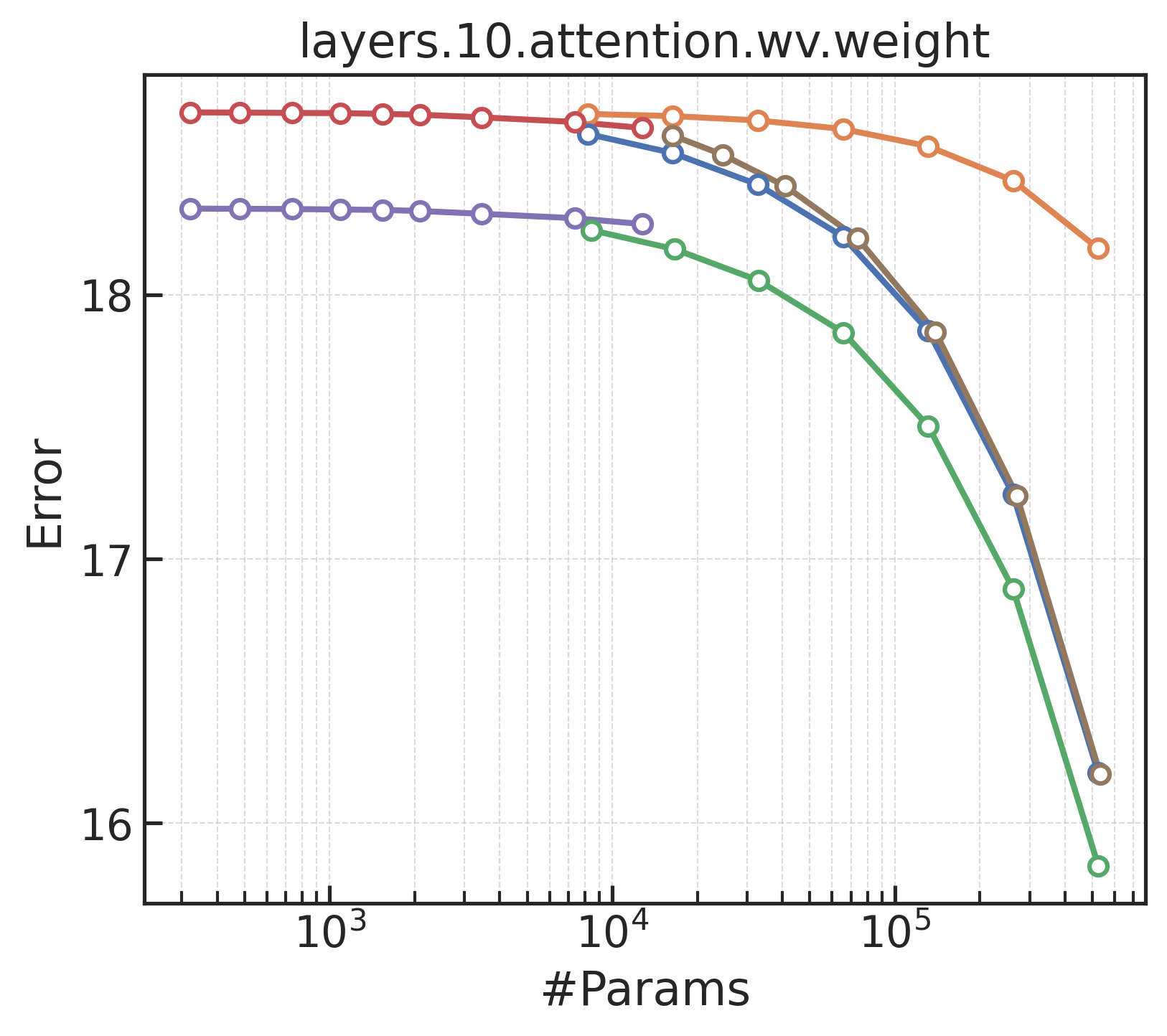}
    \end{subfigure}
    \hfill
    \begin{subfigure}[b]{0.23\linewidth}
    \includegraphics[width=\linewidth]{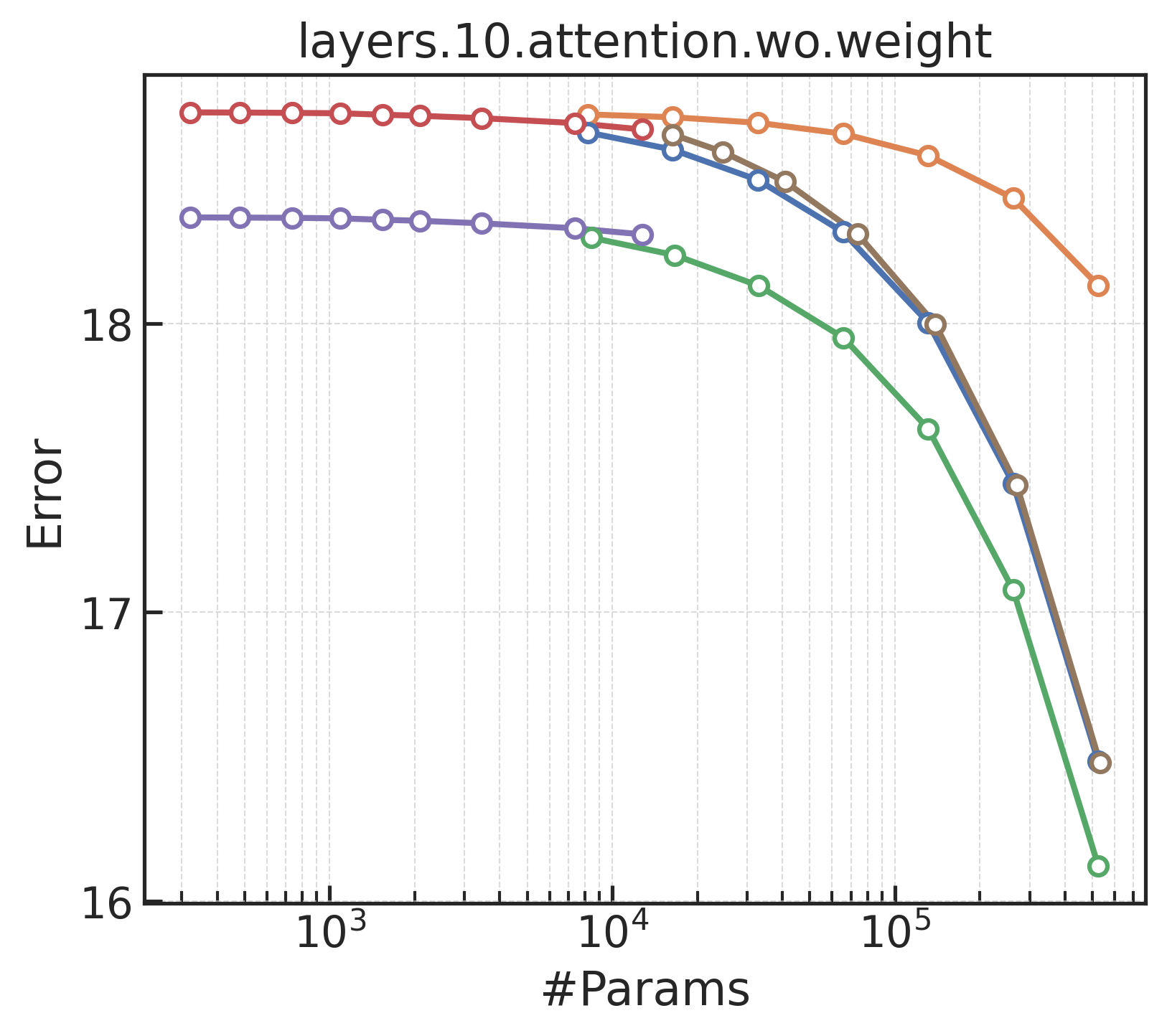}
    \end{subfigure}
    \caption{Simulation on different layers of the Llama2 7B and Llama2-chat 7B models.}\label{fig:llama-weight}
\end{figure*}

%% file: contents/fig_and_tab/db_hyper.tex
\begin{table*}[t]
\centering
  \caption{Hyper-parameters of the subject-driven generation experiment. For OFT, $b$ means the block size. For BOFT, $m$ means the number of Butterfly factors and $b$ means the block size, which is the same with OFT. For ETHER+, $n$ means the number of blocks. For LoRETTA, $r_{\text{LoRA}}$ means the rank of LoRA and $r_{\text{TT}}$ means the rank of TT decomposition. For our method TLoRA, $r_{\text{TRM}}$ means the rank of the TRM transform and $r_{\text{TR}}$ means the rank of the TR residual adaptation.}\label{tab:db-hyper}
    \begin{tabular}{lccccccc}
    \toprule
    Method & LoRA & DoRA & OFT & BOFT & ETHER+ & LoRETTA & TLoRA \\
    \midrule
    Setting & $r$=1 & $r$=1 & $b$=2 & ($m$=2, $b$=2) & $n$=1 & ($r_{\text{LoRA}}$=8, $r_{\text{TT}}$=6) & ($r_{\text{TRM}}$=1, $r_{\text{TR}}$=2) \\
    Learning rate & \num{5e-5} & \num{5e-5} & \num{1e-4} & \num{1e-4} & \num{5e-4} & \num{5e-4} & \num{5e-4} \\
    \bottomrule
    \bottomrule
    \end{tabular}
\end{table*}

\begin{table*}[t]
\scriptsize
\setlength{\tabcolsep}{3pt}
\centering
  \caption{Hyper-parameters of the controllable generation experiment. For OFT, $b$ means the block size. For BOFT, $m$ means the number of Butterfly factors and $b$ means the block size, which is the same with OFT. For ETHER+, $n$ means the number of blocks. For LoRETTA, $r_{\text{LoRA}}$ means the rank of LoRA and $r_{\text{TT}}$ means the rank of TT decomposition. For our method TLoRA, $r_{\text{TRM}}$ means the rank of the TRM transform, $r_{\text{LoRA}}$ means the rank of the LoRA residual adaptation and $r_{\text{TR}}$ means the rank of the TR residual adaptation. $\lambda$ is the scale of regularization.}\label{tab:control-hyper}
    \begin{tabular}{lcccccccccc}
    \toprule
    Method & LoRA & DoRA & OFT & BOFT & ETHER+ & LoRETTA & TLoRA* & TLoRA* & TLoRA & TLoRA \\
    \midrule
    Setting & $r$=1 & $r$=1 & $b$=2 & ($m$=2, $b$=2) & $n$=1 & ($r_{\text{LoRA}}$=8, $r_{\text{TT}}$=6) & ($r_{\text{TRM}}$=2, $r_{\text{LoRA}}$=2) & ($r_{\text{TRM}}$=2, $r_{\text{LoRA}}$=4) & ($r_{\text{TRM}}$=2, $r_{\text{TR}}$=6) & ($r_{\text{TRM}}$=1, $r_{\text{TR}}$=8)\\
    Learning rate & \num{5e-4} & \num{5e-4} & \num{1e-4} & \num{1e-4} & \num{1e-3} & \num{1e-5} & \num{5e-4} & \num{5e-4} & \num{1e-3} & \num{1e-3} \\
    $\lambda$ & - & - & - & - & - & - & 0 & 0 & \num{1e-3} & \num{1e-3} \\
    \bottomrule
    \bottomrule
    \end{tabular}
\end{table*}